\documentclass[10pt,twocolumn,letterpaper]{article}
\usepackage{iccv}
\usepackage{times}

\usepackage[T1]{fontenc}
\usepackage{graphicx}
\usepackage{amsmath}
\usepackage{amssymb}
\usepackage{xspace}
\usepackage{booktabs}
\usepackage{multirow}
\usepackage[dvipsnames]{xcolor}
\usepackage{multirow}

\usepackage[pagebackref=true,breaklinks=true,letterpaper=true,colorlinks,bookmarks=false]{hyperref}
\usepackage[capitalise]{cleveref}
\usepackage{enumitem}

\iccvfinalcopy
\def\iccvPaperID{6874}
\ificcvfinal\fi

\makeatletter
\renewcommand{\paragraph}{%
  \@startsection{paragraph}{4}%
  {\z@}{0.25em}{-1em}%
  {\normalfont\normalsize\bfseries}%
}
\makeatother

\newcommand{\datasetlong}{Replay\xspace}
\newcommand{\dataset}{Replay\xspace}

\title{\datasetlong: \\
Multi-modal Multi-view Acted Videos for Casual Holography\vspace{-1.8ex}}
\author{Roman Shapovalov$^*$
\qquad Yanir Kleiman$^*$
\qquad Ignacio Rocco$^*$
\qquad David Novotny\\
Andrea Vedaldi
\qquad Changan Chen$^\dagger$
\qquad Filippos Kokkinos
\qquad Ben Graham
\qquad Natalia Neverova\\
Meta \qquad UT Austin$^\dagger$ \qquad $^*$\textit{equal contribution}\\
{\tt\small \url{https://replay-dataset.github.io/}}
}

\begin{document}
\maketitle
\ificcvfinal\thispagestyle{empty}\fi

\begin{abstract}
We introduce \emph{\datasetlong}, a collection of multi-view, multi-modal videos of humans interacting socially. Each scene is filmed in high production quality, from different viewpoints with several static cameras, as well as wearable action cameras, and recorded with a large array of microphones at different positions in the room. Overall, the dataset contains over 4000 minutes of footage and over 7 million timestamped high-resolution frames annotated with camera poses and partially with foreground masks.
The \dataset dataset has many potential applications, such as novel-view synthesis, 3D reconstruction, novel-view acoustic synthesis, human body and face analysis, and training generative models. We provide a benchmark for training and evaluating novel-view synthesis, with two scenarios of different difficulty. Finally, we evaluate several baseline state-of-the-art methods on the new benchmark.
\end{abstract}

\vspace*{-2mm}
\section{Introduction}\label{s:intro}

\begin{figure*}[bt]
\centering%
\includegraphics[width=\textwidth]{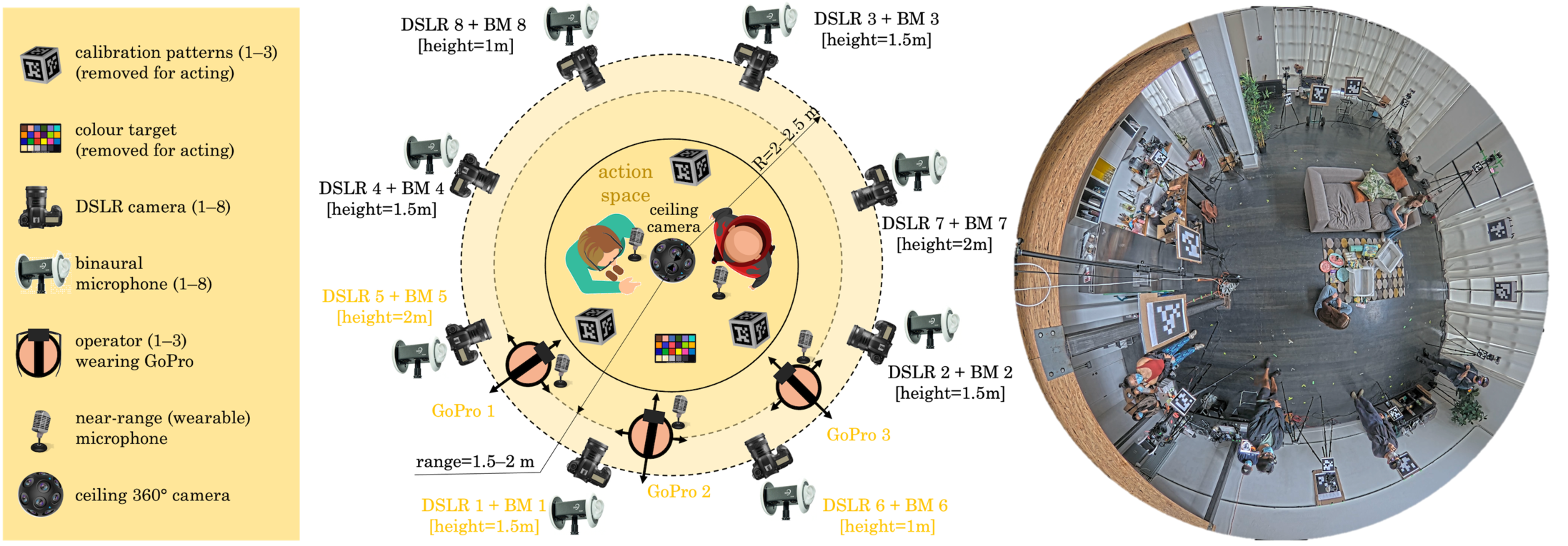}
\caption{
  Recording setup (left) and a frame from the ceiling camera during capture (right).
  Actors wearing near-range microphones are located in the centre of the scene; they are surrounded by a ring of static DSLR cameras paired with binaural microphones.
  Operators wearing GoPros and microphones are standing in front of the actors at around 2\,metres distance during the acting phase, and during the fly around phase they go around the scene filming semi-static actors.
  The colour of sensor labels reflects their usage in new-view synthesis benchmarks (\cref{s:experiments}): fly around is trained using GoPro-2 frames while evaluated on all DSLRs;
  we define an `acting benchmark' using the 6 frontal sensors (gold), of which DSLR-1 is held out for testing, and the other 5 sensors are used for training.}%
  \label{fig:splash}
\end{figure*}

A staple of science fiction is to relive past events and memories as holograms.
With technological advances in virtual, mixed and augmented reality, this vision is ever closer to become a reality.
We can now think of recording an event with a pair of AR glasses instead of a camera, and relive it later as a 360$^\circ$ re-projection in a real or virtual space.
However, there are still major technical hurdles before this can be done reliably and with sufficient quality.

High-fidelity 3D reconstruction remains one of the primary obstacles.
Given a casual recording from a single sensor like a pair of AR glasses, it is in general not possible to reconstruct the content in 3D.
Monocular data, or even data collected from cameras with a short baseline, simply does not contain sufficient information for 360$^\circ$ reconstruction.
For example, in such setup, it is not possible to observe simultaneously the front and back of an object.
Furthermore, reconstructing appearance is not enough: any engaging user experience also requires to reconstruct sounds, so the problem is inherently multi-modal.

Consumer holography requires to compensate for the intrinsic limitations of a casual data capture setup via machine learning.
However, despite the success of neural rendering~\cite{mildenhall20nerf:}, even the bests method struggle to reconstruct complex, long dynamic content from a monocular sensor.
Furthermore, none of these approaches tackles multi-modal reconstruction yet.


Here, we suggest that further progress in casual holography, and in general in the reconstruction and generation of realistic 4D (3D + time) multi-modal content, is severely hampered by the lack of suitable datasets.
We address this gap by introducing \textit{\datasetlong}, a new large dataset to study the problem of multi-modal new-view synthesis for long captures of acted dynamic content.
\dataset contains long scenes in a natural indoor environment (living room, dining room, \etc), where multiple people are interacting with props and with each other and performing a variety of activities such as exercising, playing games, or chatting.
Each scene is several minutes long, and is filmed in 4K resolution with 8 static DSLR cameras and 3 head-mounted GoPro cameras that capture the scene from all view points, allowing the evaluation of scene reconstruction from the view points that significantly differ from the source video.
For each scene, we also provide a semi-static \emph{fly-around} sequence, where the actors pause and remain still while the head-mounted camera operators walk around them.
In addition, the scene is recorded with a large array of microphones to allow novel view acoustic synthesis\,\cite{chen2023nvas}.
All sensors are temporally calibrated, and cameras are also color- and view-calibrated as well. In addition, metadata such as foreground segmentation masks is provided for some of the scenes.
The data is collected with actors' consent, addressing privacy concerns, and will be public for non-commercial research.

This paper focuses on the visual component of \dataset;
the audio part of the dataset is introduced and used for novel-view \textit{acoustic} synthesis by Chen et al.\,\cite{chen2023nvas}.
The \dataset videos, in turn, constitute a notable step up compared to existing datasets for static and dynamic novel view synthesis; so far methods have been evaluated on short sequences with a limited range of view points.
For example, the popular Dynamic Scene Dataset~\cite{yoon20novel}, which is often used to evaluate dynamic new-view synthesis, contains short scenes ($\approx$5 sec) sampled at low FPS (30 frames in total), and captured by static cameras where the farthest two cameras are about one meter apart.
Other datasets such as ZJU-Mocap~\cite{omran18neural} and AIST++~\cite{li21learn}, provide longer videos with a large variety of view points, but are human-centric and contain people with an empty background and no additional objects, which makes them less useful for evaluating full scene reconstruction.
None of these datasets contain naturally-acted events with sounds.

Due to the richness of scenes, actors, sensors and modalities, \dataset can be used to define a large variety of different tasks in multi-modal new-view synthesis.
The most direct setting for casual holography is to reconstruct a scene from a single head-mounted camera; then, reconstruction quality can be assessed in a 360$^\circ$ manner by using the static DSLRs cameras or the other head-mounts for evaluation.
However, tasks of various complexity can also be defined, such as reconstruction from any combination of static and moving cameras.
Furthermore, the fly-around segments at the beginning of each sequence can be used to test reconstruction in a decidedly simpler (semi-static) setting, and to simplify the reconstruction of the dynamic part as well.

Using \dataset we define two such benchmark tasks of increasing difficulty and assess various existing techniques on them.
Specifically, we consider baselines representing different families of radiance-field models (NeRF~\cite{mildenhall20nerf:}, TensoRF~\cite{chen22tensorf:}) and their extensions dealing with dynamic scenes (NeRF+time, HexPlane~\cite{cao23hexplane:}, Nerfies~\cite{park21nerfies:}).

\section{Related Work}%
\label{s:related}

\paragraph{Datasets for dynamic new-view synthesis.}

With the explosion of neural rendering, many datasets for studying new-view synthesis of dynamic content were proposed.
Focusing on humans,
HumanEva~\cite{sigal10humaneva:},
Human3.6M~\cite{ionescu14human3.6m:},
AIST~\cite{tsuchida19aist},
AIST++~\cite{li21learn}, and
ZJU-Mocap~\cite{peng20neural}
portray a single person in isolation, without context, performing scripted motion.
In contrast, \dataset contains groups of humans acting naturally in a familiar environment.

More complex multi-view data containing dynamic humans in context include the Immersive Light Field dataset~\cite{broxton20immersive}, which contains sixteen scenes captured from approximately 46 calibrated cameras.
The NVIDIA Dynamic Scene Dataset~\cite{yoon20novel} contains eight videos captured with 12 (mostly front-facing) calibrated GoPro Black Hero 7 cameras.
The UCSD Dynamic Scene Dataset~\cite{lin21deep} contains 96 videos collected in a similar manner to~\cite{yoon20novel}, but using 10 cameras.
The Plenoptic Video dataset~\cite{li22neural} provides 6 more scenes from 21 cameras.
All such videos are complex and visually diverse, but they all capture a single short-duration activity (typically 1 or 2 minutes at most).

Some datasets are collected using domes, and therefore do not contain natural environments or moving cameras.
Panoptic Studio~\cite{joo17panoptic} contains 3 hours of recordings of humans engaged in multiple social activities captured with roughly 500 cameras and depth sensors.
NeuralDome~\cite{zhang22neuraldome:} contains videos of a single human manipulating an object captured from 76 cameras, and additional sensor-based participant tracking data.
Our \dataset focuses on long sequences with professional actors in a familiar setting.
Furthermore, the usage of head-mounts makes our dataset particularly well-suited for studying scene reconstruction from a single egocentric device, which is one of the most realistic settings for future applications in casual holography.

Finally, all datasets above focus on visual reconstruction, and are thus not multimodal like \dataset.
See also \cref{f:dataset_comparison} for a schematic summary.

\paragraph{Reconstructing 3D dynamic humans.}

Reconstructing a 4D video remains a challenging problem, so many authors have focused on special cases, such as reconstructing individual humans.
Much work has focused on modelling articulated human bodies, including
Neural volumes~\cite{lombardi19neural},
Relightables~\cite{guo19the-relightables:},
Articulated Neural Rendering~\cite{raj20anr:},
A-NeRF~\cite{su21a-nerf:},
Neural Actor~\cite{liu21Bneural},
H-NeRF~\cite{xu21h-nerf:},
Neural Performer~\cite{kwon21neural},
Deep Dynamic Character~\cite{habermann21real-time},
Human Re-rendering~\cite{sarkar21neural},
Pixel Aligned Avatars~\cite{raj21pixel-aligned},
HumanNeRF~\cite{weng22humannerf:},
HiFi Human Avatar~\cite{zhao22high-fidelity},
Generative Neural Articulated RFs~\cite{bergman22generative},
Animatable NeIS~\cite{peng22animatable}.
Most of these works approach the problem by explicitly tracking the human body, usually by using SMPL~\cite{loper15smpl:} fits, and then modelling shape and appearance in a canonical, articulation-free, space.
Other works, including
Neural Head~\cite{grassal21neural},
Dynamic Head~\cite{wang21learning},
Dynamic Neural Faces~\cite{gafni21dynamic},
MoRF~\cite{wang22morf:},
specialise in reproducing heads, and a few such as
Artemis~\cite{luo22artemis:}
explore other animals.
Since our scenes contain several interacting humans and objects, these methods are not applicable to our problem because they focus on isolated reconstructions of specific object classes.


\paragraph{Reconstructing generic 4D videos.}

Several authors have considered the problem of reconstructing generic 4D videos.
Some have proposed to capture directly the lightfield, with no or partial understanding of scene geometry.
Examples include~\cite{gortler96the-lumigraph,chai00plenoptic,buehler01unstructured,zitnick04high-quality,kanade97virtualized,bansal204d-visualization,kalantari16learning-based,srinivasan19pushing,zhang21editable,stich08view,broxton20immersive,suhail21light}.

Other methods model shape more explicitly, often using dynamic generalizations of NeRF~\cite{mildenhall20nerf:}.
These are the most applicable to the \dataset scenarios.
Many of them, including
D-NeRF~\cite{pumarola20d-nerf:},
Deformable NeRF~\cite{park20deformable},
Dynamic NVS~\cite{yoon20novel},
Nerfies~\cite{park21nerfies:},
HyperNeRF~\cite{park21hypernerf:},
Neural Trajectory Fields~\cite{wang21neural},
NR-NeRF~\cite{tretschk21non-rigid},
NSFF~\cite{li21neural},
NeRFlow~\cite{du21neural},
STaR~\cite{yuan21star:},
NeRFPlayer~\cite{song22nerfplayer:},
Deformable Voxel Grid~\cite{guo22neural},
TiNeuVox~\cite{fang22fast},
DynIBaR~\cite{li22dynibar:},
DeVRF~\cite{liu22devrf:}
attempt to estimate a deformation field and thus explicitly model the motion in the scene, reducing the video to a single canonical reconstruction and the deformation field.
While this is statistically parsimonious, and necessary for reconstruction when the number of input viewpoints is small, estimating a correct deformation field is difficult due to the underconstrained nature of the problem.

Other methods directly add time to the radiance field parametrisation (sometimes called NeRF+$t$), thus avoiding the challenge of explicitly estimating deformations.
Examples include
NERF-W~\cite{martin-brualla21nerf},
NeuralDiff~\cite{tschernezki21neuraldiff},
Video NeRF~\cite{xian21space-time},
Dynamic View Synthesis~\cite{gao21dynamic},
Fourier PlenOctrees~\cite{wang22fourier} and
DyNeRF~\cite{li22neural}.
Finally two recent concurrent works,
K-Planes~\cite{fridovich-keil23k-planes:} and
HexPlane~\cite{cao23hexplane:}, extend the voxel grid decomposition introduced in EG3D~\cite{chan22efficient} and TensoRF~\cite{chen22tensorf:} to spatio-temporal 4D grids. 

Many of these methods explicitly require a large number of viewpoints.
In some cases, this requirement is indirect~\cite{gao22monocular}, in the sense that methods may also work from a monocular camera, but only if the camera motion dominates the scene motion.


\section{The \dataset Dataset}\label{s:dataset}

\newcommand{\inches}{${}^{\prime\prime}$}
\newcommand{\degrees}{${}^\circ$}
\newcommand{\numscenes}{68\xspace}

\begin{table*}[]
{\footnotesize
\setlength{\tabcolsep}{4pt}
\begin{tabular}{@{}lllllllll@{}}
\toprule
Dataset          & \#Sc. & Viewpoints                 & Resolution                 & Motion & Angles   & Dur.    & \#Act.               & Motion types                                                                                                                      \\ \midrule
Dynamic Scene Dataset     & 8                & 1 moving camera                  & 1920$\times$1080                   & Dynamic              & Frontal  & 5sec          & 1--4                    &      Simple body motions (facial, jump, etc.)                                                                                          \\
ZJU-Mocap        & 10               & 21 static cameras                & 1024x1024                  &       Dynamic               & 360$^\circ$    &   20sec            & 1                    & Simple body motions (punch, kick, etc.)                                                       \\
AIST++           & 1408             & 9 static cameras                 & 1920x1080                  & Dynamic              & 360$^\circ$         & 20-50sec      & 1                    & Dancing                                                                       \\
Ours: flyaround & 46               & 1 moving, 8 static       & \multirow{2}{*}{3840x2160} & $\approx$Static              & \multirow{2}{*}{360$^\circ$} & 40-60sec & \multirow{2}{*}{1--4} & Dancing, chatting, playing video games,\\
Ours: acting    & 46               & 3 quasi static, 8 static &                            & Dynamic              &                      & 3-5min & & unwrapping presents, playing ping pong                                                                                                           \\ \bottomrule
\end{tabular}
}
\caption{Comparison with related datasets. For each dataset, we report the number of scenes (\textit{\#Sc.}, which may be recorded from multiple (\textit{viewpoints}) at different resolution, contain varying amount and type of \textit{motions}, filmed either from the frontal position or from around the scene. Our dataset uniquely has multiple actors per scene (\textit{\#Act}) and duration (\emph{Dur}.) of several minutes. Our datased has a natural background (as opposed to a studio or dome), and provided foreground masks include not only actors but also objects they are interacting with. 
}\label{f:dataset_comparison}
\end{table*}

The full \datasetlong dataset consists of \numscenes scenes of social interactions between people, such as playing board games, exercising, or unwrapping presents.
Each scene contains about 5 minutes of acting following a few minutes of calibration stages, and is filmed with 12 cameras, static and dynamic.
Audio is captured separately by 8 binaural microphones and additional near-range microphones for each actor and for each egocentric device.
All sensors are temporally synchronized, undistorted, geometrically calibrated, and color calibrated.

In addition to the full dataset, we introduce the \textit{\dataset novel view reconstruction benchmark}, a curated subset of scenes with given training and test sets and supplemental information such as foreground/background masks.
We run several state-of-the-art novel view reconstruction methods on this benchmark and report the results in \cref{s:experiments}.

\paragraph{Content.}

The videos depict human social interaction in a large variety of indoor settings and contexts.
Examples include meeting friends, talking, sitting in a living room, making hand gestures, playing charades, exercising on a yoga mat, playing video games, playing board games, arranging ornaments, having a meal, unwrapping presents, and more.
Each scene contains up to 4 actors, with a total of 42 actors of diverse age, gender, and ethnicity across the scenes. In particular, 21 of them are white, 11 are Asian, 5 are black, and 5 are mixed race.

\paragraph{Scene setup.}

In each scene, there are three human operators wearing wearable \textit{egocentric} cameras that provide eye-level views of the scene. The focus on monocular and binocular wearable cameras and microphones is a unique feature of \dataset, which enables evaluating methods targeting AR/VR applications where a scene captured by one wearable device may have to be rendered in a world-locked manner on another device, from a new viewpoint.

In addition, the scene is shot by 8 static DSLR cameras arranged in a full circle around the action, approximately 45\degrees\,apart from one another.
Each static DSLR camera has a binaural microphone attached to it, and each actor and egocentric camera operator is equipped with a near-range microphone.
We also provide an auxiliary capture of the entire scene with a 360\degrees\ ceiling camera. This capture is not intended to be used as input, and is included to provide an overview of the scene for users of the dataset.
\Cref{fig:splash} shows a bird-eye view of the scene setup, as well as a schematic representation.

\paragraph{Phases.}

Scenes are divided into three logical phases: calibration, flyaround, and acting.
The \emph{Calibration} phase is part of the scene setup, and contains images of calibration patterns and of the digital clapper.
%
%
The \emph{Flyaround} phase 
shows the actors take their place in the scene and remain still.
Then, the wearable camera operators walk around the scene while looking at the central action area.
This provides a continuous 360\degrees{} view of a  scene which remains nearly static.
This stage takes 40--60 seconds.
Finally, during the \emph{Acting} phase the actors perform for about 3--5 minutes.

To reduce the amount of data necessary to process for the benchmark, we limit the fly-around part of the scene to 40 seconds and the acting part to 60 seconds.
We further segment the acting part into two 30-seconds segments, since we found that most state-of-the art-methods are incapable of handling longer sequences.
However, we strongly encourage future users of the dataset to test reconstructing at lest an entire minute.


\paragraph{Sensors.}
As shown in \cref{fig:splash}, \dataset uses several sensors. These are:
Eight static DSLR cameras (Sony A7 III\@; 24\inches{} lens; 30 FPS\@; 4K resolution);
Eight 3Dio binaural microphones co-located with the DSLR cameras;
A ceiling camera (AXIS M4308-PLE: wide angle, 30 FPS, 2880$^2$ resolution, circular frame);
Three GoPro cameras (Hero 9; RAW model, 60 FPS, 4K resolution);
And a near-range lavalier microphone for each camera operator and actor in the scene.


\paragraph{Publicly-available assets.}



We will make the pre-processed data available to resaerchers upon publication, delivering the following assets.
For each imaging sensor $s$, we provide
(1) a collection of video frames $I_{st}$ indexed by $t$;
(2) the distortion and intrinsic calibration parameters $(\rho_s,K_s)$;
(3) the camera pose $\pi_{st}$ with respect to the scene reference frame;
(4) and, for 10 of the scenes, foreground segmentation masks $M_{st}$ for each frame, including furniture, actors, and objects they interact with.
For each audio sensor $a$, we provide
(1) a collection of audio frames $A_{at}$ indexed by $t$;
(2) the location $\pi_{at}$ of the sensor (which usually coincides with a certain imaging sensor), except for the near-range microphones, which are mounted on the actors and camera operators, whose dynamic location is therefore difficult to estimate.
All sensors are temporally synchronised; for this, we provide the time information $\tau_{st}$ and $\tau_{at}$ for each video and audio frame.

We also provide benchmark definitions (\cref{s:bench}) and corresponding evaluation code.


\section{Data collection}\label{s:collection}

Collecting a dataset such as \dataset{} is a major endeavour.
We describe the key aspects of the data collection to better understand the properties of the dataset, and because they can be helpful for other researchers that wish to engage in a similar experimental activity.

\subsection{Production}

The data was produced with the help of a vendor who took care of finding locations, ordering hardware, hiring professional actors, running the filming, quality assurance, and assigning basic metadata.
Production lasted for more than six months, and resulted in the collection of 119 scenes, of which \numscenes have been processed so far to be released.
The vendor was instructed to calibrate sensors before recording each scene, as described below.

To be able to diversify recording locations and keep natural backgrounds, we required a relatively mobile capture setup that (contrary to a dome) required non-trivial calibration and synchronisation from scratch before each recording.
The large amount of sensors of different types and setup phases significantly increased the likelihood of human and equipment failures, such as camera or microphone malfunction, accidental camera movement during filming, wrong focus or focal length, missing or low-quality setup step for a specific camera, \etc
Secure data storage and transfer was also a challenge due to the data volume (60 GB per scene).

\subsection{Processing}

The captured data required substantial pre-processing, including, in order: intrinsic calibration, temporal synchronisation, temporal segmentation, extrinsic calibration, photometric calibration, foreground/background segmentation.
The principal challenges and solutions are discussed next.

\paragraph{Intrinisic calibration.}

The focal length, principal point, and lens distortion of each sensor was estimated by asking the vendor to show a moving ChARuco calibration board at least once to each sensor, and for latter recordings, before recording each scene.
A ChARuco board is a commercially available checkerboard combined with ARuco tags \cite{aruco-Garrido2014}, which allows disambiguating the pose of the board in camera coordinates.
These  estimations were then used initialise the cameras in the COLMAP SfM software~\cite{schonberger16structure-from-motion}, which then refined the intrinsic parameters through joint optimisation with camera poses; see below for details.

\paragraph{Temporal synchronisation.}
Accurate temporal synchronisation of the various sensors is crucial for training and evaluating a dynamic scene from multiple view points.
While specific sensors support hardware-based synchronisation, this is inapplicable to our heterogeneous mix of sensors.
Instead, the sensors are synchronised using their audio signature. We take a salient segment of the audio recorded by each sensor, and match it with the entire audio of a second sensor using cross correlation. To increase robustness, we take several segments from different positions in the audio file and use a majority vote to estimate the true offset between each two sensors.
After the computation of the pairwise correspondence, we check the stability of the estimated offsets by analysing cycles of three or more sensors. That allows us to identify sensors which were not matched correctly, due to errors in the data collection, for example noisy cameras or cameras which did not record audio due to technical issues. In such cases, manual temporal synchronisation was required for specific sensors in a small portion of the shots.


\paragraph{Camera pose calibration.}

Camera poses were estimated using COLMAP~\cite{schonberger16structure-from-motion} with several improvements for robustness and scalability.
Specifically, camera intrinsics were initialising via  ChARuco calibration (see above); we then fixed the principal point, but let COLMAP refine the focal length and distortion parameters.
The environment was first reconstructed using the head-mounted camera from the fly-around phase of the capture at a low frame rate (3 FPS).
This produced a sufficiently small number of frames ($\sim$1,000) with sufficient parallax for COLMAP to run successfully.
Then, all the other frames in the capture were triangulated against this initial reconstruction after masking out image regions prone to contain dynamic objects (instances of person, cat, dog segmented using PointRend~\cite{kirillov20pointrend:}).
Finally, since the absolute (and thus relative) positions of the static cameras are constant, they are assumed to form a rig; we thus use bundle adjustment with rig constraints to reconstruct them.
For environments that did not provide sufficient texture detail to robustly triangulate the DSLR poses, we further refined calibration of the DSLR cameras by showing to pairs of them a ChARuco board.

\paragraph{Photometric Calibration.}

We bring all frames from different sensors in the same sRGB colour space, despite differences in the sensor type, factory calibration, and possible processing steps over which we had no control.
This was done by
(1) using the color model of each sensor to map colors to linear space;
(2) filming a reference colour chart from each camera and fitting a linear map to align colours between sensor pairs;
(3) moving colours back to sRGB space.

\paragraph{Foreground segmentation.}

We provide high-quality foreground masks for part of the dataset.
Fully automatic foreground segmentation in videos is still an extremely challenging research problem.
We therefore adopted a human-in-the-loop approach, leveraging the state-of-the-art XMem~\cite{cheng22xmem:} video segmenter.
The model extends scribbles annotated by an operator to generate a segmentation mask for each video frame, and then automatically propagates masks to the subsequent frames.
When an error is detected, the operator intervenes drawing additional scribbles to correct the current prediction, and then re-initiating the propagation process.
Since the definition of foreground objects is open to interpretations, we made sure the same person worked on each scene, ensuring consistency within and across its videos.
Although the process requires only a limited intervention, we still found it time consuming due to the long duration of the videos and the number of cameras per scene.
Thus, only 10 scenes are annotated with foreground masks.

\section{Experiments}%
\label{s:experiments}

After defining two benchmarks (\cref{s:bench}), we conduct several experiments on \dataset for the purpose of demonstrating its application to the development of new-view synthesis methods, as well as to assess current state-of-the-art neural rendering techniques (\cref{s:baselines}) on this challenging data (\cref{s:results}).

\subsection{Benchmarks}%
\label{s:bench}

We define two novel-view synthesis benchmarks using \dataset:
\textit{flyaround}, with semi-static actors filmed from $360^{\circ}$ trajectory, and the more challenging \textit{acting}, where naturally behaving actors are filmed with frontal cameras.

\begin{table}[b]
\centering
{\small
\begin{tabular}{@{}lcccccc@{}}
\toprule
\multirow{2}{*}{Method} & \multicolumn{3}{c}{Flyaround ($\approx$static)} & \multicolumn{3}{c}{Acting (dynamic)} \\ \cmidrule(lr){2-4}\cmidrule(l){5-7} 
                        & {\footnotesize PSNR}          & {\footnotesize IOU}           & {\footnotesize LPIPS}        & {\footnotesize PSNR}       & {\footnotesize IOU}       & {\footnotesize LPIPS}      \\ \midrule
NeRF~\cite{mildenhall20nerf:}            & 21.85         & \textbf{0.95}        & \textbf{0.22}        & 20.22      & 0.93      & 0.23      \\
NeRF+t                  & 20.86         & 0.92        & 0.25        & \textbf{21.28}      & \textbf{0.94}      & \textbf{0.22}      \\
TensoRF~\cite{chen22tensorf:}          & 20.58         & 0.92        & 0.22        & 17.26      & 0.78      & 0.42      \\
HexPlane~\cite{cao23hexplane:}         & 15.08         & 0.87        & 0.29        & 17.66      & 0.72      & 0.44      \\
Nerfies~\cite{park21nerfies:}         & \textbf{23.22}         & N/A          & 0.61        & 18.08      & N/A        & 0.71       \\ \bottomrule
\end{tabular}
}
\caption{Quantitative results on the two proposed benchmarks. Please note that the numbers are not comparable across benchmarks: in \textit{flyaround}, methods have to model a wider range of viewpoints. In \textit{acting}, all training and evaluation cameras are frontal, but the methods have to model the dynamic geometry of the scene.\vspace{-0.2cm}}\label{tab:results}
\end{table}

\paragraph{Flyaround.}

This simpler benchmark allows evaluating standard reconstruction approaches such as neural radiance fields that cannot model time, alongside with those that model time dependency and shape deformations.
The 40-second segments are extracted from each of the \dataset sequences (see \cref{s:dataset}).
Training frames are extracted from one dynamic wearable camera at 30 FPS (GoPro-2 in \cref{fig:splash}), and all 8 static DSLR cameras are used for evaluation.
This amounts to 1,200 training and 64 evaluation frames per scene.
Despite the fact that there is no significant motion in the scene in these segments, the task is already quite challenging since, unlike typical NVS datasets, we require \textit{extrapolating} beyond camera the trajectory, as the DSLRs are located far away from the trajectory of the wearable camera used for capture.

\paragraph{Acting.}

This is the most challenging setting since the actors are allowed to move freely, while the operators of wearable cameras stand fairly still in front of them.
Monocular reconstruction in this context is still beyond the capability of state-of-the-art reconstruction methods, so in this instance we consider a multi-view reconstruction setup.
We consider a 30-second segment sampled at 30 FPS, which is significantly longer than the data used for testing modern deformable NVS methods, so it stretches their limits.
We hold out one static DSLR for evaluation, while using two relatively frontal DSLRs (DSLR-5 and DSLR-6 in \cref{fig:splash}) and 3 wearable cameras for training, resulting in 4,500 training frames and 50 randomly sampled evaluation frames per scene.
While this scenario is more challenging because of the changing geometry of the scene, all training and evaluation sensors are located in front of the actors, so that the methods do not have to generalise to a wide range of viewing angles.

\begin{figure*}[th]
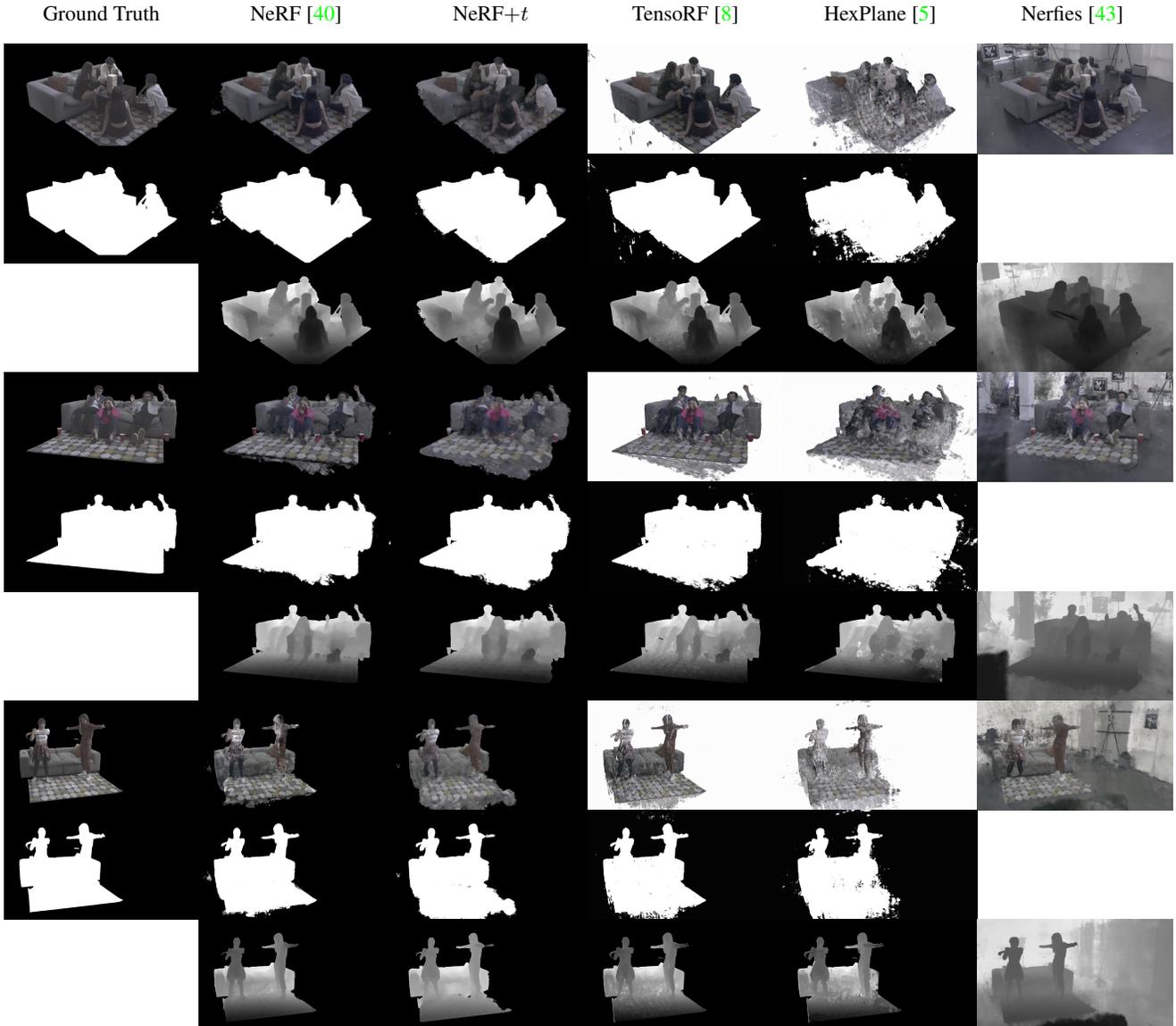

\centering%
\graphicspath{{figure/qual-flyaround/}}

\newcommand{\methcolimw}{2.9cm}%
\newcommand{\methcolimh}{1.55cm}%
\newcommand{\methodcol}[1]{%
\IfFileExists{figure/qual-flyaround/#1}
{\includegraphics[width=\methcolimw,height=\methcolimw,keepaspectratio]{#1}}{}%
}%
\newcommand{\examplerow}[2]{
\methodcol{gt\_#1\_#2.png}&%
\methodcol{maskednerf\_#1\_#2.png}&%
\methodcol{maskedtimenerf\_#1\_#2.png}&%
\methodcol{tensorf\_#1\_#2.png}&%
\methodcol{hexplane\_#1\_#2.png}&%
\methodcol{nerfies\_#1\_#2.png}%
}%

\setlength\tabcolsep{0cm}%
\renewcommand{\arraystretch}{0.0}
\centering\small%
\begin{tabular}{cccccc}
Ground Truth &%
NeRF \cite{mildenhall20nerf:} &%
NeRF$+t$ &%
TensoRF \cite{chen22tensorf:} &%
HexPlane \cite{cao23hexplane:} &%
Nerfies \cite{park21nerfies:} \vspace{0.3cm}\\%
\examplerow{\detokenize{SC-1018_DSLR-5_frame0110333}}{render}\\
\examplerow{\detokenize{SC-1018_DSLR-5_frame0110333}}{mask}\\
\examplerow{\detokenize{SC-1018_DSLR-5_frame0110333}}{depth}\\
\examplerow{\detokenize{SC-1015_DSLR-6_frame0091033}}{render}\\
\examplerow{\detokenize{SC-1015_DSLR-6_frame0091033}}{mask}\\
\examplerow{\detokenize{SC-1015_DSLR-6_frame0091033}}{depth}\\
\examplerow{\detokenize{SC-1005_DSLR-1_frame0107467}}{render}\\
\examplerow{\detokenize{SC-1005_DSLR-1_frame0107467}}{mask}\\
\examplerow{\detokenize{SC-1005_DSLR-1_frame0107467}}{depth}
\end{tabular}

\let\methcolimw\undefined%
\let\methcolimh\undefined%
\let\methodcol\undefined%
\let\examplerow\undefined%

\caption{Qualitative results on the flyaround (semi-static) phase of 3 different scenes. Each section contains 3 rows: rendered RGB image, rendered opacity mask, and rendered depth map. Note that NeRFies is not trained to produce opacity masks, so we skip these renders.}\label{fig:qual-flyaround}
\end{figure*}

\begin{figure*}[th]
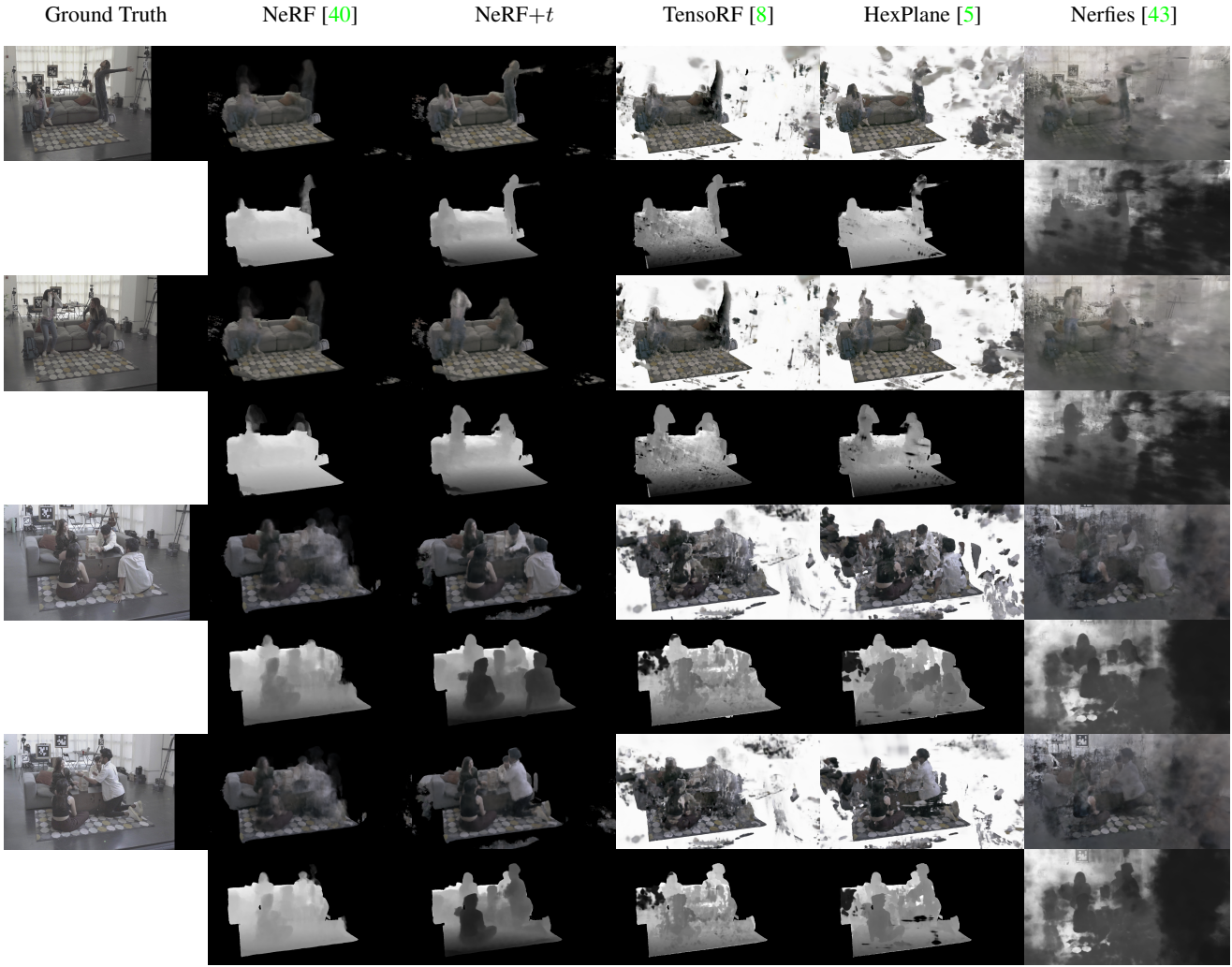

\centering%
\graphicspath{{figure/qual-acting-1/}}

\newcommand{\methcolimw}{2.9cm}%
\newcommand{\methcolimh}{1.55cm}%
\newcommand{\methodcol}[1]{%
\IfFileExists{figure/qual-acting-1/#1}
{\includegraphics[width=\methcolimw,height=\methcolimw,keepaspectratio]{#1}}{}%
}%
\newcommand{\examplerow}[2]{
\methodcol{gt\_#1\_#2.png}&%
\methodcol{maskednerf\_#1\_#2.png}&%
\methodcol{maskedtimenerf\_#1\_#2.png}&%
\methodcol{tensorf\_#1\_#2.png}&%
\methodcol{hexplain\_#1\_#2.png}&%
\methodcol{nerfies\_#1\_#2.png}%
}%

\setlength\tabcolsep{0cm}%
\renewcommand{\arraystretch}{0.0}
\centering\small%
\begin{tabular}{cccccc}
Ground Truth &%
NeRF \cite{mildenhall20nerf:} &%
NeRF$+t$ &%
TensoRF \cite{chen22tensorf:} &%
HexPlane \cite{cao23hexplane:} &%
Nerfies \cite{park21nerfies:} \vspace{0.3cm}\\%
\examplerow{\detokenize{SC-1002_DSLR-1_frame0192567}}{render}\\
\examplerow{\detokenize{SC-1002_DSLR-1_frame0192567}}{depth}\\
\examplerow{\detokenize{SC-1002_DSLR-1_frame0200767}}{render}\\
\examplerow{\detokenize{SC-1002_DSLR-1_frame0200767}}{depth}\\
\examplerow{\detokenize{SC-1018_DSLR-1_frame0156600}}{render}\\
\examplerow{\detokenize{SC-1018_DSLR-1_frame0156600}}{depth}\\
\examplerow{\detokenize{SC-1018_DSLR-1_frame0171533}}{render}\\
\examplerow{\detokenize{SC-1018_DSLR-1_frame0171533}}{depth}
\end{tabular}
\let\methcolimw\undefined%
\let\methcolimh\undefined%
\let\methodcol\undefined%
\let\examplerow\undefined%

\caption{Qualitative results on the acting (dynamic) phase of 3 different scenes. Each section contains 2 rows: rendered RGB image and rendered depth map.
Sections 1--2 and 3--4 come from the same scene; note that NeRF and TensoRF produce the same average render with ghosting artifacts because it is missing time input.
NeRFies is not trained to produce opacity masks, so we skip these renders.}\label{fig:qual-acting}
\end{figure*}

\subsection{Metrics}

We compare the methods using a range of metrics evaluating the faithfulness of the rendering, their perceptual quality, and the quality of the opacity mask (for the methods that produce it).
To this end, we use the following metrics: Peak Signal-to-Noise Ratio (PSNR), Learned Perceptual Image-Patch Similarity (LPIPS), both evaluated in the foreground region only, and the Intersection over Union (IoU) between the produced opacity mask and ground-truth foreground mask.
We compute the metrics at a fixed resolution of $960 \times 540$, which can be handled by all the methods.

\subsection{Baselines}\label{s:baselines}

We evaluate the dataset on a range of novel-view synthesis methods, including those modelling dynamic scenes.

\paragraph{NeRF and NeRF$+t$.}

Neural Radiance Field (NeRF)~\cite{mildenhall20nerf:} fitting is a cornerstone of modern novel-view synthesis.
The method learns the radiance field of the scene through an MLP $\Psi$ that predicts the colour $c_i \in \mathbb{R}^3$ and density $o_i \in \mathbb{R}$ for points $r_i$ along the rays $\mathbf{r}$ emitted from the camera center passing through each pixel:
\begin{equation}\label{eq:nerf}
    [c_i, o_i] = \Psi(\gamma_{R_x}(r_i), \gamma_{R_d}(d_i)),
\end{equation}
where $d_i$ is a normalised vector pointing from the camera centre to $r_i$, $\gamma_{R}$ is an order-$R$ harmonic encoding, $R_x, R_d \in \mathbb{N}$ are hyperparameters.
The predicted colours and opacities are then integrated along the ray using the emission-absorption raymarching function to get the final RGB value in the corresponding pixel.
Unlike standard NeRF, we do not model view-dependent colours (\ie $R_d=0$) due to sparsity of input views:
in \textit{flyaround} setting, we noticed the model generalises poorly to DSLR camera poses that are located farther away from the scene centre than the wearable camera's trajectory;
in \textit{acting} setting, we found that 5 viewpoints per timestamp are not sufficient to fit view-directional colours reliably,
i.e. we set $R_d = 0$.
Since we are interested only in the foreground, we pre-process the images by masking out background pixels.

NeRF assumes that the scene is static and produces blurry renders even in case of a limited non-deliberate motion.
Hence, we consider the temporal extension NeRF$+t$ (used in various video reconstruction methods~\cite{martin-brualla21nerf,tschernezki21neuraldiff,xian21space-time}):
\begin{equation}\label{eq:timenerf}
    [c_i, o_i] = \Psi(\gamma_{R_x}(r_i), \gamma_{R_t}(t)),
\end{equation}
where $t$ is a frame's timestamp normalised to $[0, 1]$ range.
This model is thus tasked in modelling a 4D time-space. 

\paragraph{Nerfies.} Nerfies~\cite{park21nerfies:} extend the vanilla NeRF model to handle deformations, but does this in a different way than NeRF$+t$. Instead of treating $t$ as an additional input dimension, Nerfies model the dynamics by considering a time-invariant rigid radiance field $\Psi$ in \emph{canonical space}, and a time-dependant deformation field $\Delta$, which allows to convert points from posed to canonical space. Therefore, in order to compute the color of a pixel of an image $I_k$, the points $r_i$ along the ray $\mathbf{r}$ are first offset to canonical coordinates by applying $\Delta$, before obtaining the colors $c_i$ and opacities $o_i$ by evaluating the implicit function $\Psi$:
\begin{equation}\label{eq:nerfies}
\begin{split}
    \bar{r}_i &= \Delta(\gamma_{R_x}(r_i), \phi_k) \\
    [c_i, o_i] &= \Psi(\gamma_{R_x}(\bar{r}_i), \gamma_{R_d}(d_i)),
\end{split}
\end{equation}
where $\phi_k$ is an appearance code corresponding to image $I_k$. We found that the quality of Nerfies degrades with masking, so we train it on the unmasked videos. Note that we still report the foreground-only PSNR and LPIPS.


\paragraph{TensoRF and HexPlane.} 
TensoRF~\cite{chen22tensorf:} shares with NeRF the underlying idea of volumetric rendering through emission-absortion ray marching. However, instead of modelling the radiance field with an MLP, TensoRF proposes to model it through a product of a set of 2D ($M^{XY}$, $M^{YZ}$, $M^{XZ}$) and 1D tensor components ($v^{X}$, $v^{Y}$, $v^{Z}$), which factorize the 3D density and color spatial fields in a memory- and computationally-efficient manner.

HexPlane~\cite{cao23hexplane:} extends TensoRF by considering the factorization of the 4D space-time density and color fields (analogousy to how NeRF$+t$ extends NeRF), and therefore including into the factorization 2D tensors $M^{Xt}$, $M^{Yt}$, $M^{Zt}$ which span the temporal axis $t$ along with each spatial axis $x$, $y$, or $z$.

\subsection{Results}%
\label{s:results}

\paragraph{Flyaround.}
The result on a flyaround benchmark are shown in Figure \ref{fig:qual-flyaround} and summarised in the left half of Table \ref{tab:results}. NeRF and NeRFies produce best results, with the latter better adapting to small movements present in the scene. While TensoRF shows results comparable to NeRF, its temporal extension, HexPlane, falls short to generalise to a wide range of viewing angles, presumably due to overfitting to the additional temporal dimension.

\paragraph{Acting.}
The results on acting benchmark are shown in Figure \ref{fig:qual-acting} and summarised in the right half of Table \ref{tab:results}. Here, time-extension of NeRF shows the best quality, being able to learn the dynamic geometry to a better extent, while time-agnostic NeRF produces the ghost-looking shapes wherever an actor changed the pose. HexPlane, on the other hand, in spite of a better ability to model deformations, does not improve much over TensoRF, and NeRFies fails to reconstruct the geometry explaining the errors away with floaters.

\section{Conclusion}%
\label{s:conclusions}

We presented \datasetlong, a collection of scenes captured with egocentric and scene-static sensors.
We aim primarily to support research in new-view synthesis of dynamic and multi-modal content from egocentric sensors, including in particular reconstruction from a single viewpoint or a very narrow baseline.
In the future, this technology will enable breakthrough applications such as personal holography.
While this task is still too challenging for existing new-view synthesis methods, as generative AI matures, it will become possible to better hallucinate information missing in the capture, and \dataset can spur further research in this direction.
Furthermore, while in this paper we have only discussed the visual component of the data, \dataset also contains fully-calibrated and synchronised audio information for research in multi-modal new-view synthesis, which is as of today largely unexplored.







\paragraph{Limitation.} There are a few limitations of the released part of the dataset. First, we did not record stereo videos, which might become an important modality with the next generation of wearable devices. Second, the operators of dynamic cameras do not change location during acting, only moving their heads in a natural way, which represents a subset of AR/VR applications. Finally, all released scenes were filmed in the same room, albeit with various furniture and props. These limitations may be addresses in future work.

{\small\bibliographystyle{ieee_fullname}\bibliography{vedaldi_general,vedaldi_specific,local}}

\iccvfinalcopy
\def\iccvPaperID{6874}
\ificcvfinal\pagestyle{empty}\fi

\hfuzz=5.002pt
\hbadness=10000





\clearpage
\appendix
\ificcvfinal\thispagestyle{empty}\fi

\section{Scene Variety}

\begin{figure*}[th]
\centering%
\graphicspath{{figure/all_scenes/}}

\newcommand{\imw}{4.5cm}%
\newcommand{\sceneimage}[1]{%
\IfFileExists{figure/all_scenes/#1}
{\includegraphics[width=\imw,keepaspectratio]{#1}}{}%
}%

\setlength\tabcolsep{0cm}%
\renewcommand{\arraystretch}{0.5}
\centering
\begin{tabular}{c}

\sceneimage{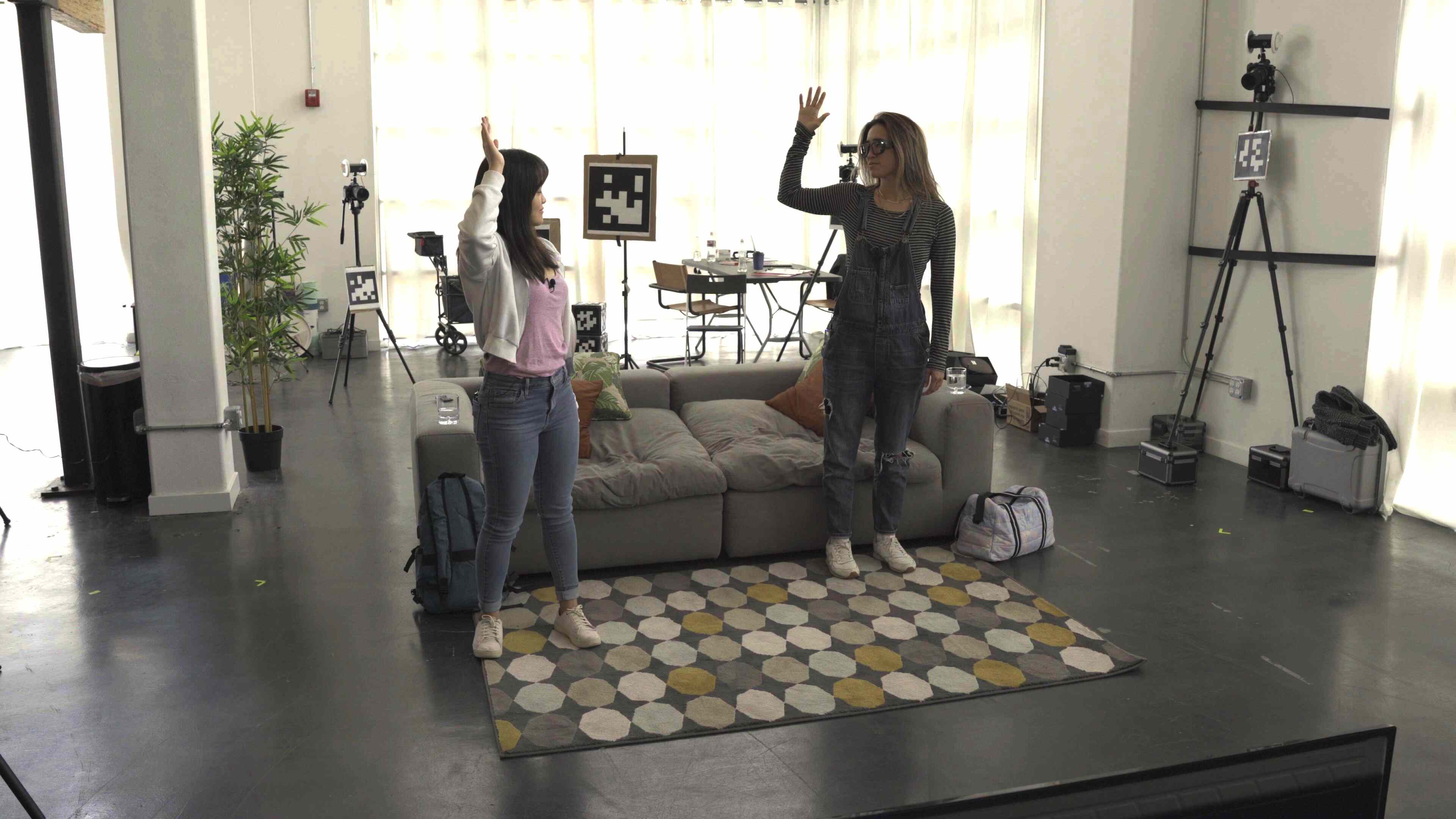} 
\sceneimage{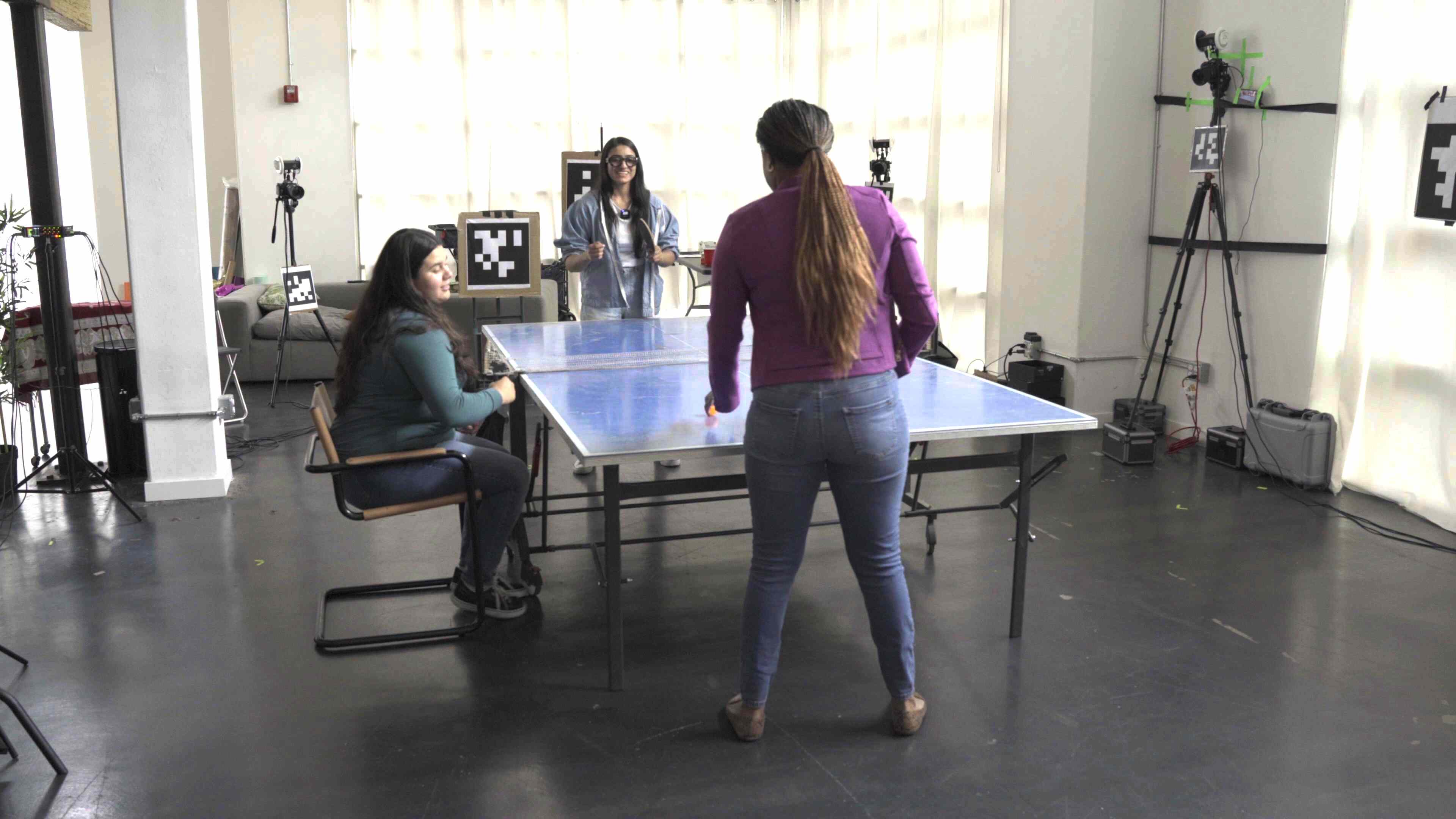} 
\sceneimage{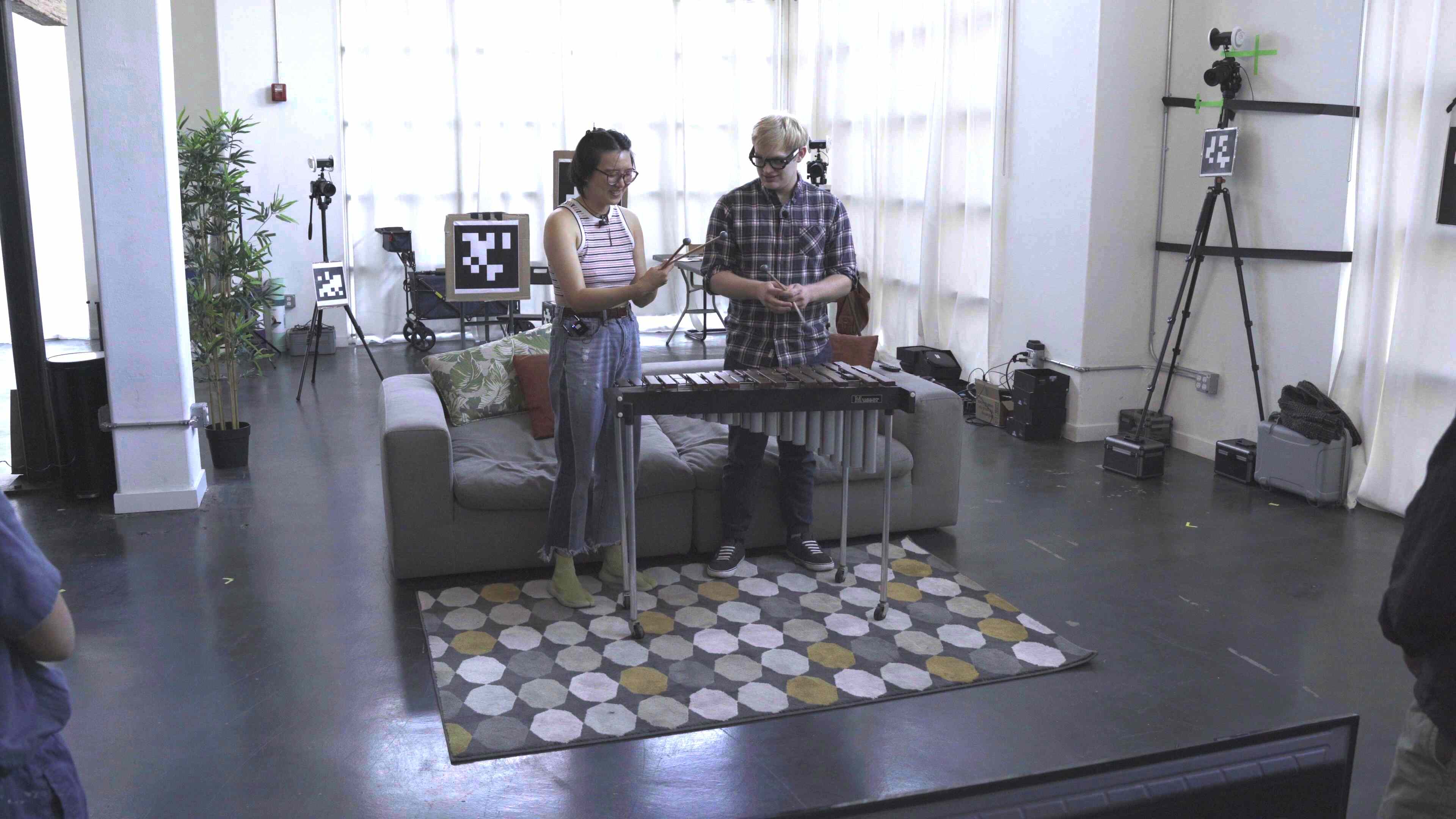} 
\sceneimage{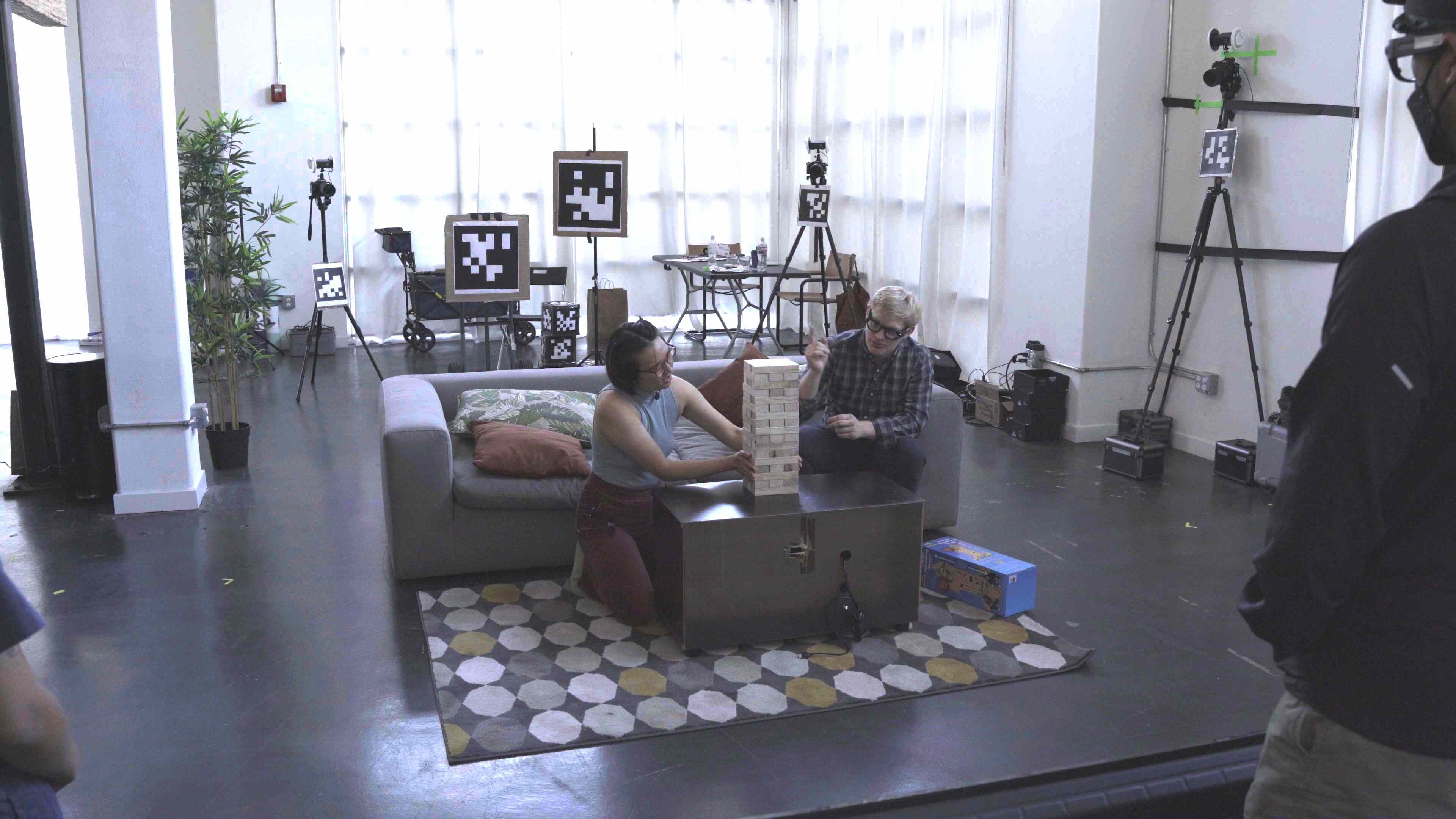}
\\
\sceneimage{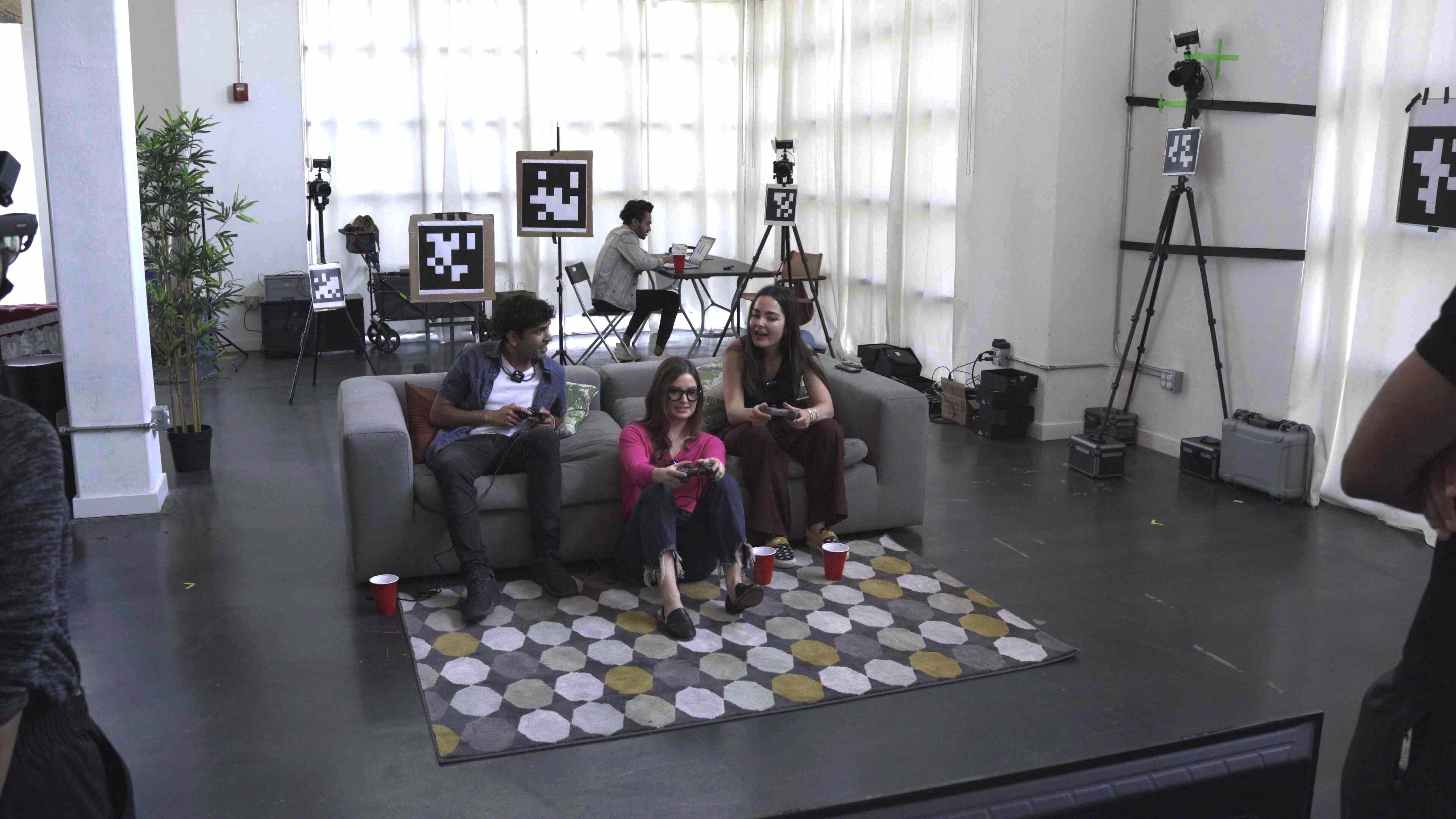} %
\sceneimage{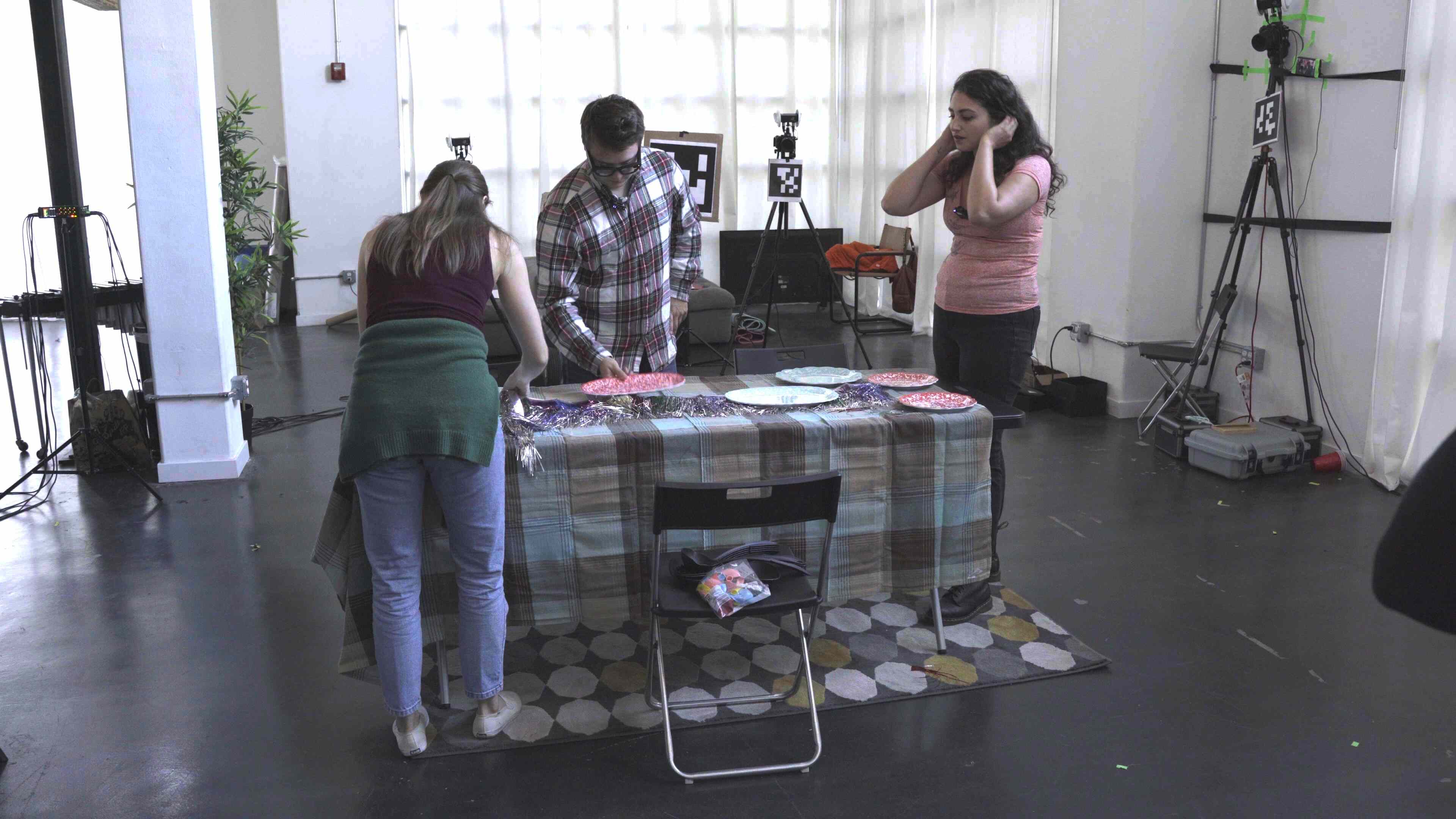} 
\sceneimage{1018_DSLR-1_frame0156000.jpg} 
\sceneimage{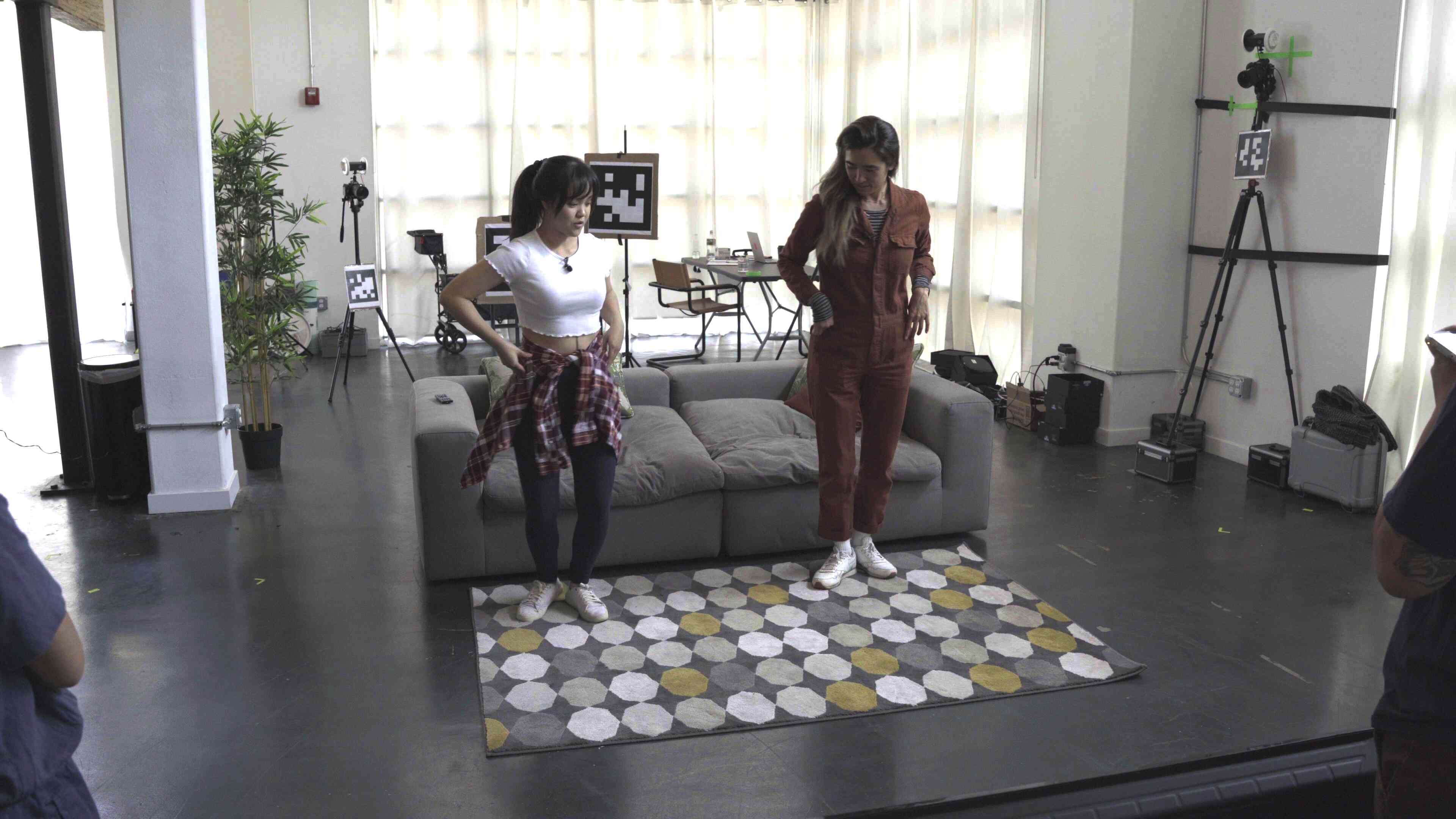}
\\
\sceneimage{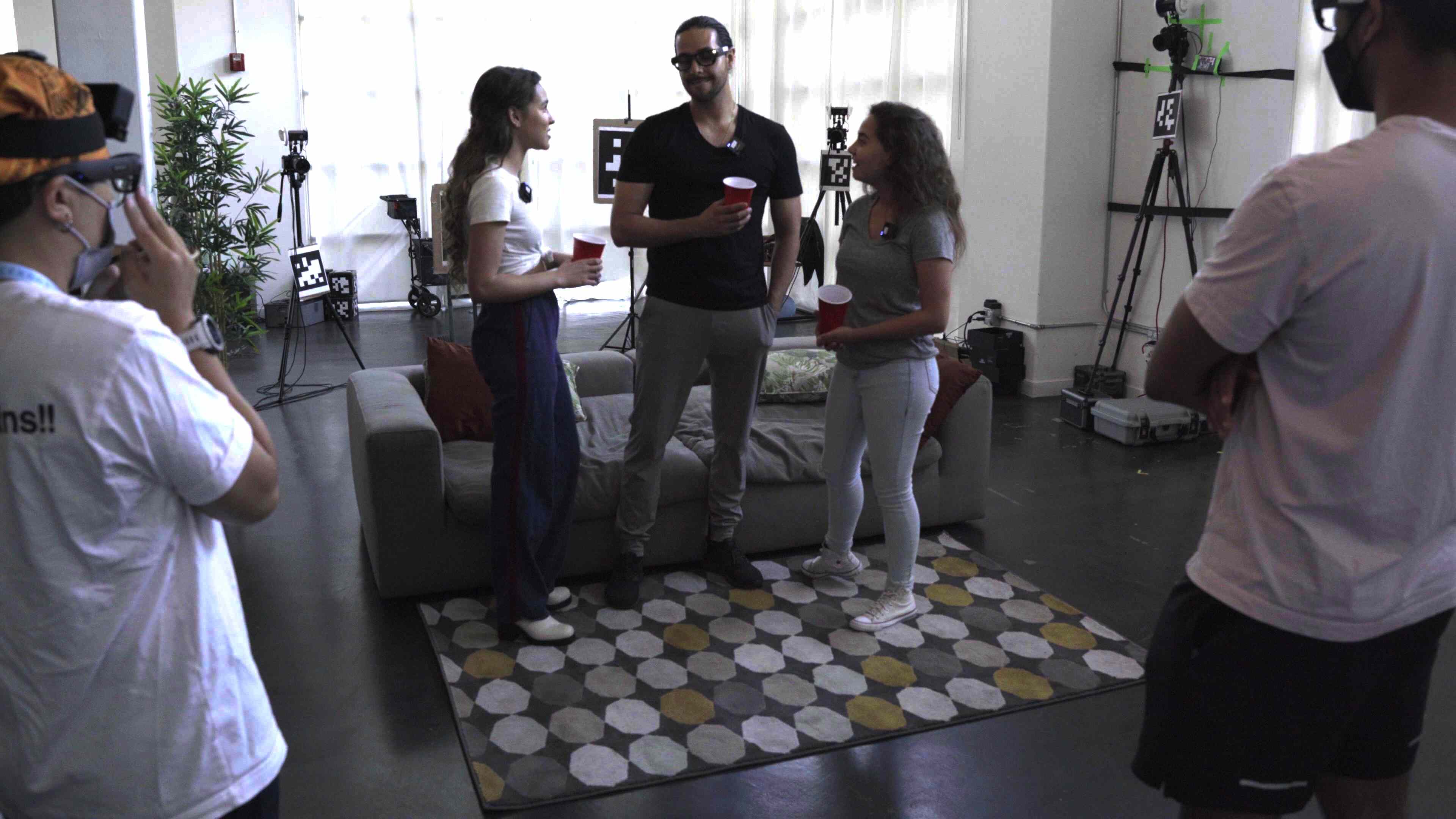} 
\sceneimage{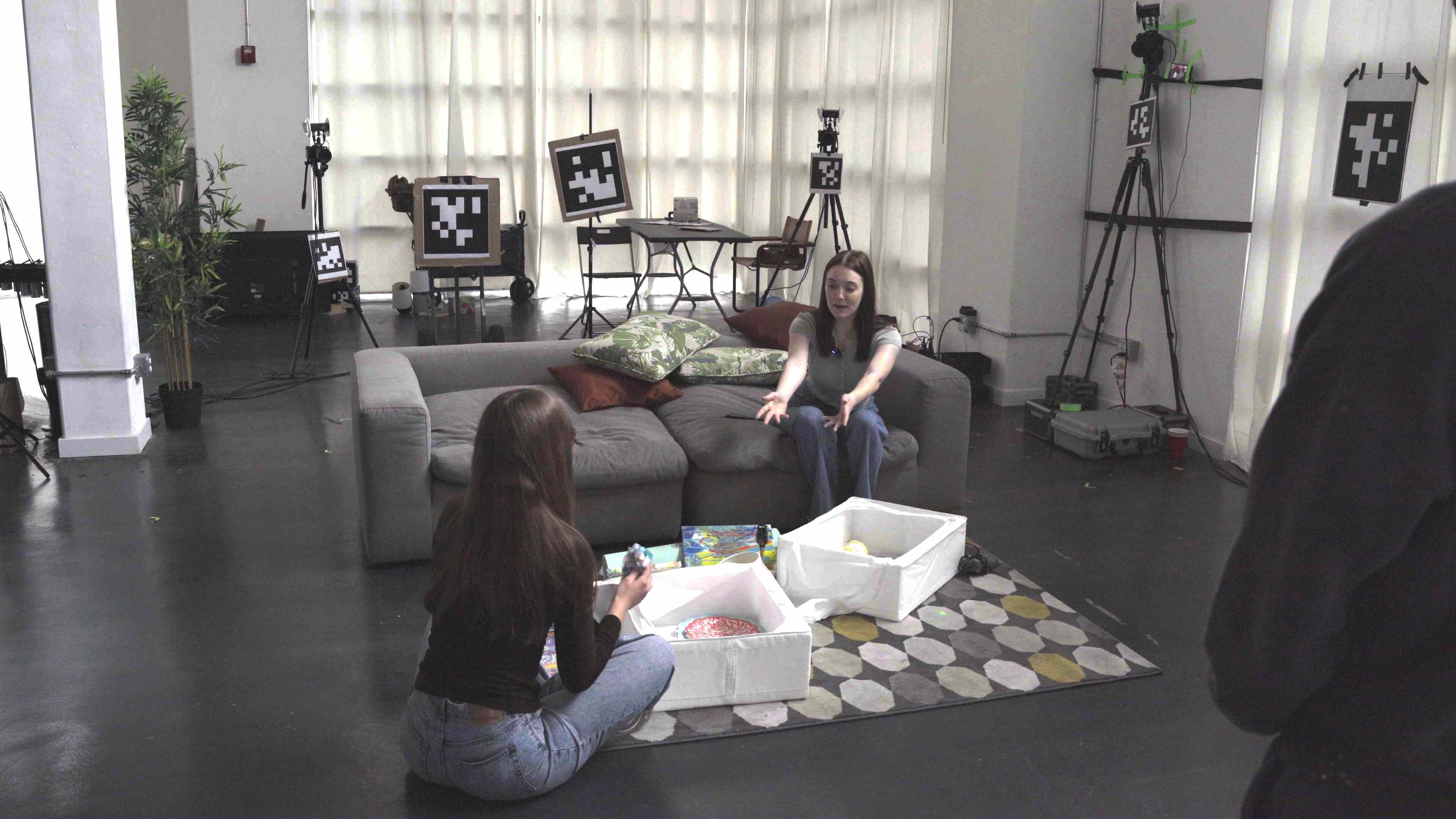} %
\sceneimage{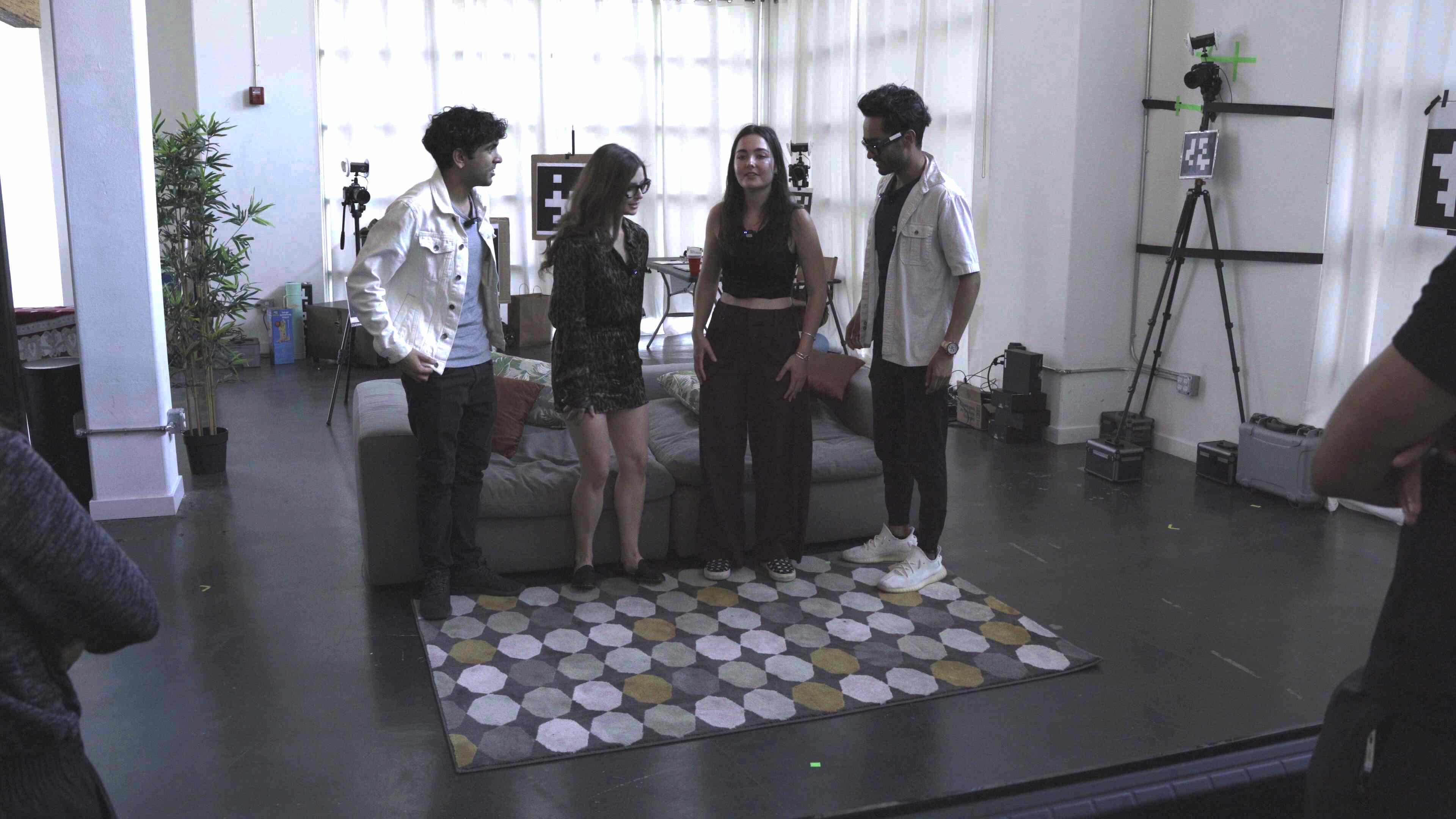} 
\sceneimage{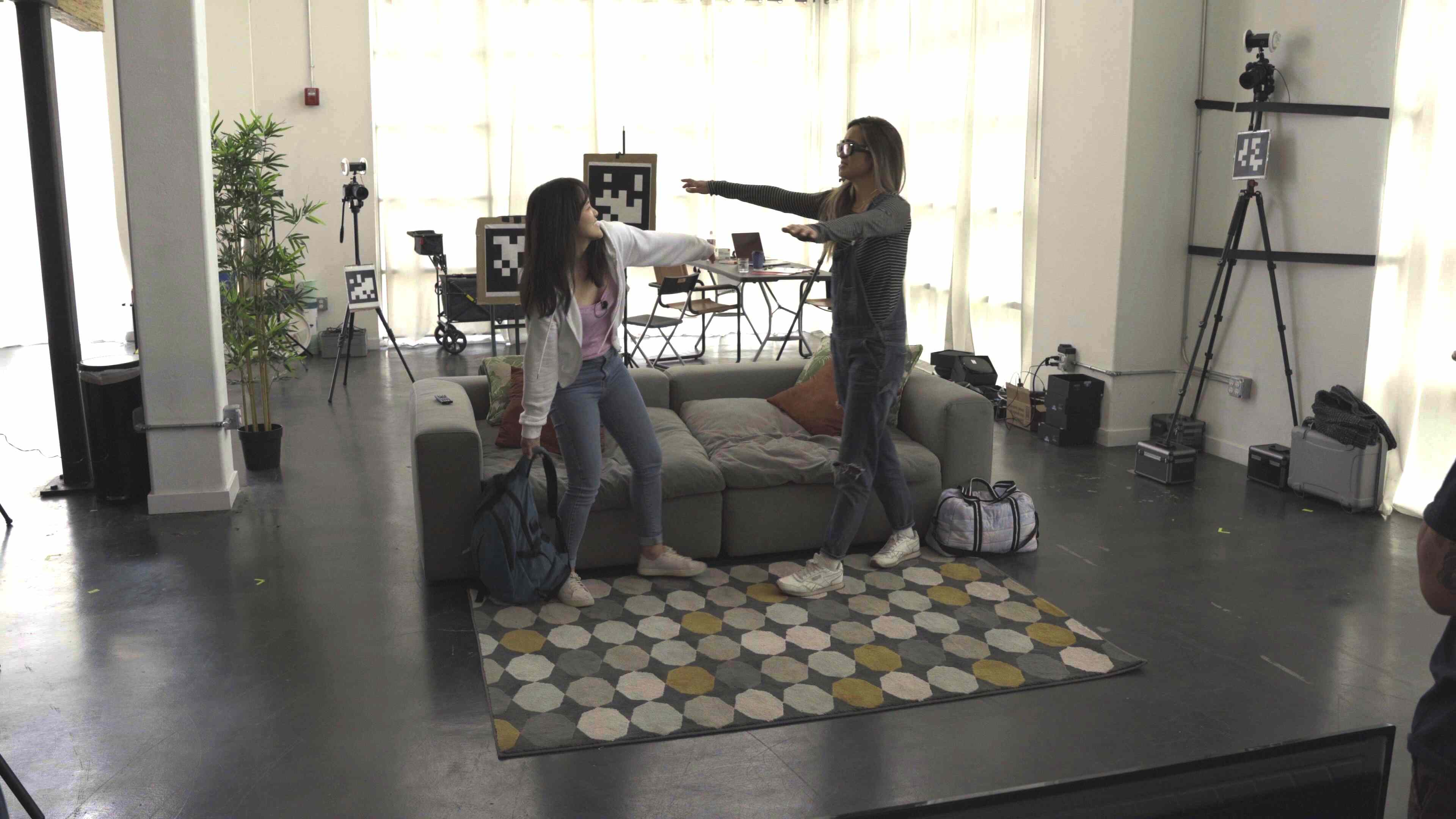}
\\
\sceneimage{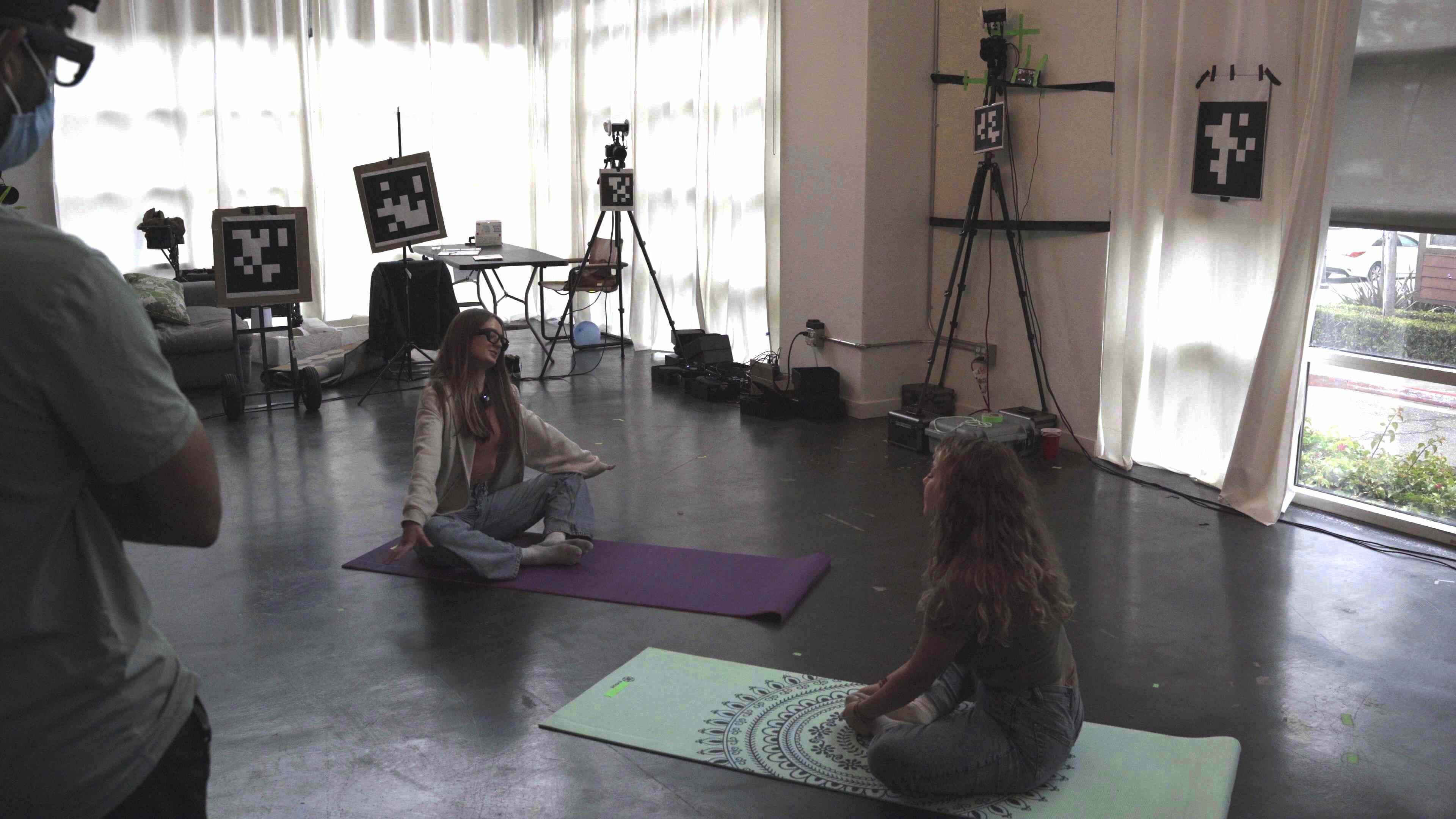} 
\sceneimage{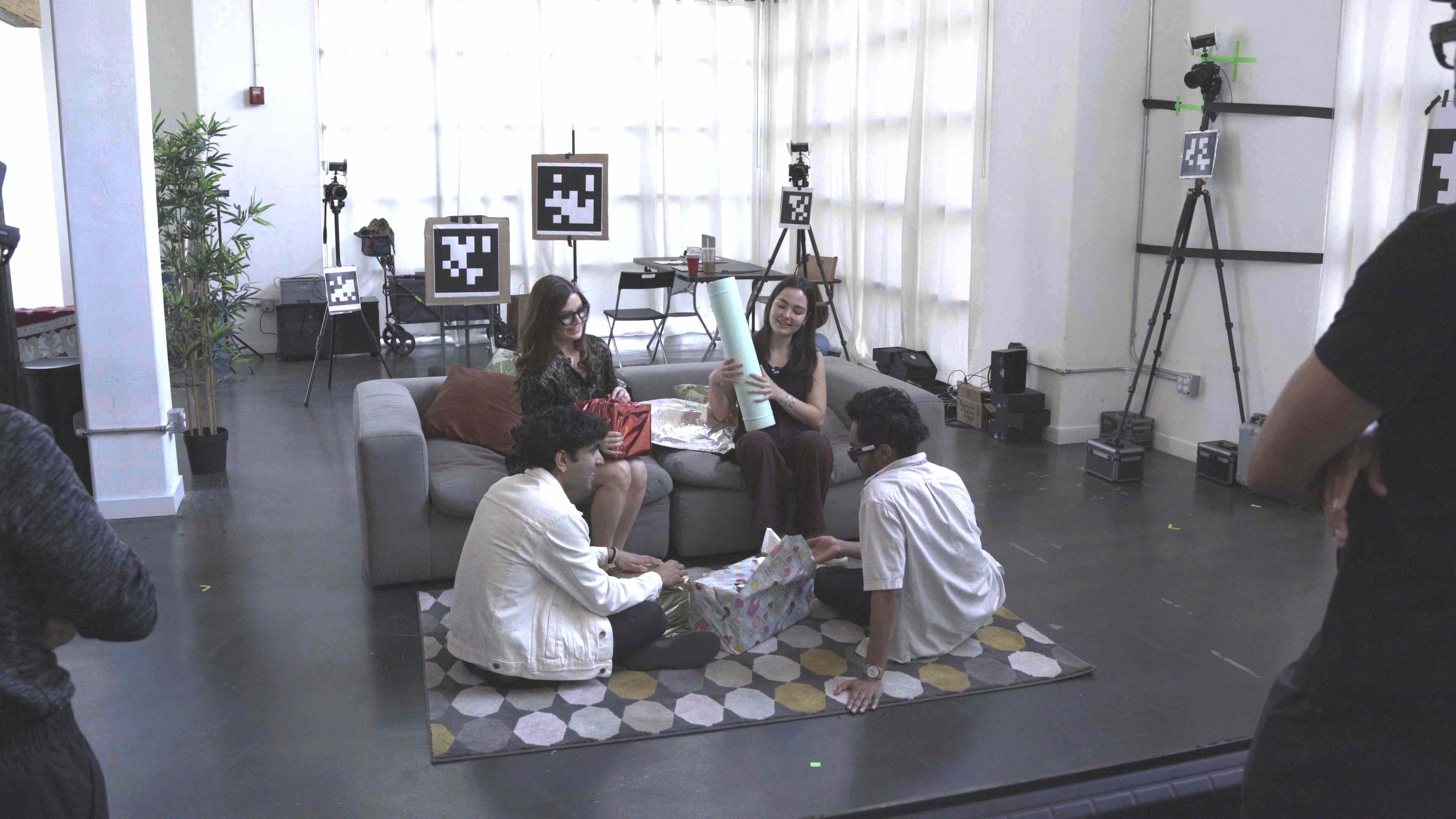} 
\sceneimage{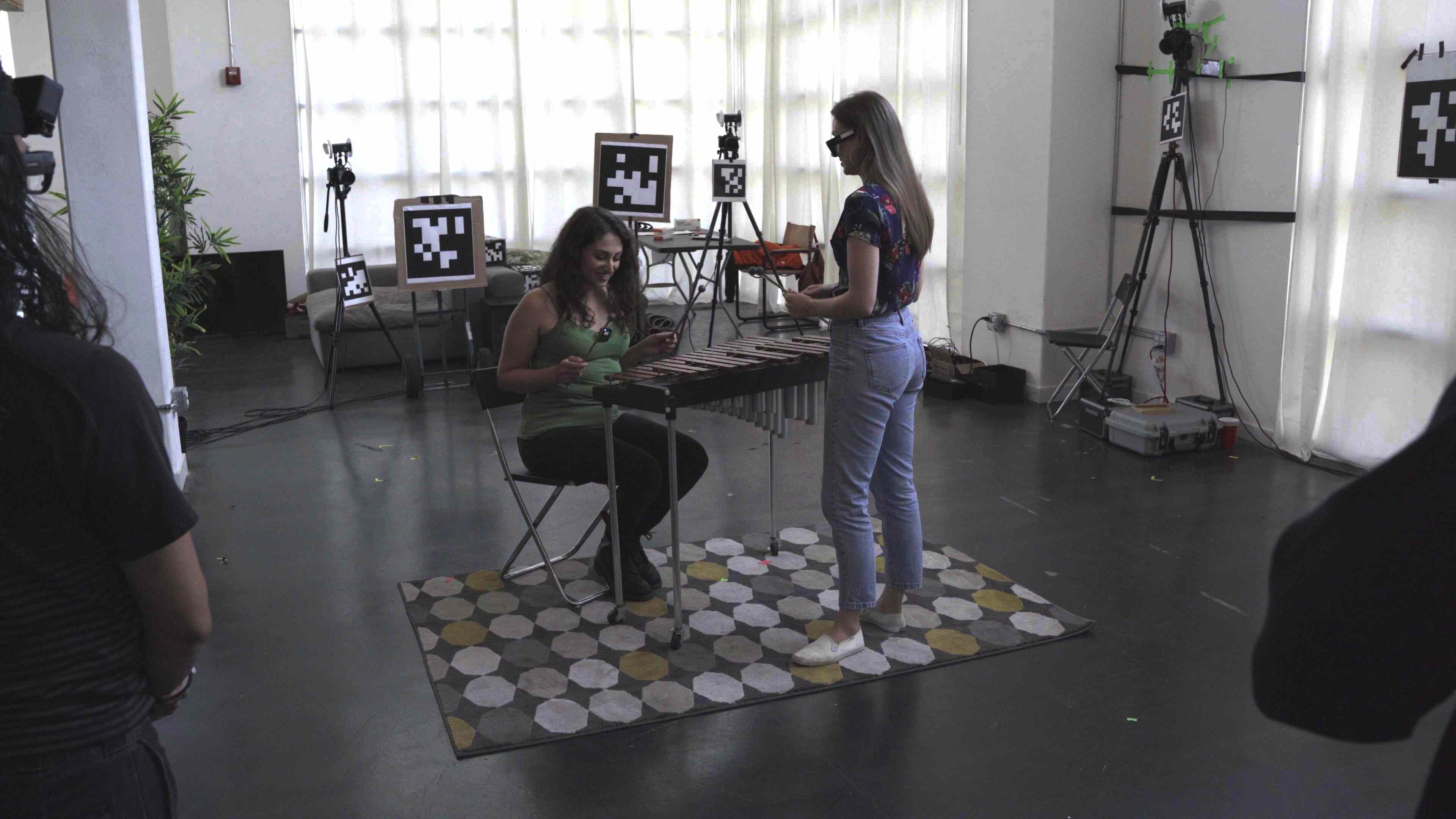} %
\sceneimage{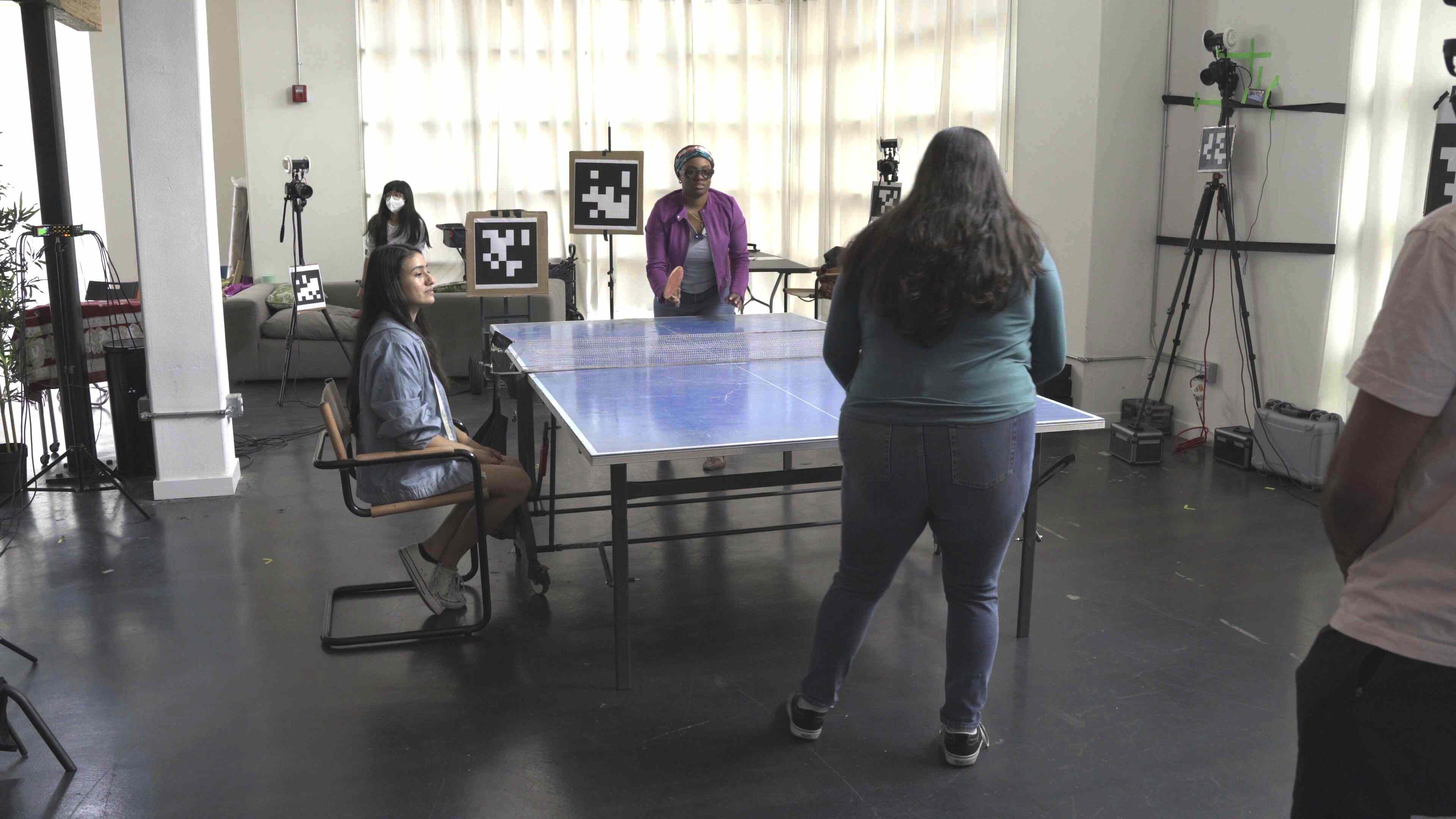}
\\
\sceneimage{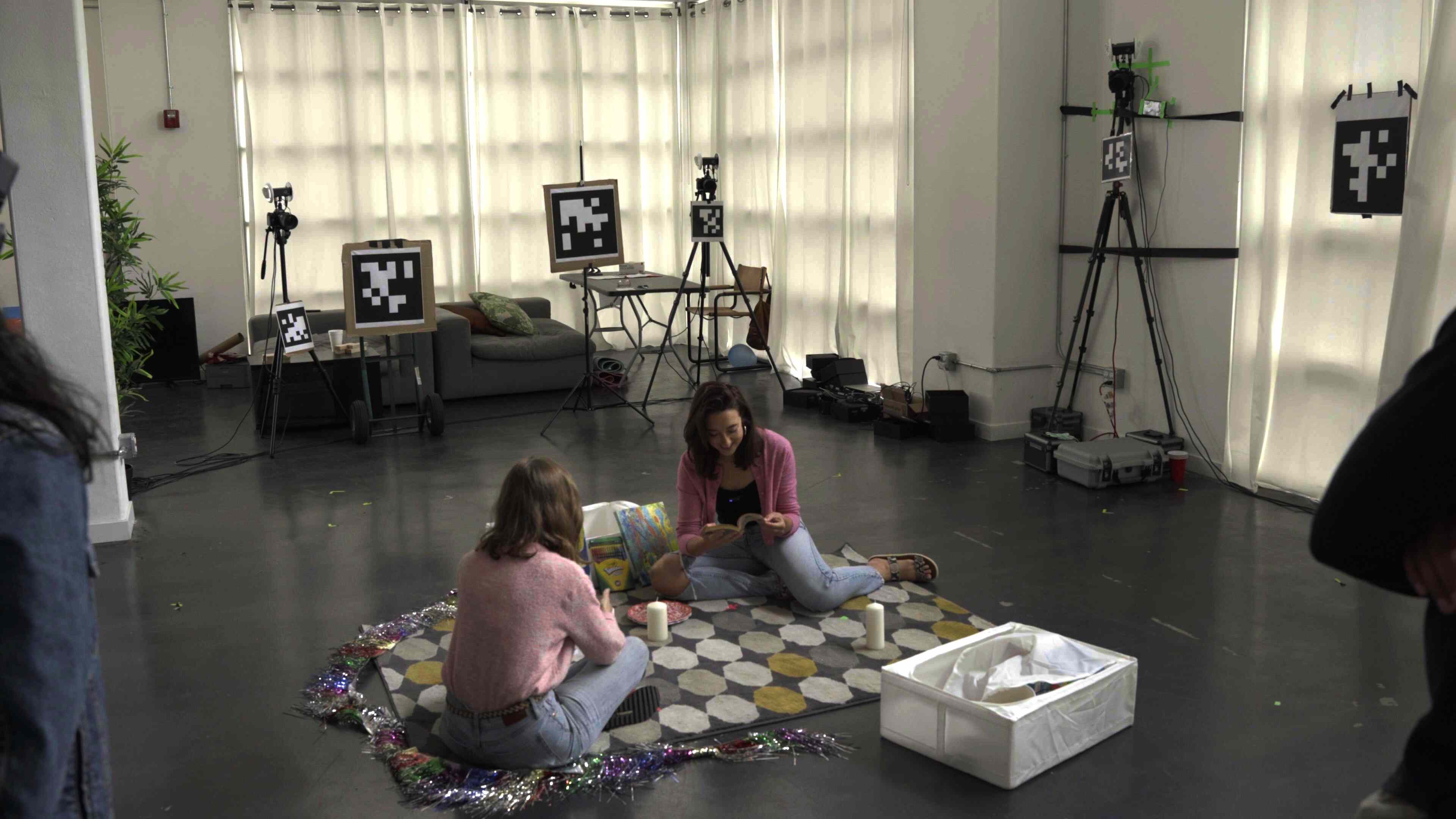} 
\sceneimage{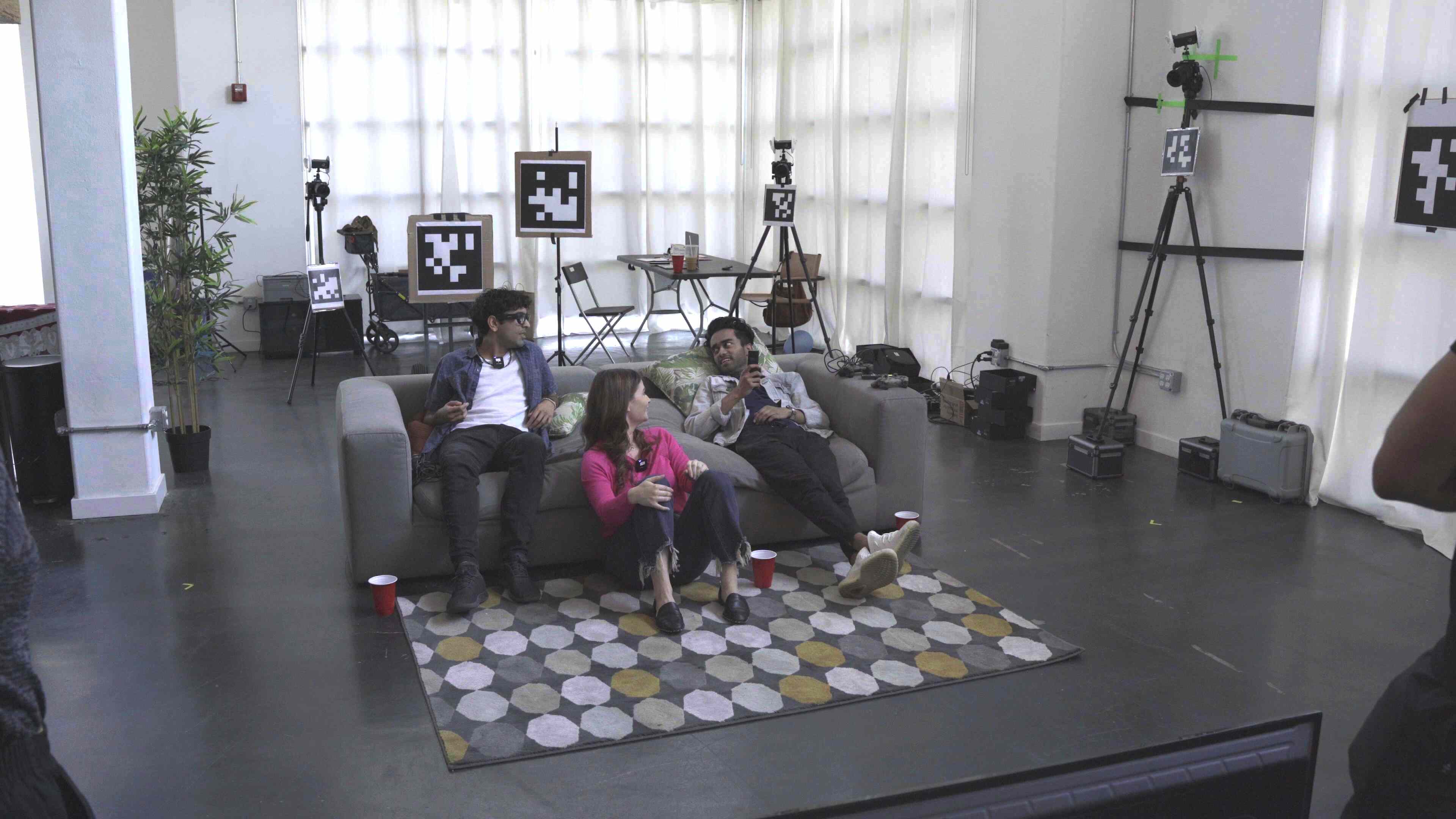} 
\sceneimage{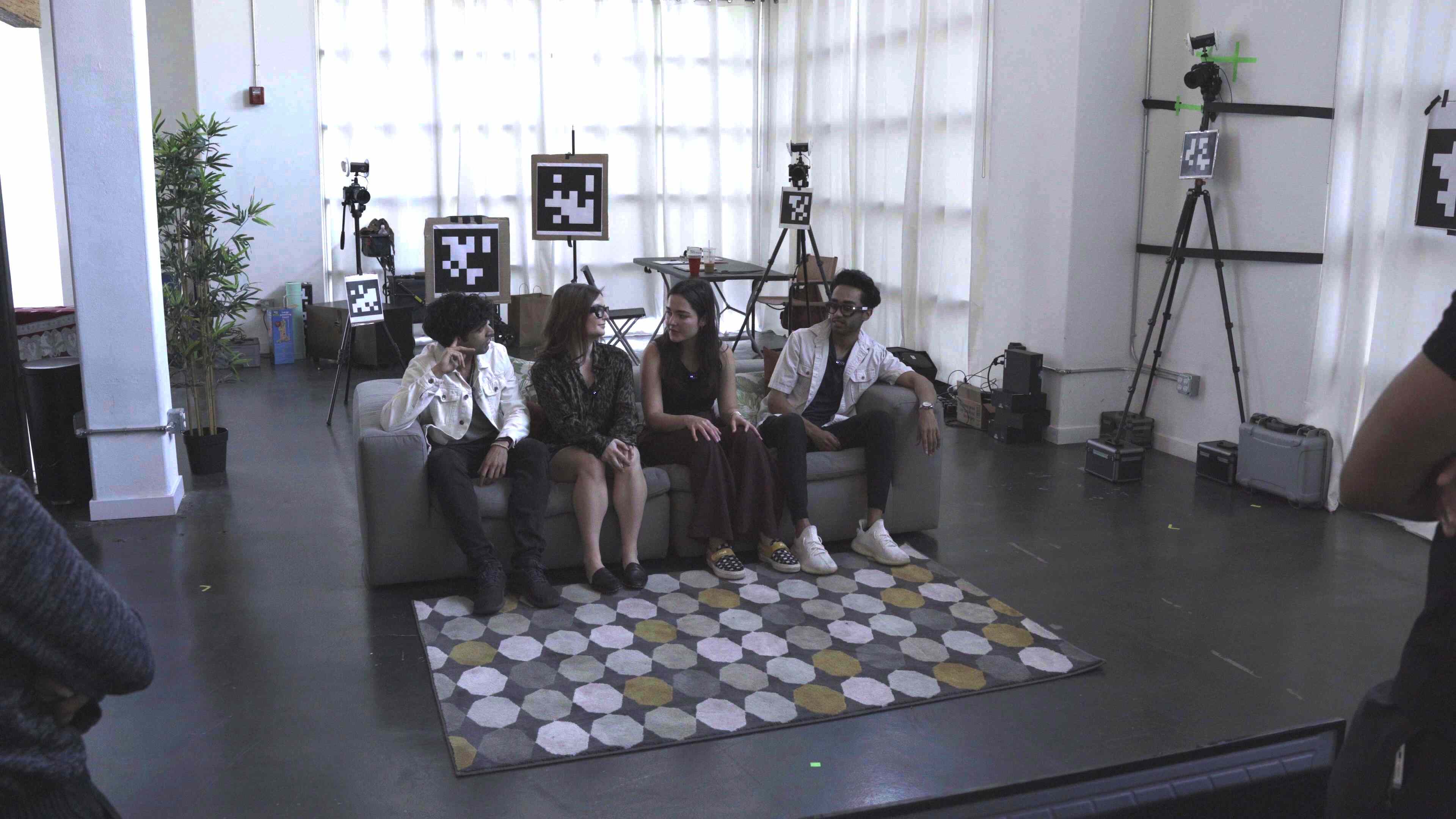} 
\sceneimage{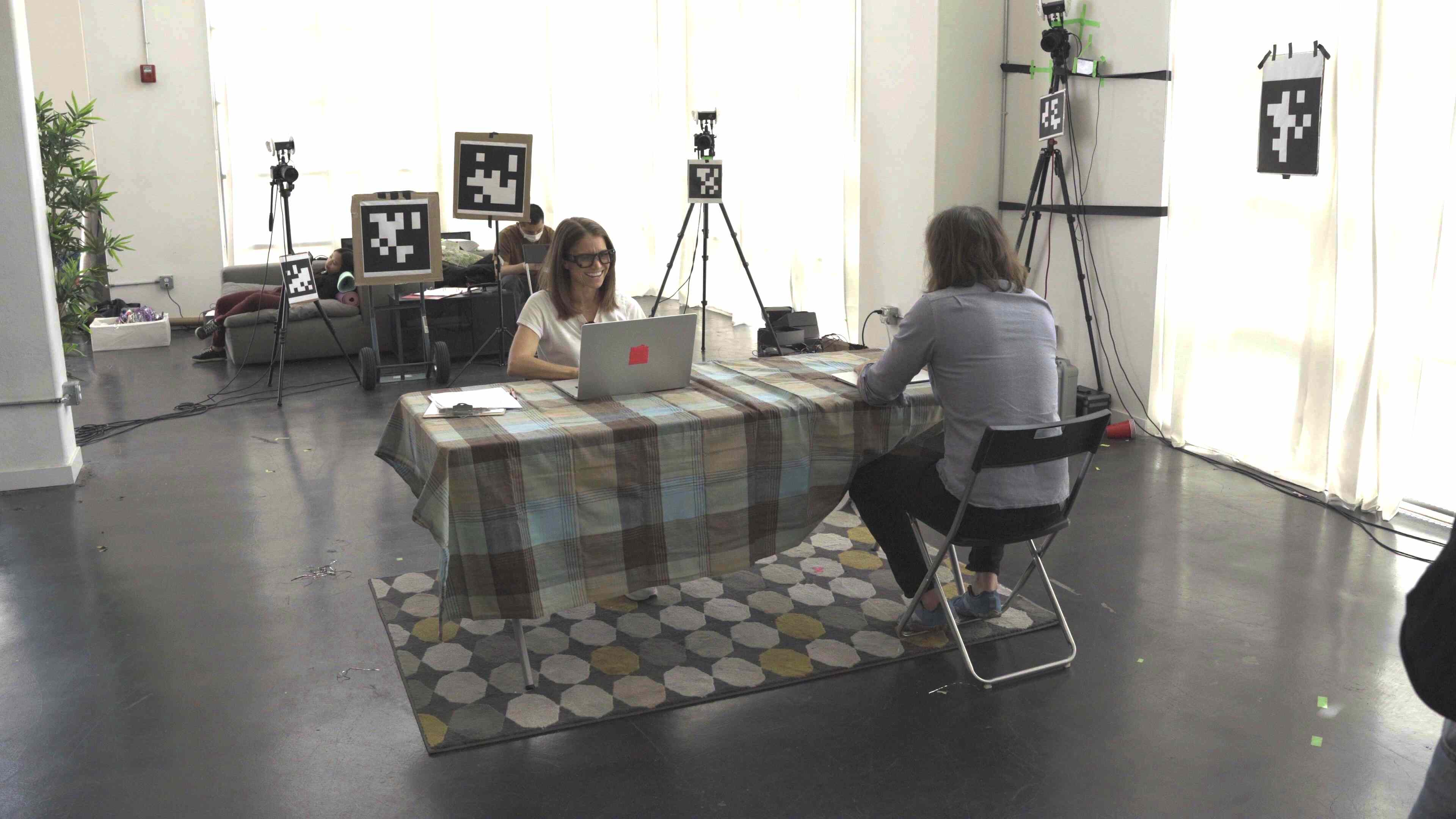}
\\
\sceneimage{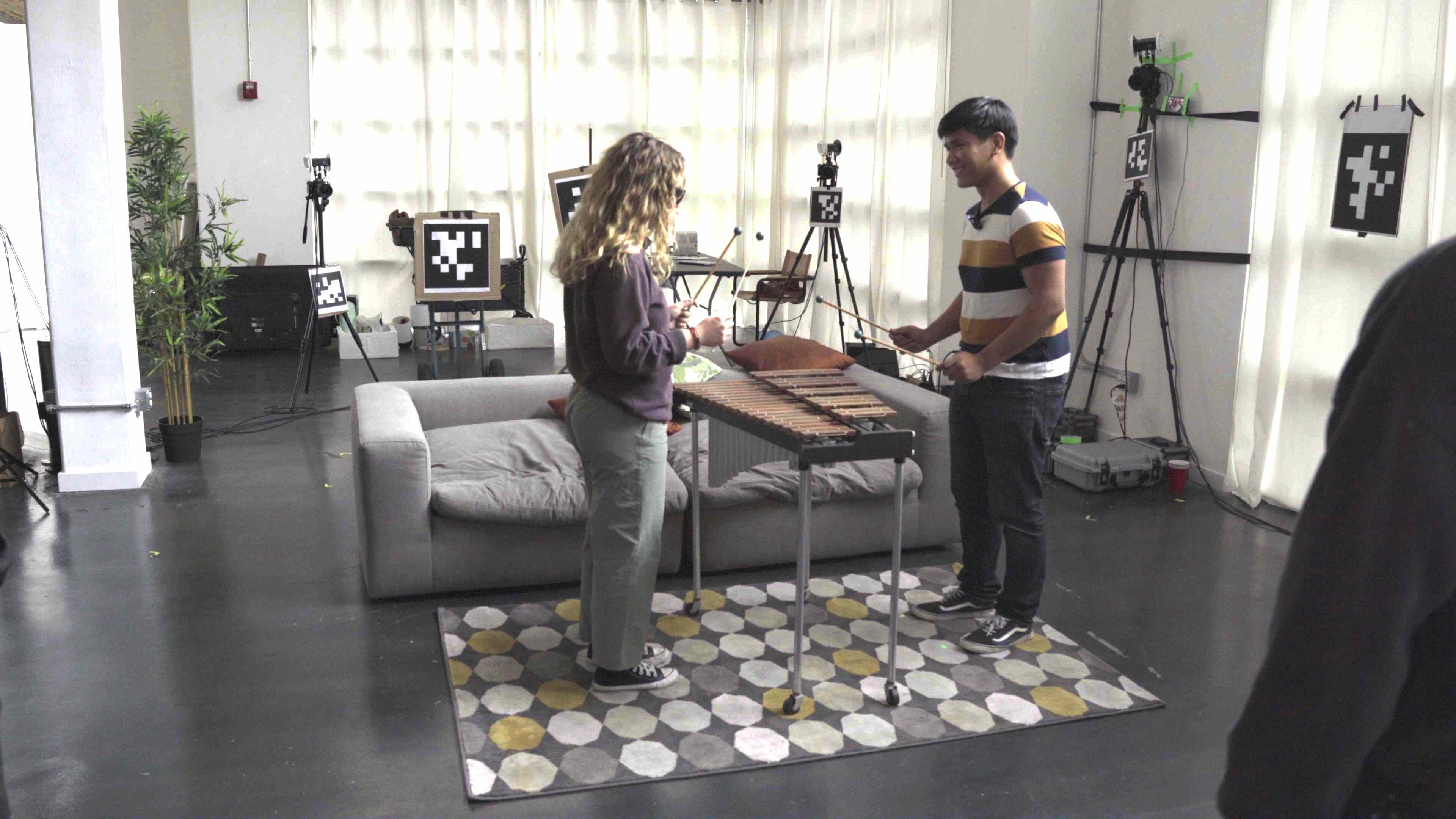} 
\sceneimage{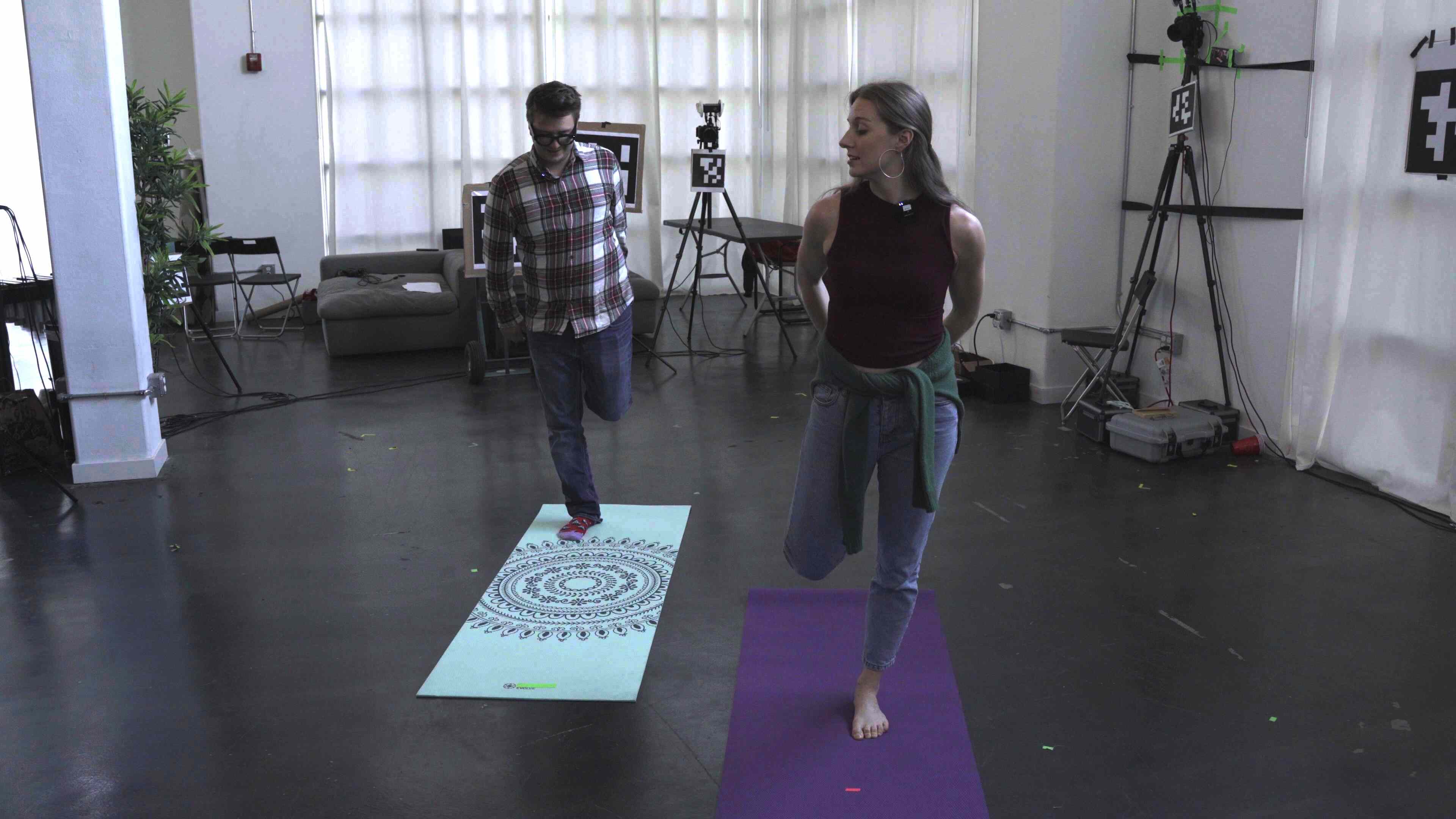} 
\sceneimage{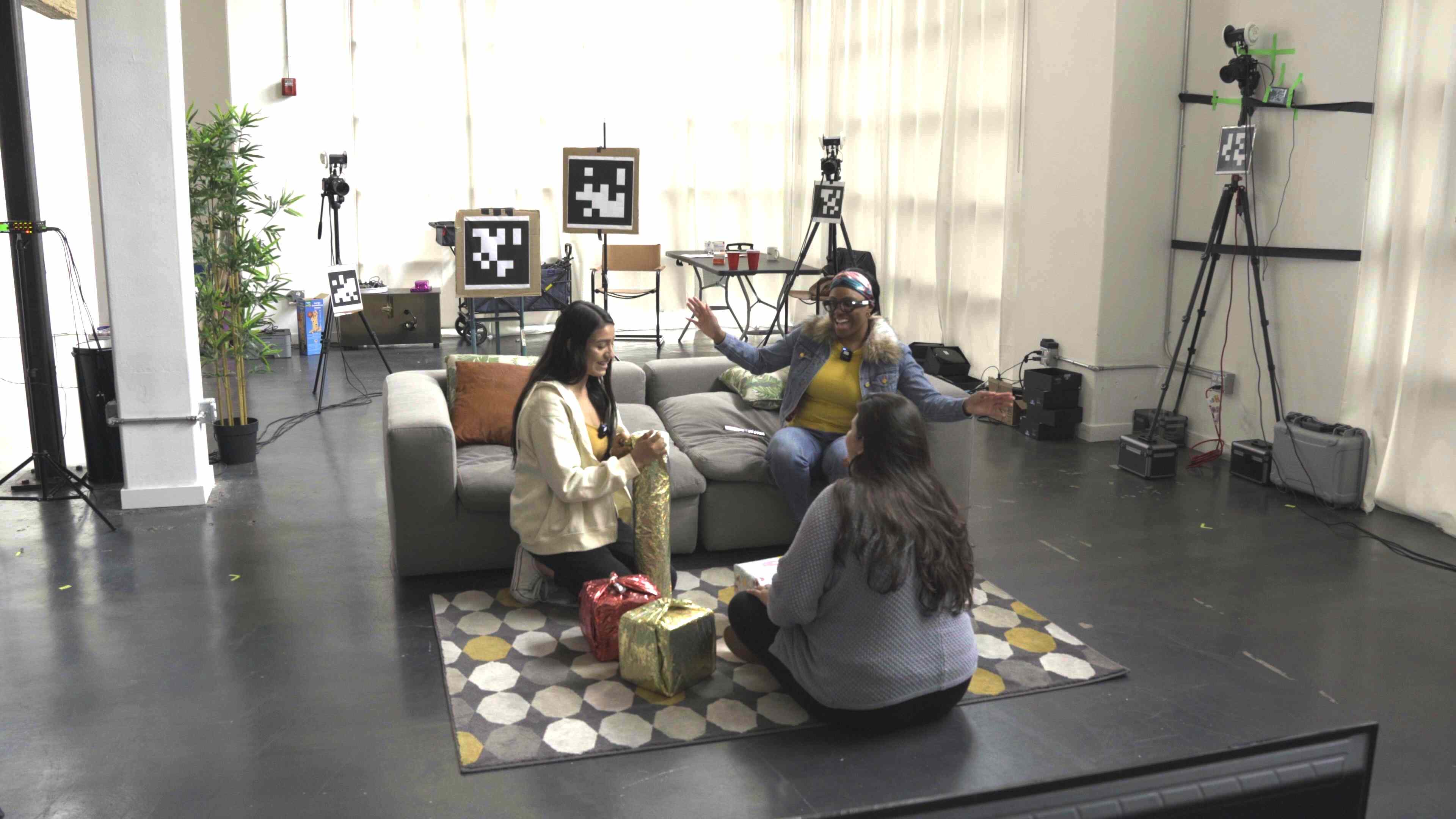} 
\sceneimage{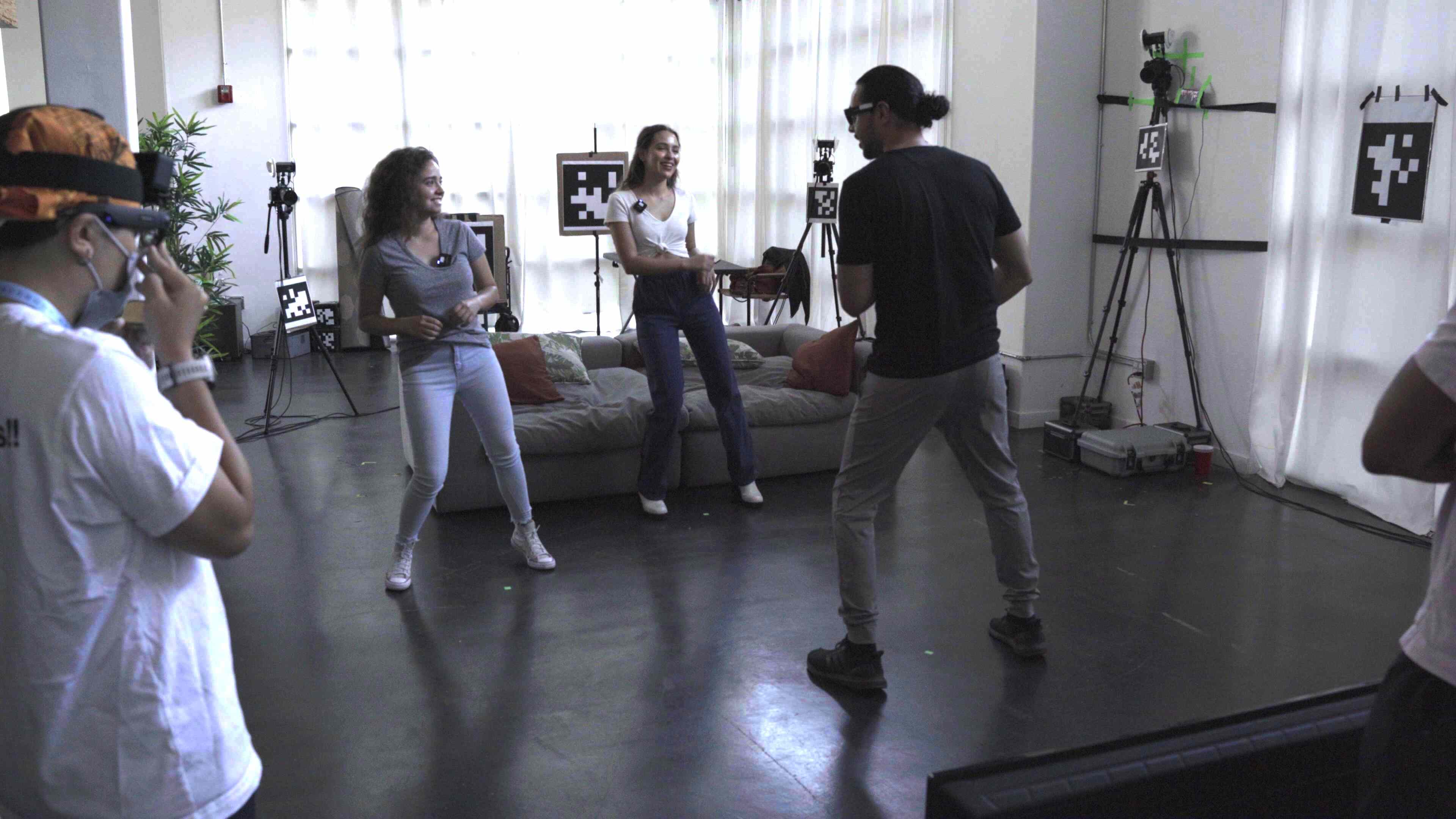}
\\
\sceneimage{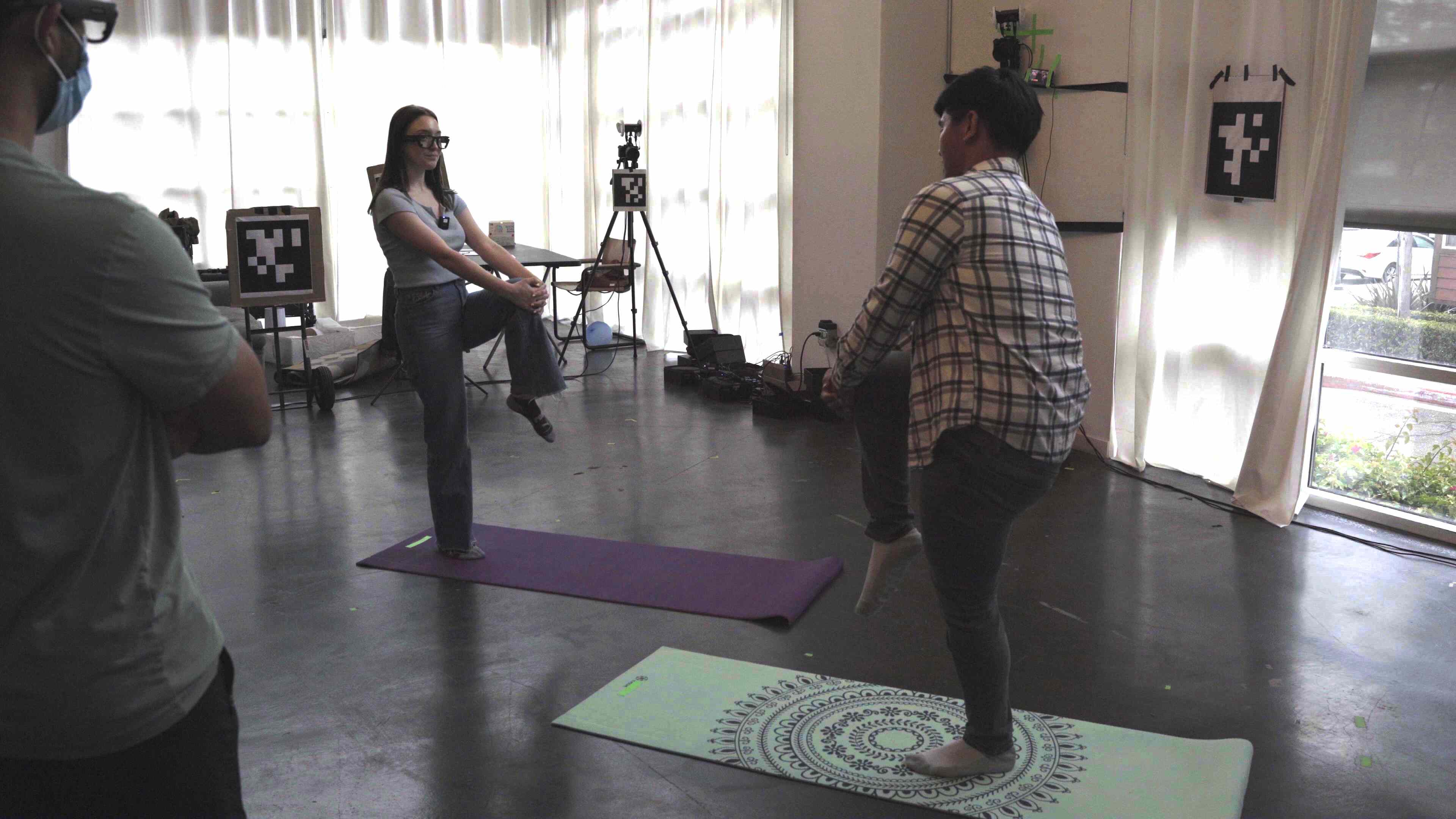} %
\sceneimage{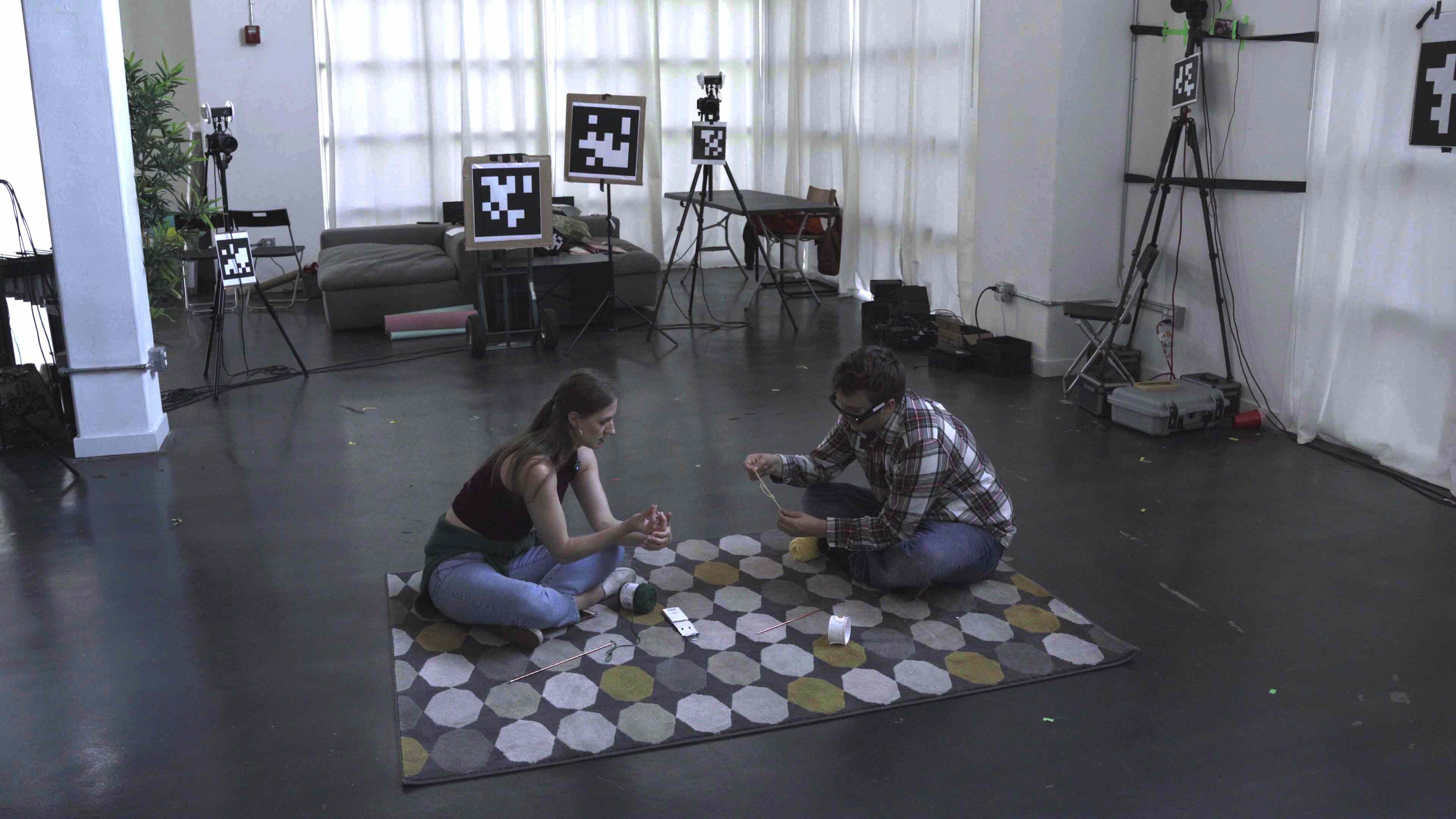} 
\sceneimage{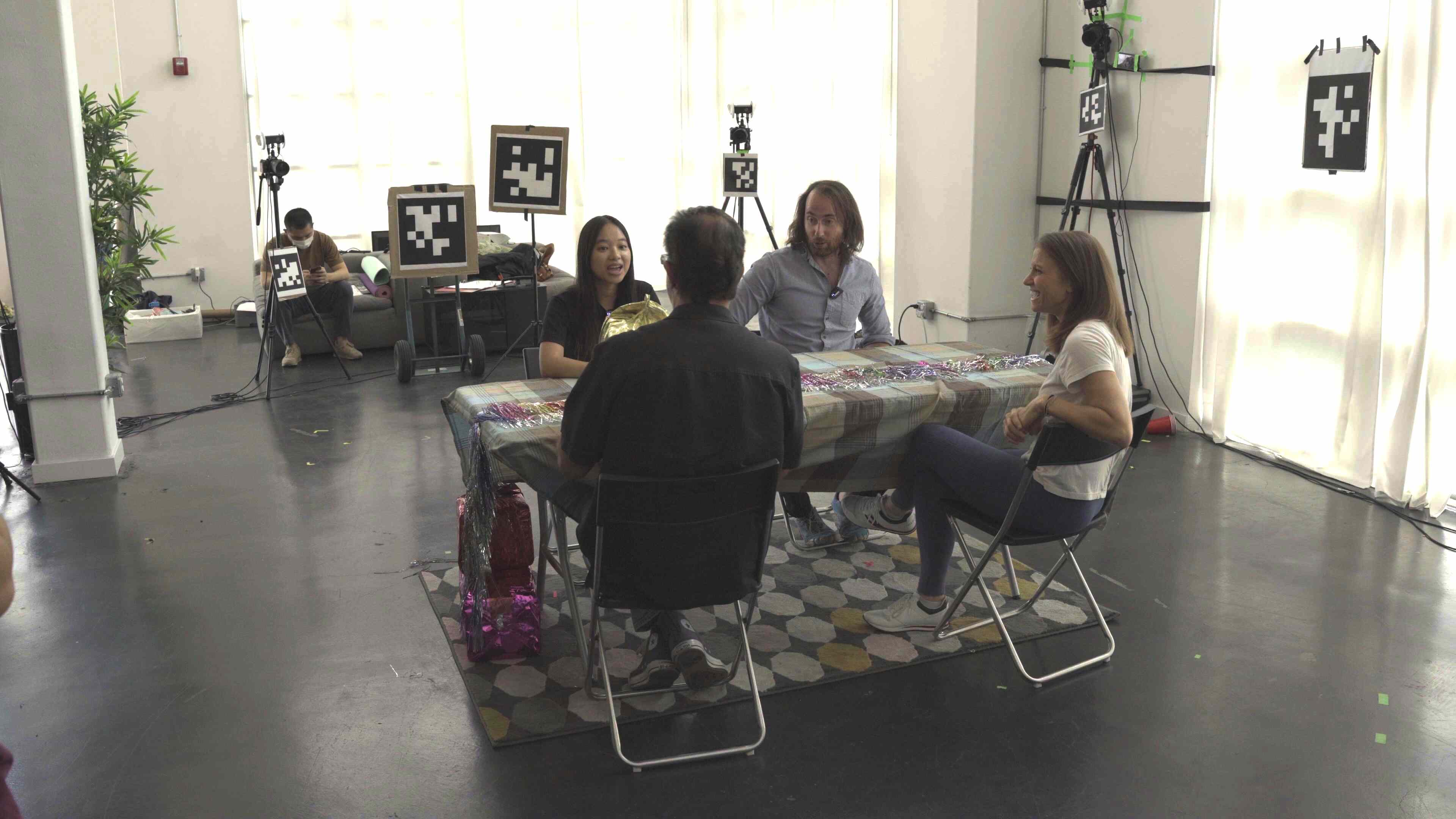} 
\sceneimage{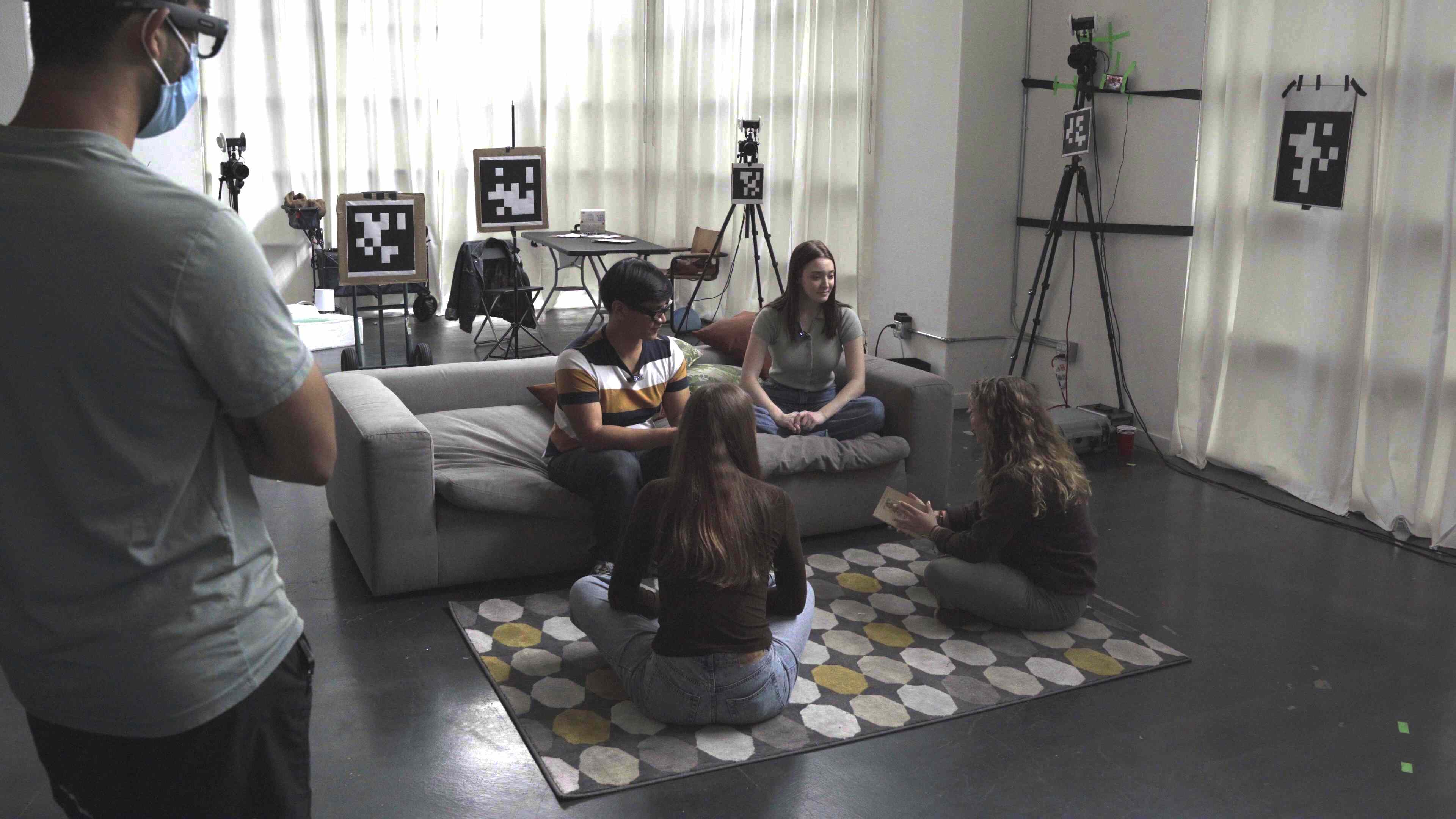}%

\end{tabular}

\let\sceneimage\undefined%
\let\imw\undefined%

\caption{\textbf{Scene diversity.} Representative frames from \textbf{28 different scenes} in the Replay dataset.
Scenes are individually synchronised and calibrated.}\label{fig:all_scenes}
\end{figure*}

\begin{figure*}[th]
\graphicspath{{figure/all_sensors/}}

\newcommand{\imh}{2.5cm}%
\newcommand{\sceneimage}[1]{%
\IfFileExists{figure/all_sensors/#1}
{\includegraphics[height=\imh,keepaspectratio]{#1}}{}%
}%

\setlength\tabcolsep{0cm}%
\renewcommand{\arraystretch}{0.5}
\centering
\begin{tabular}{c}

\sceneimage{1018_DSLR-1_frame0156000.jpg} 
\sceneimage{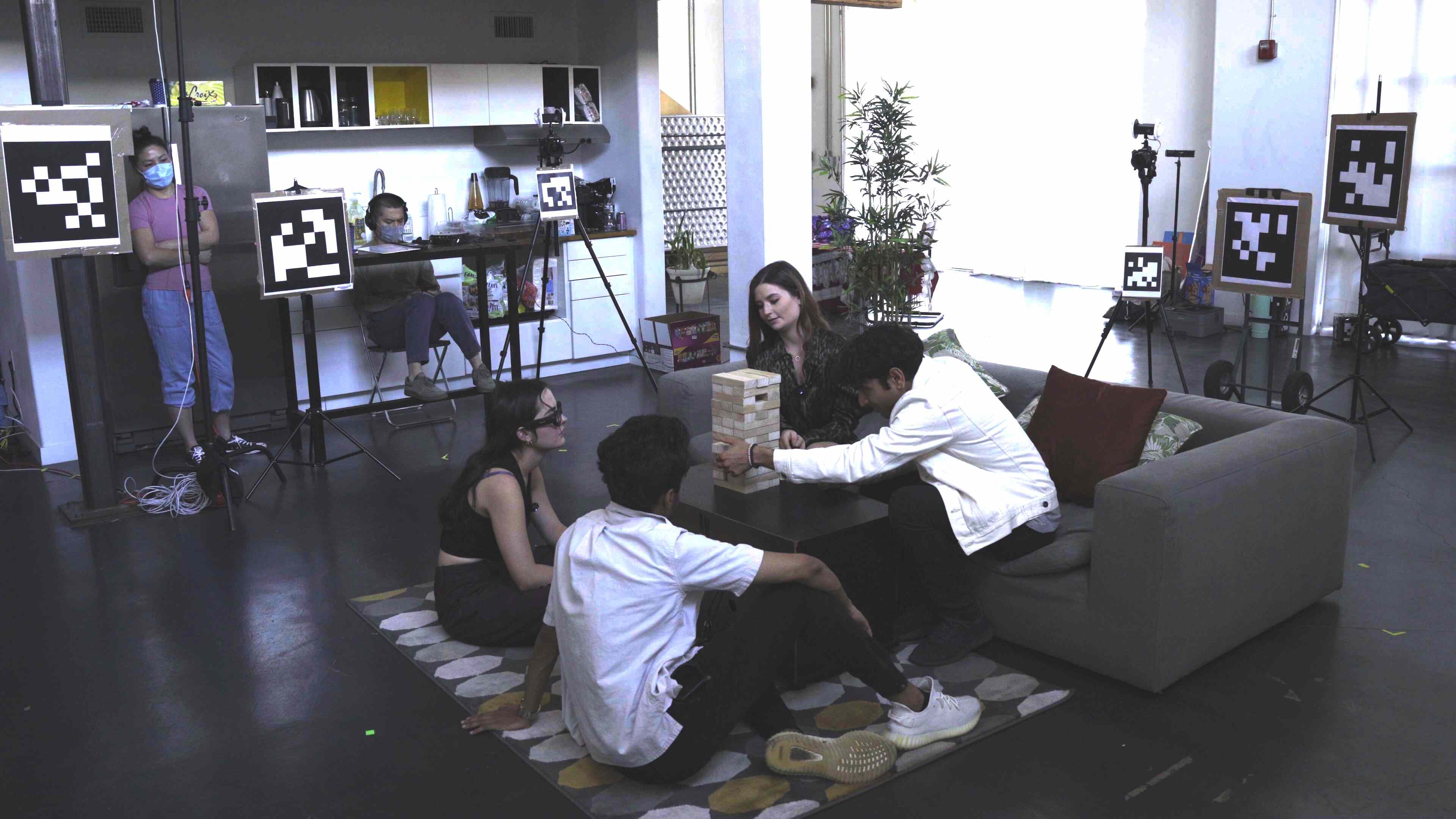} 
\sceneimage{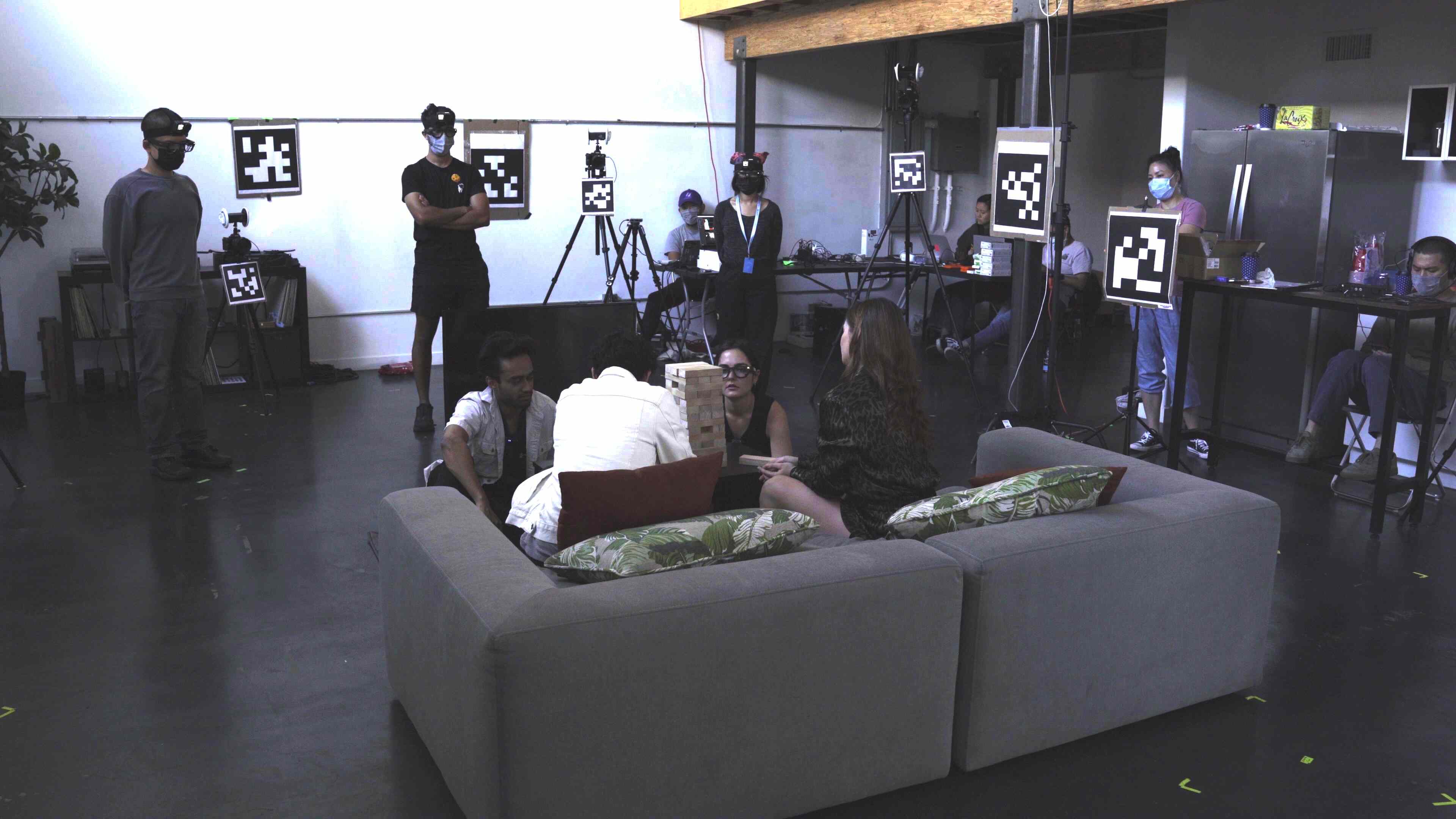} 
\sceneimage{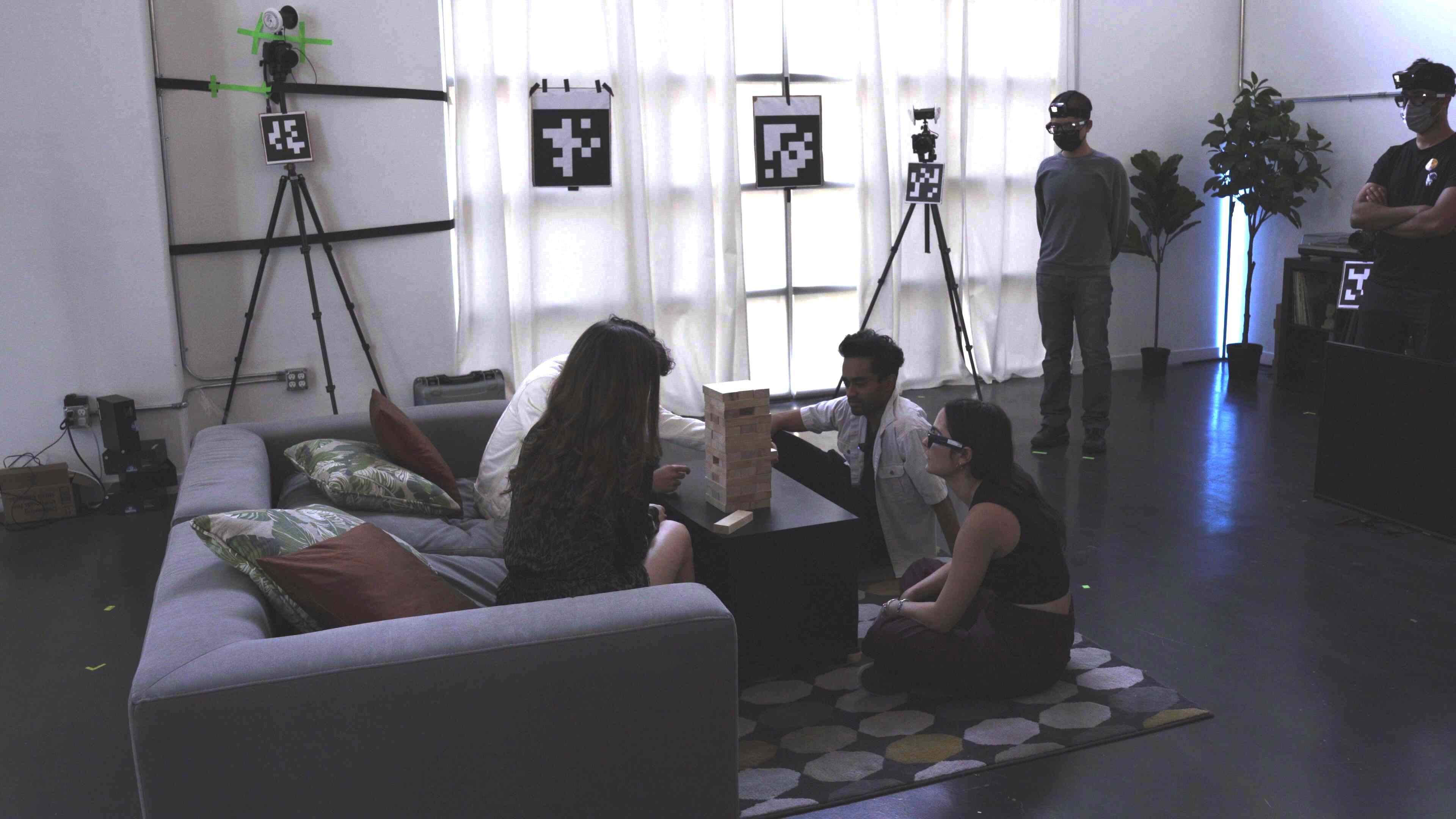}
\\
\sceneimage{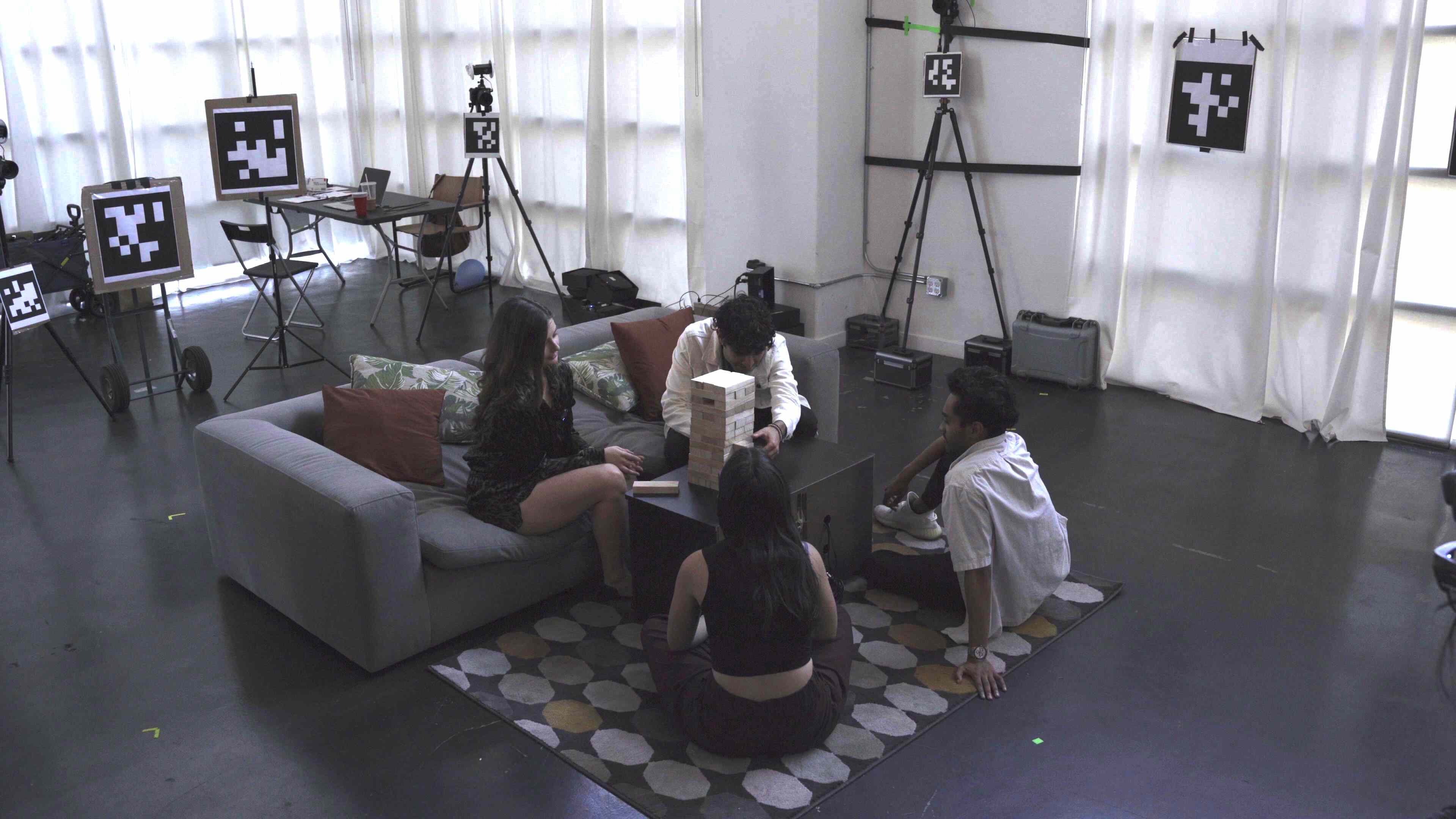} 
\sceneimage{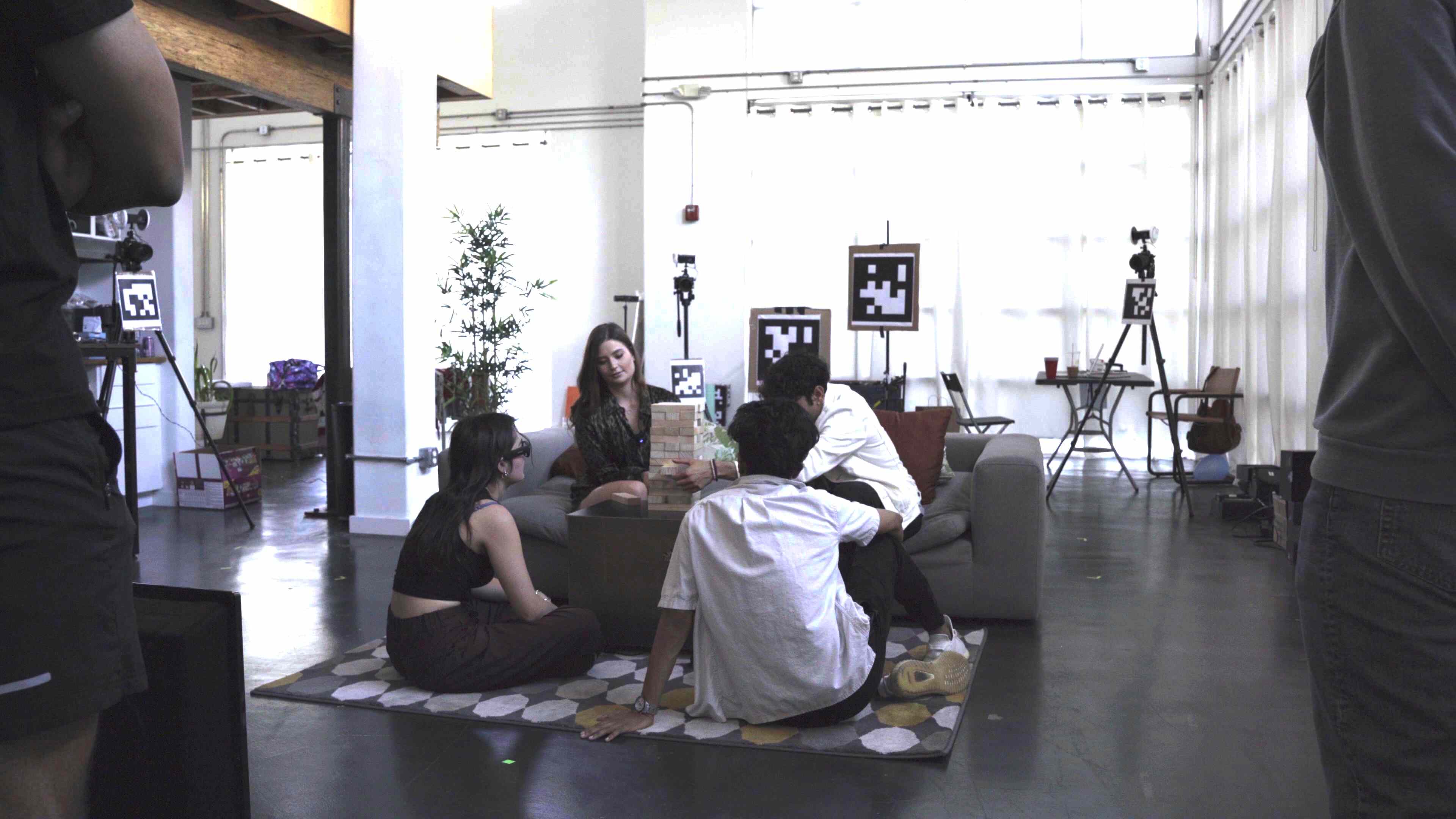} 
\sceneimage{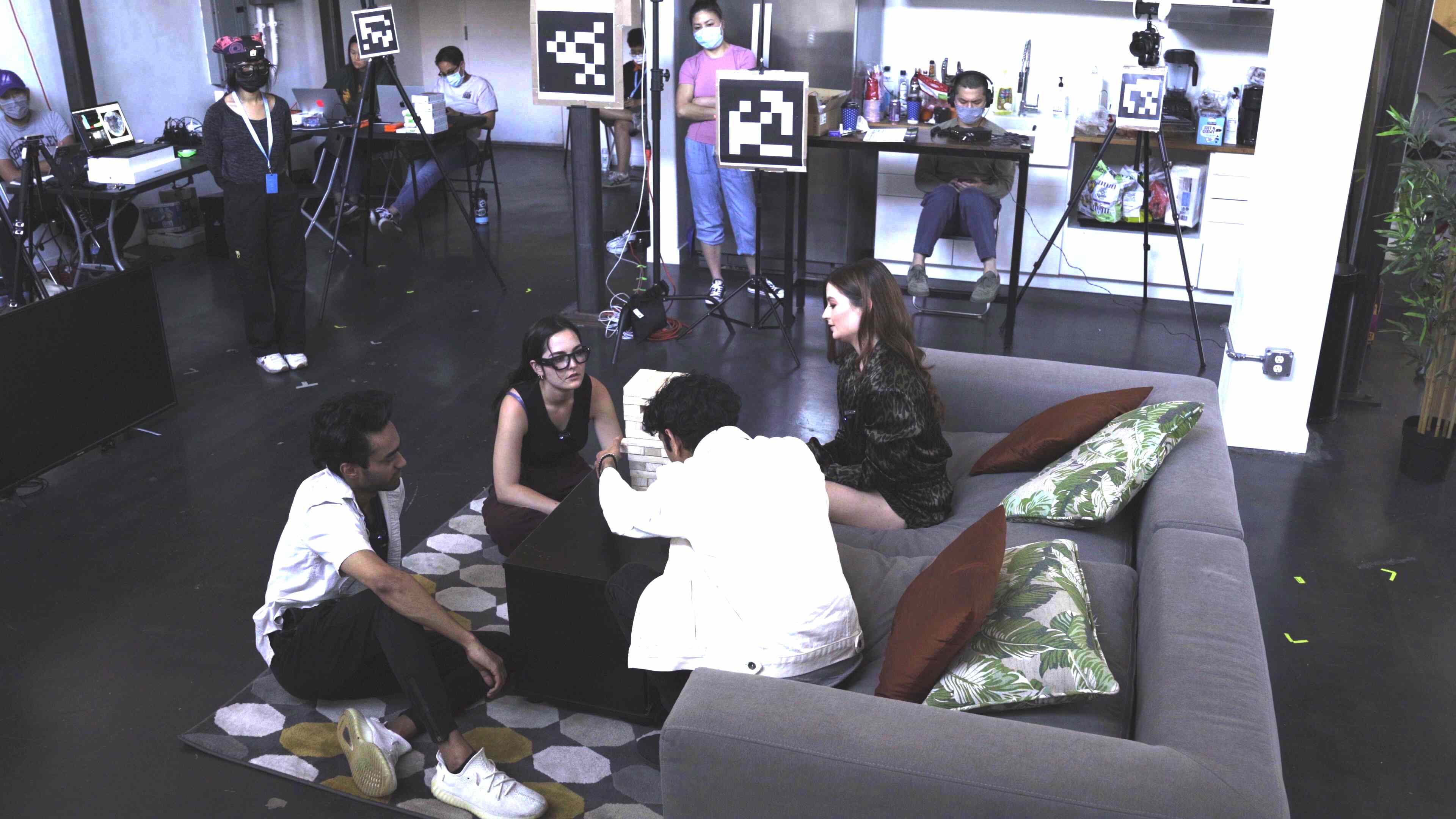} 
\sceneimage{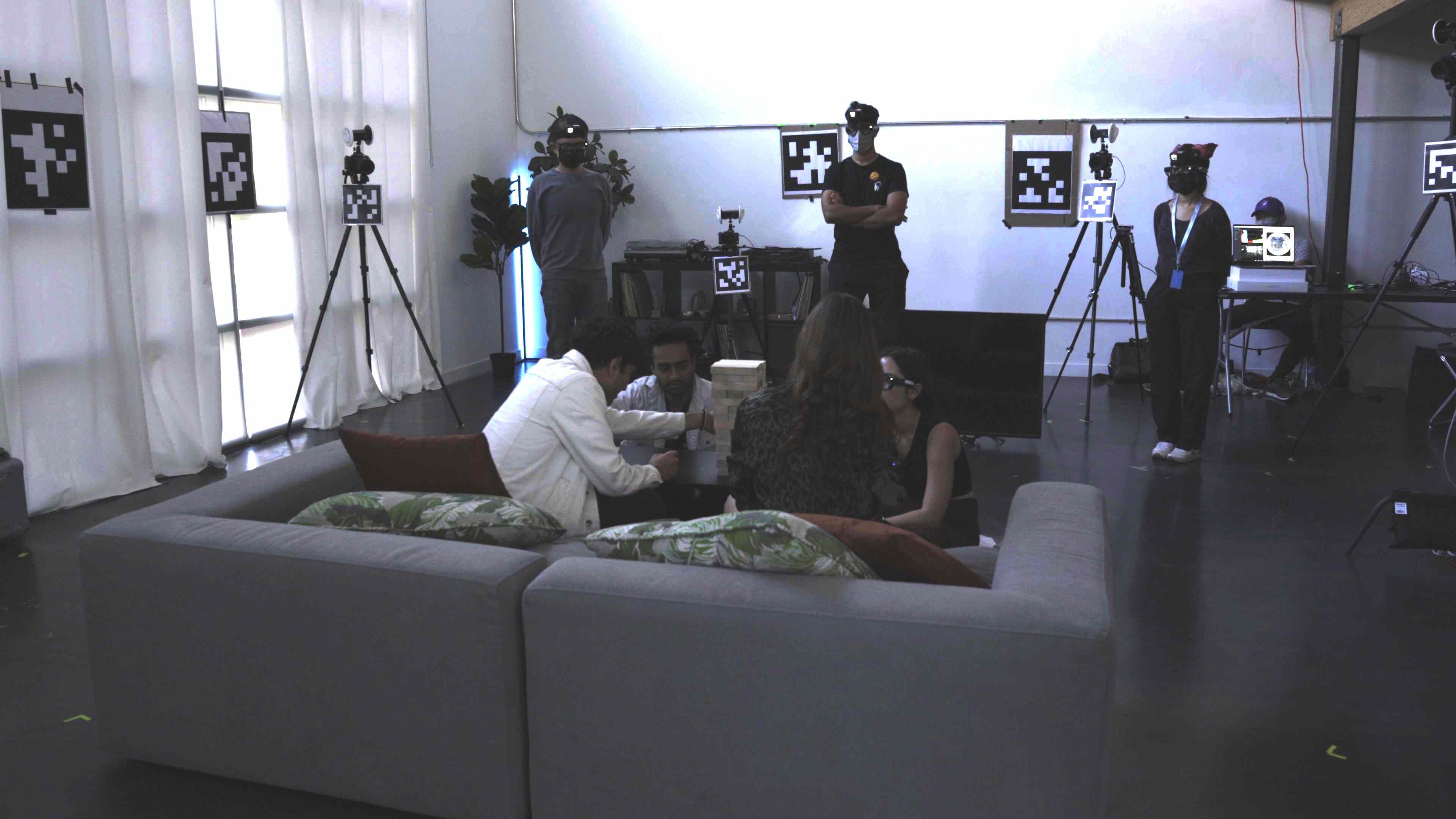}
\\
\sceneimage{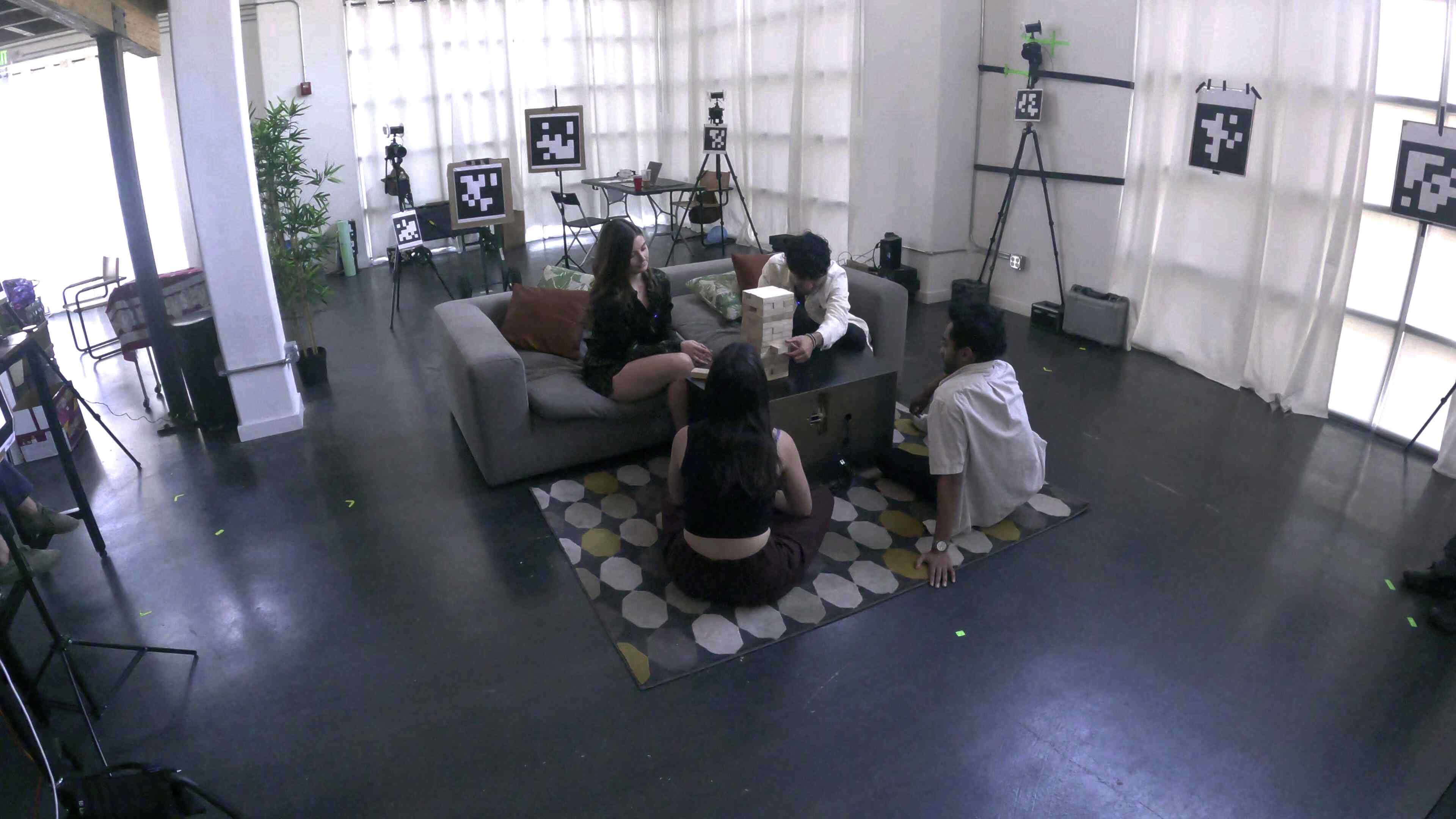} 
\sceneimage{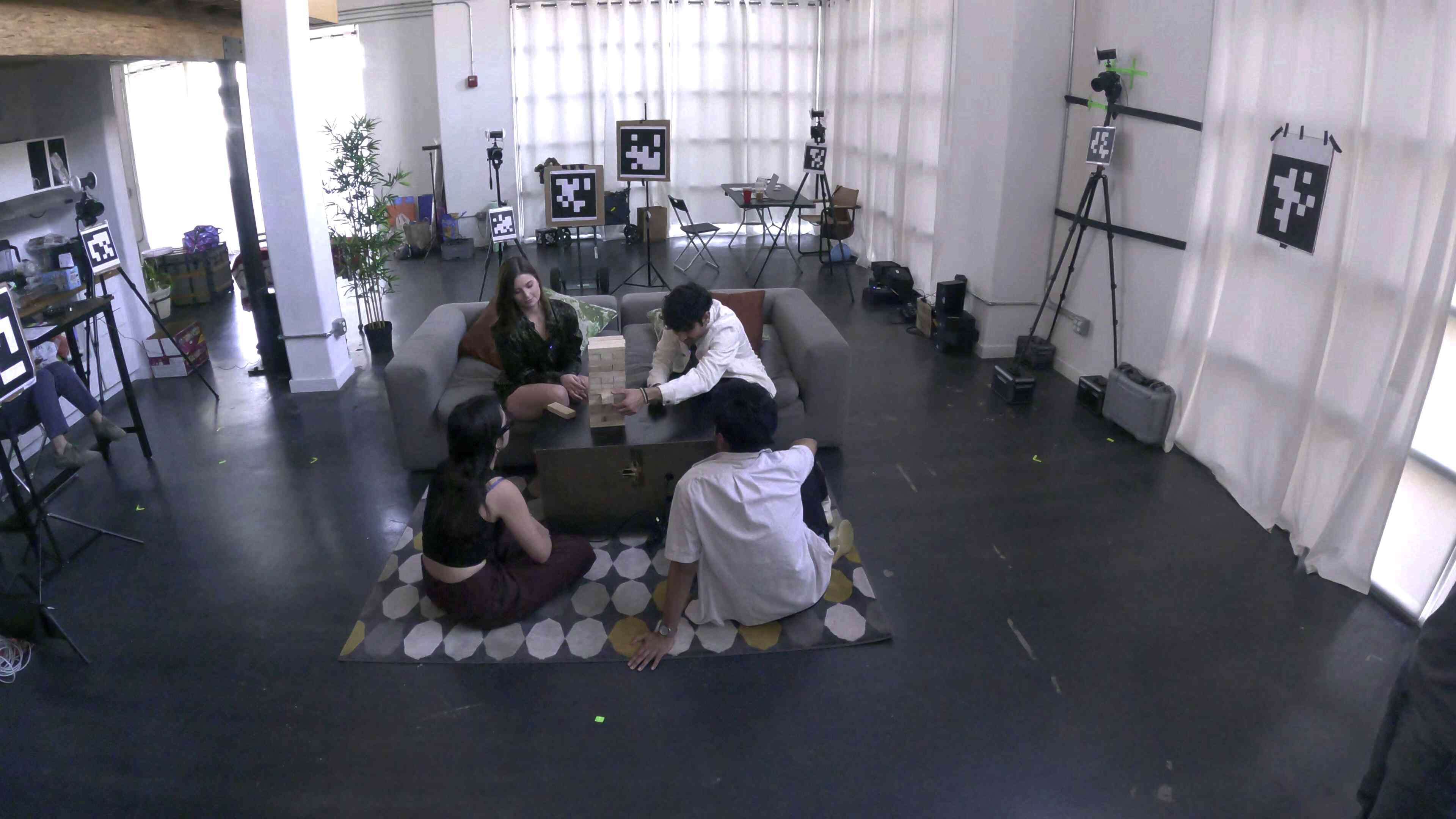} 
\sceneimage{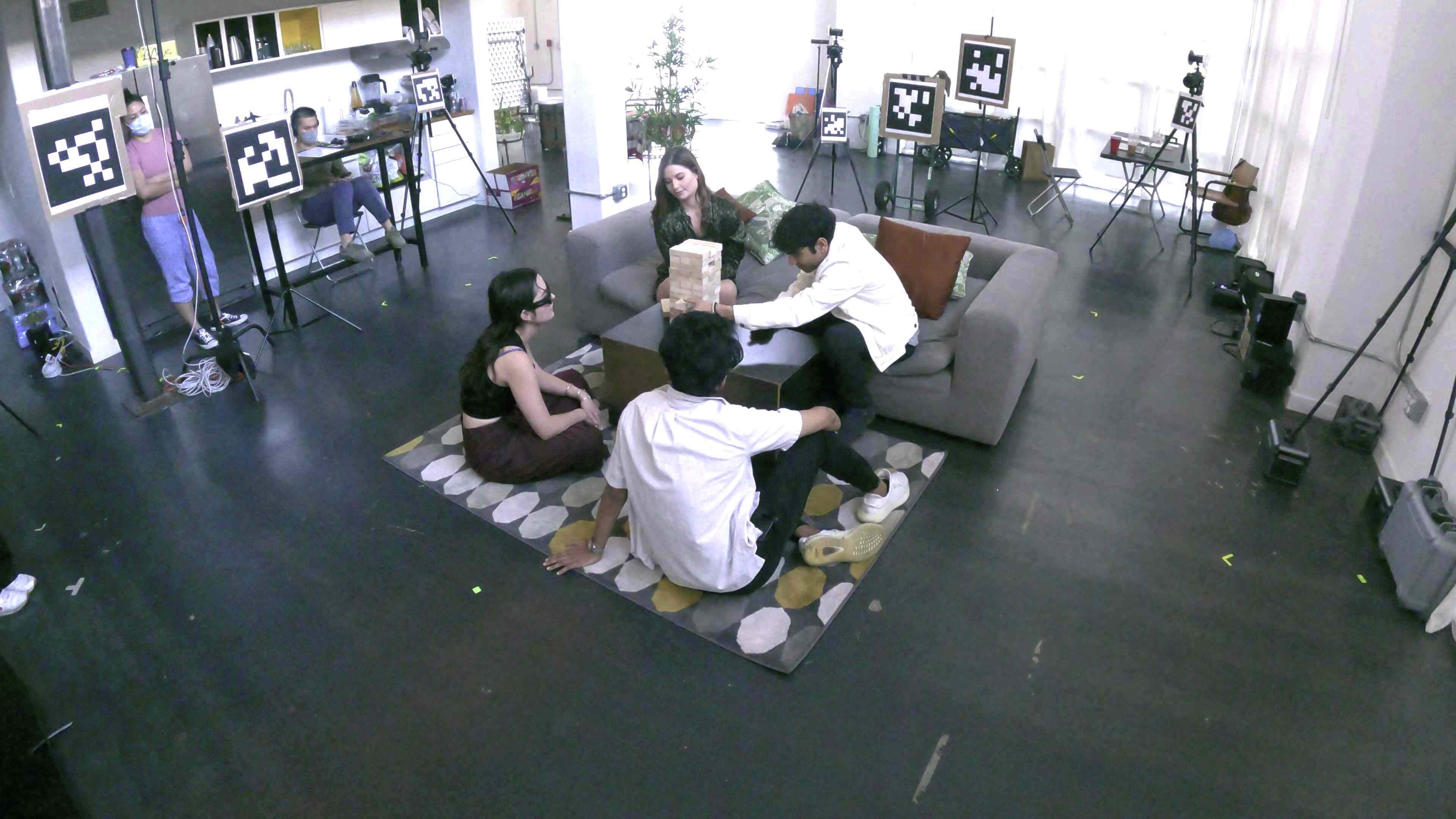} 
\sceneimage{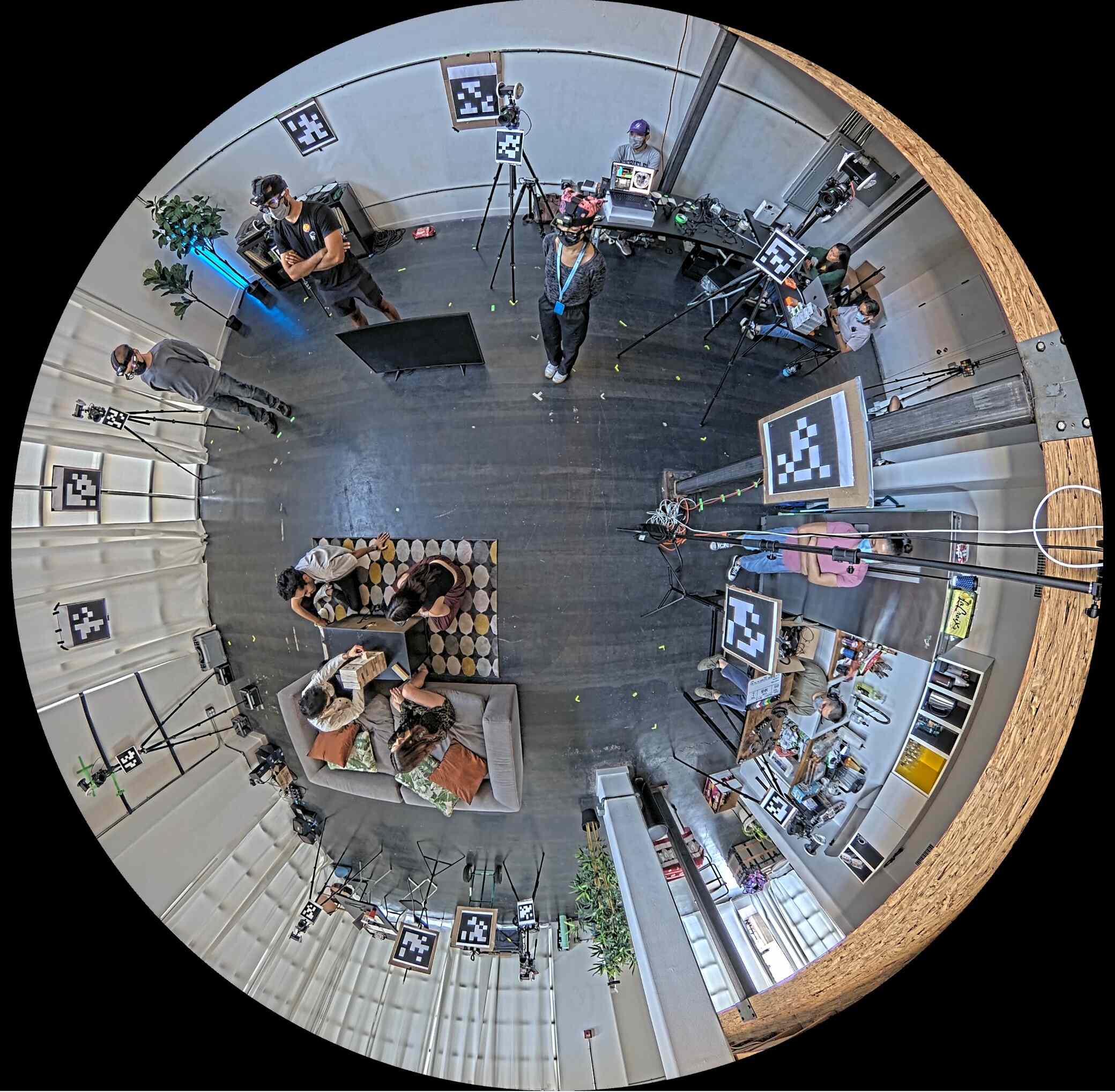}%

\end{tabular}

\let\sceneimage\undefined%

\caption{\textbf{Viewpoint diversity.} The \textbf{same moment in time} as viewed by the twelve different sensors we provide for each scene.
Top two rows: DSLR cameras. Bottom row: Head mounted GoPro cameras, and a 360\degrees ceiling camera.}\label{fig:all_sensors}
\end{figure*}

The \dataset dataset contains a large variety of scenes in terms of actions, number of participants, environments, and props. The scenes are acted out by a diverse cast of actors of different age, gender, and ethnicity. 
In the overview video and \cref{fig:all_scenes}, we show the representative sample of the different scenes one can find in the dataset.
In \cref{fig:all_sensors}, we show an example of a moment in time that was captured by the twelve visual sensors that we provide for each scene. 
All sensors are temporally synchronized: we provide frame timestamps in the time frame that is shared across sensors of each scene.

\section{Overview video and dynamic results}

We attach a video demonstrating:
\begin{itemize}
\item the diversity of the collected scenes (cf. \Cref{fig:all_scenes}),
\item results of the color calibration (cf. \Cref{fig:cc-before-after}),
\item visualisation of the estimated camera poses and static scene points,
\item the diversity of viewpoints (cf. \Cref{fig:all_sensors}) and foreground segmentation available for benchmark scenes,
\item the results of the novel-view synthesis when moving in space or time.
\end{itemize}

We show the novel-view synthesis rendered in three settings, specified in the slide titles. Below is the description of those.

\paragraph{Fly-around -- Fixed Timestamp.}

In this setting, we train a model on a fly-around stage where the camera used for training goes around the scene.
To render these videos, we generated camera poses by fitting a circle to camera centres from the training data.
The camera wearer went along an elliptical trajectory during the fly-around,
so approximating it with a circle forces the method to extrapolate the views in some parts of the trajectory.
For the models taking time as input, we fix the timestamp to the middle of the sequence.

\paragraph{Acting -- Fixed Timestamp.}

In this setting, we train the model using captures from the frontal cameras during the acting stage that contain substantial actors' motions.
Here we again fit a circle to the centres of cameras used for training and fix the timestamp to the middle of the stage, hence visualisations are static.
This setup shows how well the models are able to decouple viewpoints from motions in time.

\paragraph{Acting -- Fixed Novel Viewpoint.}

We additionally render the acting stage in fixed-viewpoint mode.
We fix the camera pose to the one used for evaluation (\ie, static DSLR-1, held out from training) and use the sub-sampled range of training timestamps.
This setting shows how well the methods can model the geometry and appearence changing in time.
In particular, time-agnostic methods (NeRF and TensoRF) are bound to produce static renders in this mode.
Please note that this is a viewpoint extrapolation scenario, since all training cameras were located at a significant distance from DLSR-1.

\graphicspath{{figure/colour-calibration}}

\newcommand{\imw}{4.2cm}%
\newcommand{\scenerow}[1]{%
\includegraphics[width=\imw,keepaspectratio]{before_sc-#1_DSLR-1.jpg}%
\includegraphics[width=\imw,keepaspectratio]{before_sc-#1_GOPRO-2.jpg}\hspace{0.5cm}%
\includegraphics[width=\imw,keepaspectratio]{after_sc-#1_DSLR-1.jpg}%
\includegraphics[width=\imw,keepaspectratio]{after_sc-#1_GOPRO-2.jpg}\vspace{0.1cm}\\
}%

\begin{figure*}[th]

\setlength\tabcolsep{0cm}%
\renewcommand{\arraystretch}{0.5}
\centering
\begin{tabular}{c}
\scenerow{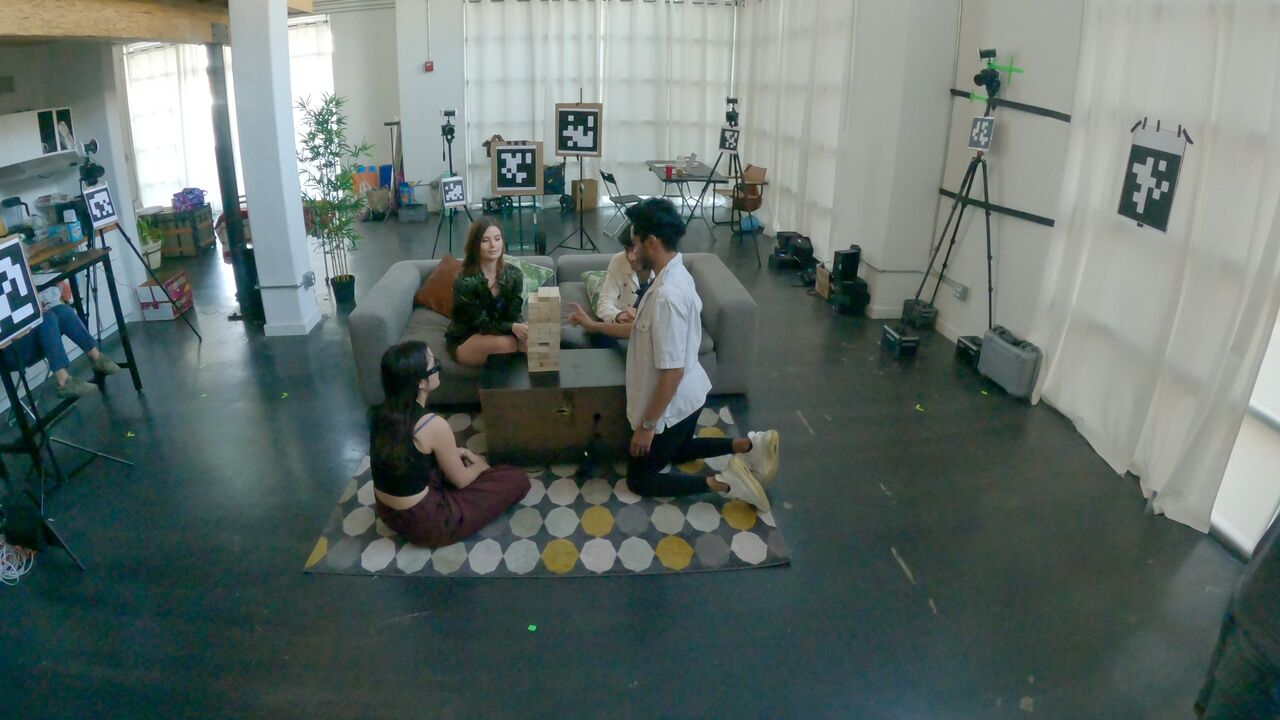}
\scenerow{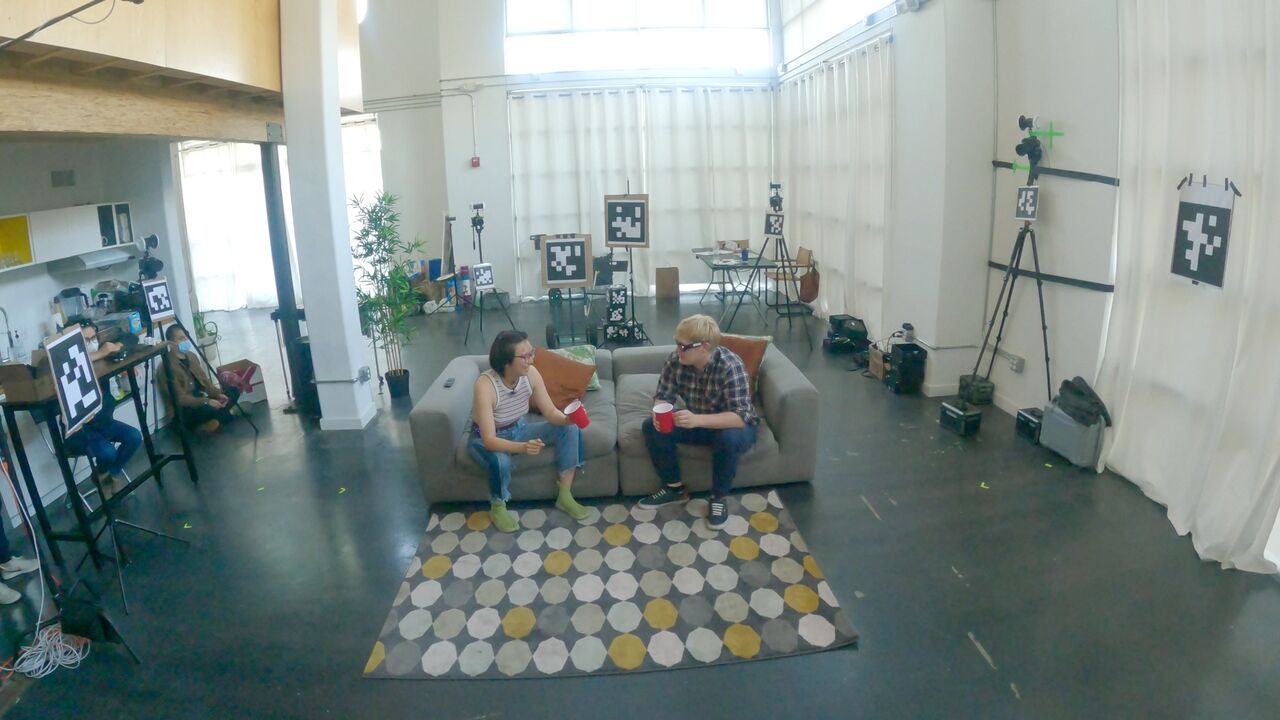}
\scenerow{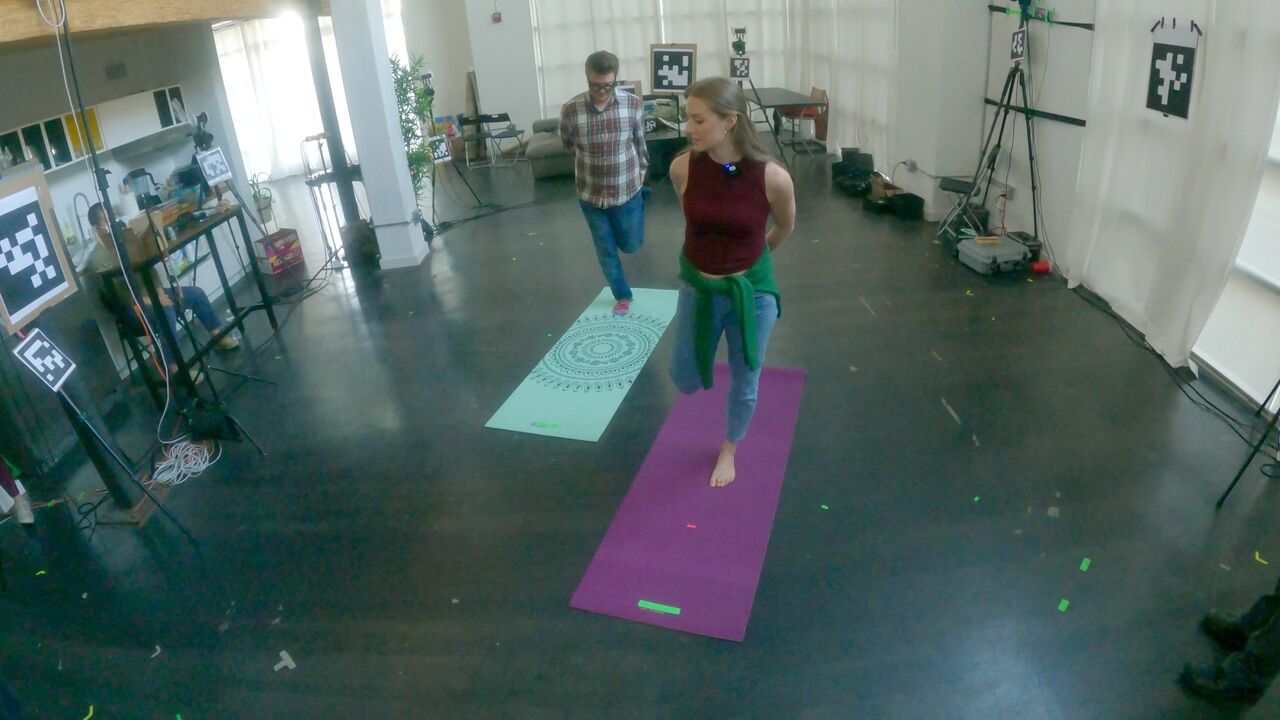}
\scenerow{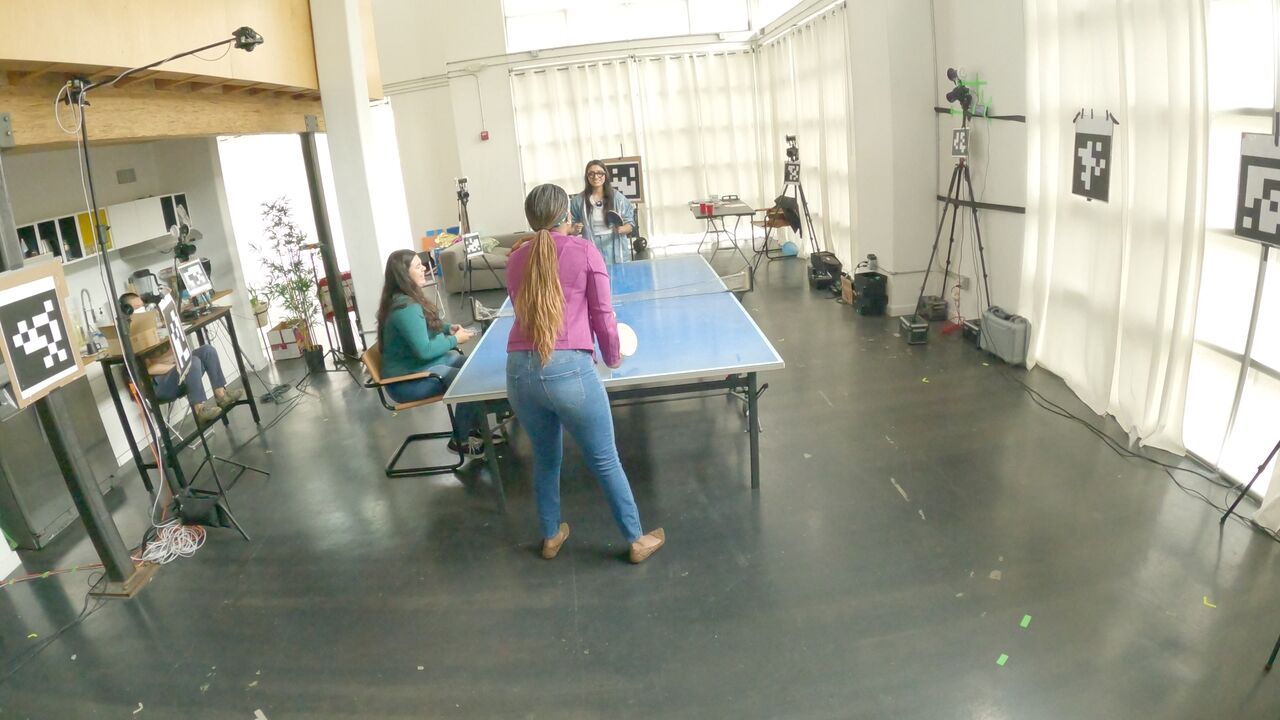}
\scenerow{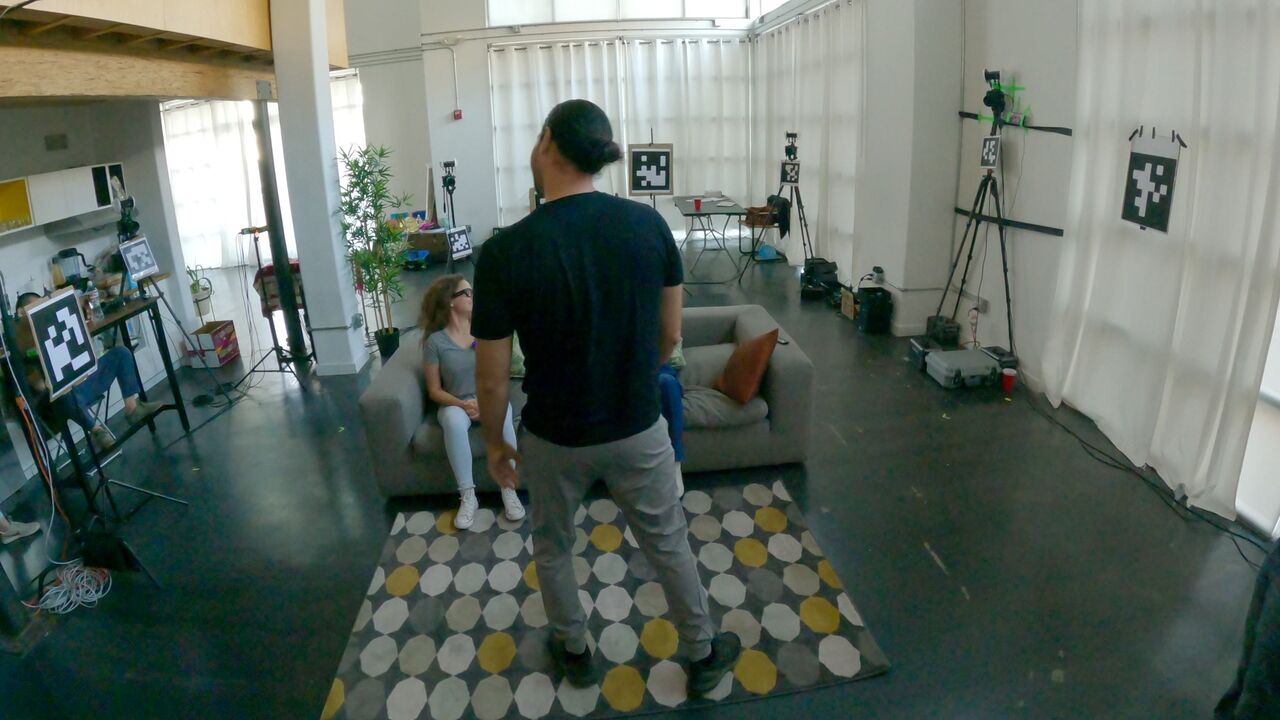}
\scenerow{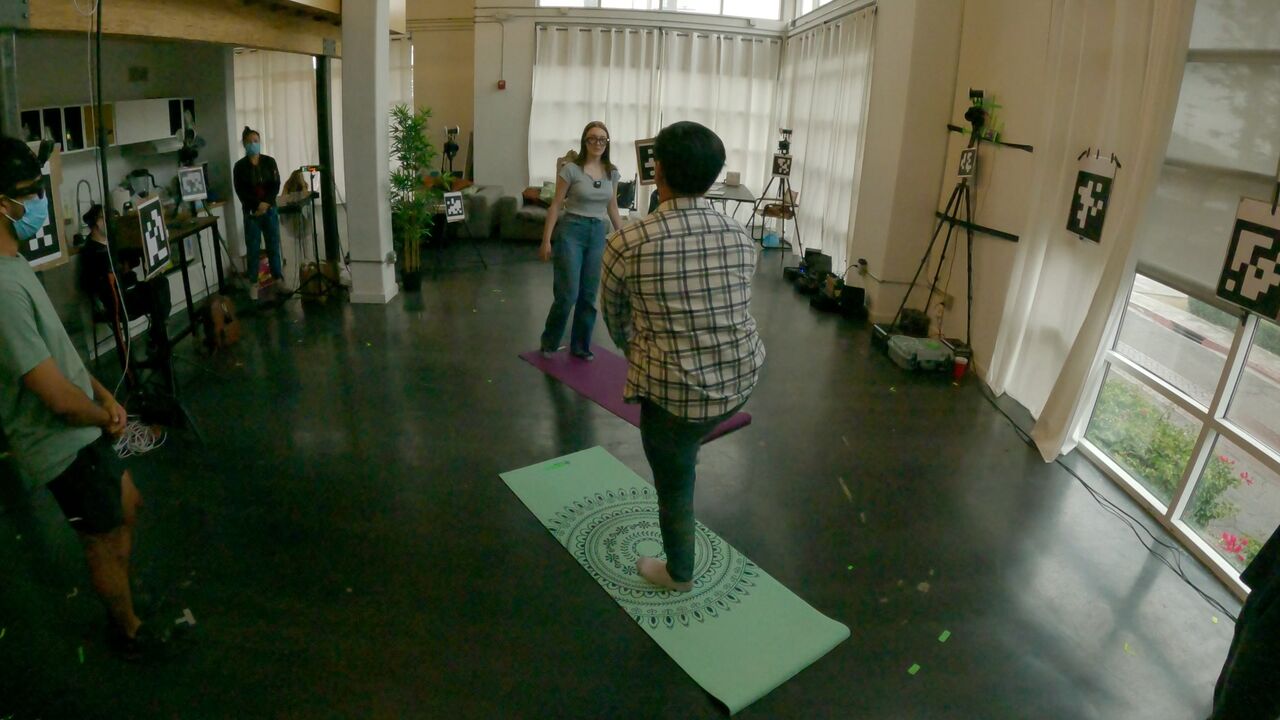}
\scenerow{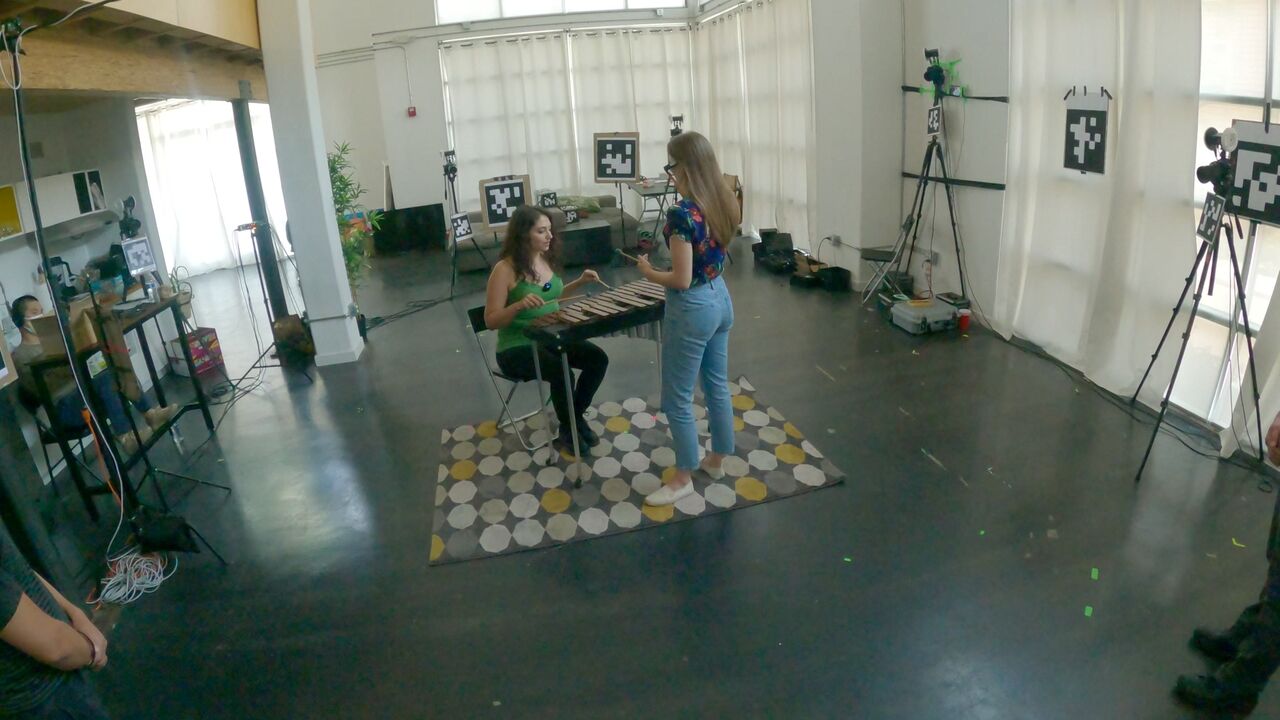}
\scenerow{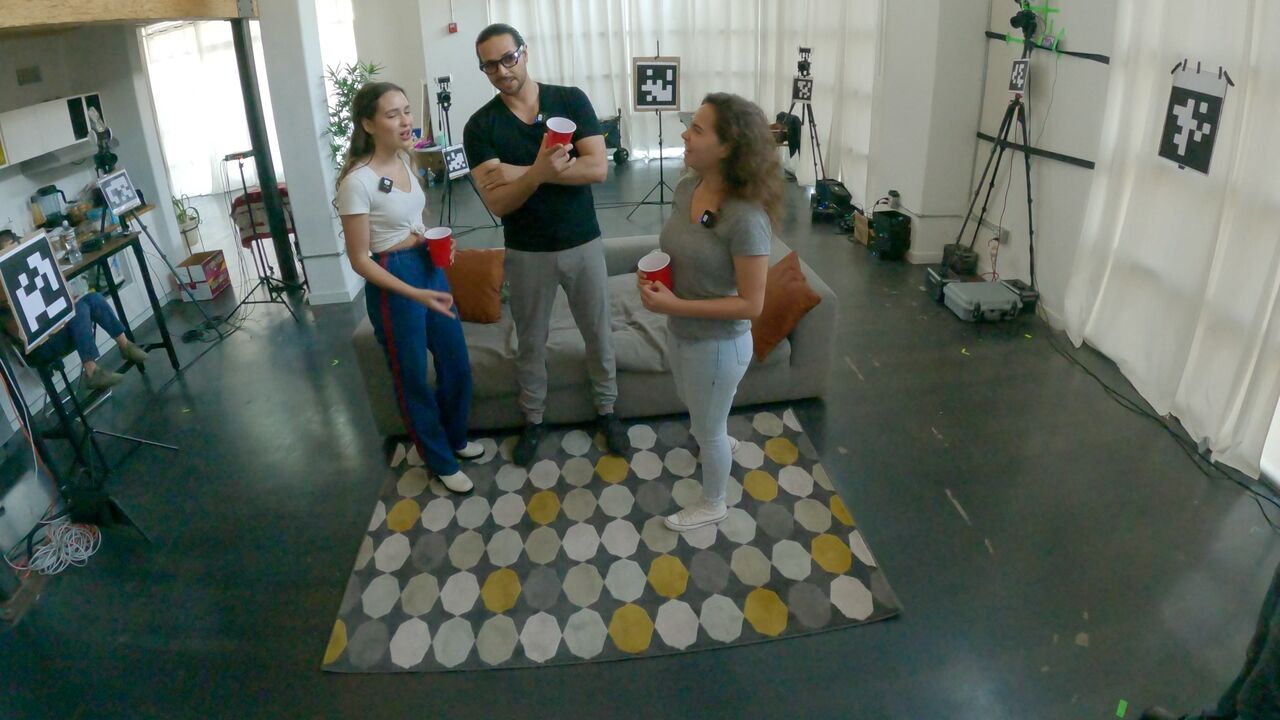}
\end{tabular}

\caption{\textbf{Color calibration.} For 8 different scenes (rows), we show a pair of frames from a DSLR and a GoPro sensor before color equalisation (left) and after equalisation (right).
Please note the discrepancy in colors on the left due to hardware differences.
We equalise images by transforming them into a common sRGB color space and matching a color-checker target.
Equalised colors enable combining sensors for training new-view synthesis methods and using a sensor of different type for evaluation.}\label{fig:cc-before-after}

\end{figure*}

\let\scenerow\undefined%
\let\imw\undefined%





\end{document}